\documentclass[letterpaper,twocolumn,10pt]{article}
\usepackage{usenix2019_v3}

\usepackage[utf8]{inputenc} 
\usepackage[T1]{fontenc}    
\usepackage{url}            
\usepackage{booktabs}       
\usepackage{amsfonts}       
\usepackage{nicefrac}       
\usepackage{balance}
\usepackage{microtype}      
\usepackage{makecell}
\usepackage{times}
\usepackage{soul}
\usepackage[small]{caption}
\usepackage{graphicx}
\usepackage{amsmath}
\usepackage{algorithm}
\usepackage{algorithmic}
\usepackage[algo2e,ruled,vlined]{algorithm2e}
\usepackage{xspace}
\usepackage{colortbl}
\usepackage{xcolor}
\usepackage{color}
\usepackage{listings}
\usepackage{epsfig}
\usepackage[title]{appendix}
\usepackage{makecell}
\usepackage{multirow}
\usepackage{enumitem}
\usepackage{subcaption}
\usepackage{caption}

\usepackage{tikz}
\usepackage{amsmath}

\usepackage{filecontents}

\usepackage{url}

\DeclareFixedFont{\ttb}{T1}{txtt}{bx}{n}{8}
\DeclareFixedFont{\ttm}{T1}{txtt}{m}{n}{8}

\usepackage{color}
\definecolor{deepblue}{rgb}{0,0,0.5}
\definecolor{deepred}{rgb}{0.6,0,0}
\definecolor{deepgreen}{rgb}{0,0.5,0}
\definecolor{purple}{rgb}{0.5,0,0.5}
\definecolor{gray}{rgb}{0.33,0.33,0.33}

\definecolor{dkgreen}{rgb}{0,0.6,0}
\definecolor{gray}{rgb}{0.5,0.5,0.5}
\definecolor{mauve}{rgb}{0.58,0,0.82}

\newcommand{\eat}[1]{}
\newcommand{\at}[1]{\protect\ensuremath{\mathsf{#1}}\xspace}
\newcommand{\stitle}[1]{\vspace{0.5ex}\noindent{\bf #1}}

\newcommand{\preeq}{\vspace{0mm}\begin{small}}
\newcommand{\posteq}{\vspace{0mm}\end{small}}

\newcommand{\system}{\textsc{HB}\xspace}
\newcommand{\lsystem}{\textsc{Hummingbird}\xspace}

\def\compactify{\itemsep=0pt \topsep=0pt \partopsep=0pt \parsep=0pt}
\let\latexusecounter=\usecounter
\newenvironment{CompactItemize}
  {\def\usecounter{\compactify\latexusecounter}
   \begin{itemize}}
  {\end{itemize}\let\usecounter=\latexusecounter}
\newenvironment{CompactEnumerate}
  {\def\usecounter{\compactify\latexusecounter}
   \begin{enumerate}}
  {\end{enumerate}\let\usecounter=\latexusecounter}

\newcommand{\mw}[1]{[\textcolor{blue}{MW: #1}]}
\newcommand{\KK}[1]{[\textcolor{purple}{KK: #1}]}

\newcommand{\ks}[1]{[\textcolor{cyan}{ks: #1}]}

\def\Snospace~{\S{}}
\def\Fnospace~{\mbox{Fig.\hspace{0.25em}}}
\def\Tnospace~{\mbox{Tab.\hspace{0.25em}}}
\def\Enospace~{\mbox{Equation\hspace{0.25em}}}

\title{A Tensor Compiler for Unified Machine Learning Prediction Serving}

\author{
	Supun Nakandala{\normalfont \textsuperscript{c,}}\thanks{The work was done while the author was at Microsoft.}~,
	Karla Saur{\normalfont \textsuperscript{m}},
	Gyeong-In Yu{\normalfont \textsuperscript{s,}\footnotemark[1]}~,
	Konstantinos Karanasos{\normalfont \textsuperscript{m}},\\
	Carlo Curino{\normalfont \textsuperscript{m}},
	Markus Weimer{\normalfont \textsuperscript{m}},
	Matteo Interlandi{\normalfont \textsuperscript{m}}\\[2mm]
	{\normalfont \textsuperscript{m}Microsoft,
	\textsuperscript{c}UC San Diego,
	\textsuperscript{s}Seoul National University}\\
 	{\small \texttt{\{<name>.<surname>\}@microsoft.com,snakanda@eng.ucsd.edu},~\texttt{gyeongin@snu.ac.kr}}
}

\usepackage[available,functional,reproduced]{usenixbadges}

\begin{document}

\makeatletter
\newcommand\notsotiny{\@setfontsize\notsotiny{9.0}{11}}
\newcommand\tablecolumnmarginsmall{\setlength\tabcolsep{3.5pt}}
\makeatother
\maketitle

\begin{abstract}
\vspace{-0ex}
\eat{
    Due to the wide adoption of ML based techniques in web, enterprises, and other domains, ML prediction serving has emerged as an important workload.
    To cater this demand, a new class of systems called \textit{prediction serving systems} have been developed (e.g., \textit{onnxruntime}, \textit{tensorrt}, and \textit{TorchScript}).
    However, the focus of almost all of these systems is on accelerating deep neural network prediction serving.
    Yet in many domains classical ML methods are still the preferred way to perform advanced data analytics, especially when dealing with structured and semi-structured data.
    In this work we present \system, a system to compile classical ML pipelines into tensor computations and thereby leverage the optimized systems developed for neural network prediction serving.
    We discuss the challenges, research directions, and promising initial empirical results.

Recent advances in Deep Neural Networks (DNNs) and the subsequent explosion of DNN frameworks have fostered the creation of a new class of systems. 
\KK{"the creation of a new class of ML inference systems" (or model scoring) would be better.}
ONNX \KK{I think you mean ONNX Runtime here.}, TVM, and TensorRT are notable examples of such systems: they share the same goal of providing a runtime for DNN model inference with state-of-the-art performance, ease of deployment on hardware accelerators (e.g., GPUs), and portability across platforms and devices.
Yet, in the enterprise space, data is mostly tabular and classical Machine Learning (ML) techniques, such as tree methods, are frequently used, often within complex pipelines composed of data featurizers and feature selection operators. 
Unfortunately, in the classical ML space no unified inference serving system exists. \KK{What do you mean by unified inference serving systems? I don't think it is clear here. For example, what is that systems for classical ML are lacking?} Therefore, developers are forced to resort to bespoke solutions or subpar performance.
In this work we present \system: a system able to compile classical ML pipelines end-to-end into tensor computations. It thereby seamlessly leverages the features provided by DNN inference systems, e.g., ease of deployment, operator optimizations and GPU support.
We discuss the challenges, our initial system prototype, and promising initial empirical results.

}

\eat{Machine Learning adoption across enterprises requires simpler and more efficient software infrastructure. The bespoke solutions typical in large web companies are simply untenable. The primary target for reducing infrastructure complexity and cost is model scoring (or inference)---since models are trained once but used many times and across different environments. 

In this paper, we propose \system, a novel approach to model scoring. \system compiles featurization operators and classical ML models to a narrow set of highly optimized core operators. This approach reduces infrastructure complexity. Furthermore, we show that this set of operators can be chosen to match common ones in Deep Neural Network (DNN), allowing us to leverage large existing investments in compilers/runtimes/specialized hardware. Our performance results are counter-intuitive:  despite replacing cheap sparse computations (tree traversals) with costly dense ones (tensor multiplications), our ability to better leverage instruction level and hardware parallelism (SIMD/GPU/IPUs/FPGAs), leads to massive speed-ups in thousands of benchmarks and production pipelines. \mw{Do we get to claim IPUs and FPGAs here?}\ks{i think it's premature to say it. and most people wouldn't know what IPU is yet, but have probably heard TPU} \system is competitive and even outperforms (by up to $3\times$) hand-crafted GPU kernels, and steadily beats state-of-the-art CPU solutions (by up to $2000\times$). \mw{Do we have average or minimal speedups as well? As it stands, this comes across as a bit braggalicious. It is deserved, but still :)}}

\vspace{-0.5ex}
Machine Learning (ML) adoption in the enterprise requires simpler and more efficient software infrastructure---the bespoke solutions typical in large web companies are simply untenable. Model scoring, the process of obtaining predictions from a trained model over new data, is a primary contributor to infrastructure complexity and cost as models are trained once but used many~times. 
In this paper we propose \lsystem, a novel approach to model scoring, which compiles featurization operators and traditional ML models (e.g., decision trees) into a small set of tensor operations. This approach inherently reduces infrastructure complexity and directly
leverages existing investments in Neural Network compilers and runtimes to generate efficient computations for both CPU and hardware accelerators. Our performance results are intriguing: despite replacing imperative computations (e.g., tree traversals) with tensor computation abstractions,
\lsystem is competitive and often outperforms hand-crafted kernels on micro-benchmarks on both CPU and GPU, while enabling seamless end-to-end acceleration of ML pipelines.
We have released \lsystem as open source.




\end{abstract}

\vspace{-1.5ex}
\section{Introduction}\label{sec:intro}
\vspace{-1.5ex}

Enterprises increasingly look to Machine Learning (ML) to help solve business challenges that escape imperative programming and analytical querying~\cite{flock}---examples include predictive maintenance, customer churn prediction, and supply-chain optimizations~\cite{firmai}. %
To do so, they typically turn to technologies now broadly referred to as {\em ``traditional ML''}, to contrast them with Deep Neural Networks (DNNs). A recent analysis by Amazon Web Services found that $50$ to $95\%$ of all ML applications in an organization are based on traditional ML~\cite{sagemaker-tco}. 
An analysis of $6M$ notebooks in public GitHub repositories~\cite{dsonds} paints a similar picture:
NumPy~\cite{numpy}, Matplotlib~\cite{matplot}, Pandas~\cite{pandas}, and scikit-learn~\cite{scikit} are the four most used libraries---all four provide functions for traditional ML. As a point of comparison with DNN frameworks, scikit-learn is used about $5$ times more than PyTorch~\cite{pytorch} and TensorFlow~\cite{tensorflow} combined, and growing faster than both. 
Acknowledging this trend, traditional ML capabilities have been recently added to DNN frameworks, such as the ONNX-ML~\cite{onnx-ml} flavor in ONNX~\cite{onnx} and TensorFlow's TFX~\cite{tfx}.

When it comes to owning and operating ML solutions, enterprises differ from early adopters in their focus on long-term costs of ownership and amortized return on investments~\cite{enterprise-app-lifespan}. As such, enterprises are highly sensitive to: (1)~complexity, (2)~performance, and (3)~overall operational efficiency of their software infrastructure~\cite{ai-cost}. 
In this work we focus on {\em model scoring} (i.e., the process of getting a prediction from a trained model by presenting it with new data), as it is a key driving factor in each of these regards.
First, each model is trained once but used multiple times for scoring in a variety of environments, thus scoring dominates infrastructure complexity for deployment, maintainability, and monitoring.
Second, model scoring is often in the critical path of interactive and analytical enterprise applications, hence its performance (in terms of latency and throughput) is an important concern for enterprises. Finally, {\em model scoring is responsible for 45-65\% of the total cost of ownership of data science solutions}~\cite{sagemaker-tco}. 

\stitle{Predictive Pipelines.} 
The output of the iterative process of designing and training traditional ML models is {\em not just} a model but a {\em predictive pipeline}: a Directed Acyclic Graph (DAG) of operators. Such pipelines are typically comprised of up to tens of operators out of a set of hundreds~\cite{dsonds} that fall into two main categories: (1)~{\em featurizers}, which could be either stateless imperative code (e.g., string tokenization) or data transformations fit to the data (e.g., normalization); and (2)~{\em models}, commonly decision tree ensembles or (generalized) linear models, fit to the data. Note that the whole pipeline is required to perform a prediction.

\begin{figure}[t]
\centering
\includegraphics[clip, trim=2cm 0cm 0.5cm 0cm, width=0.95\columnwidth]{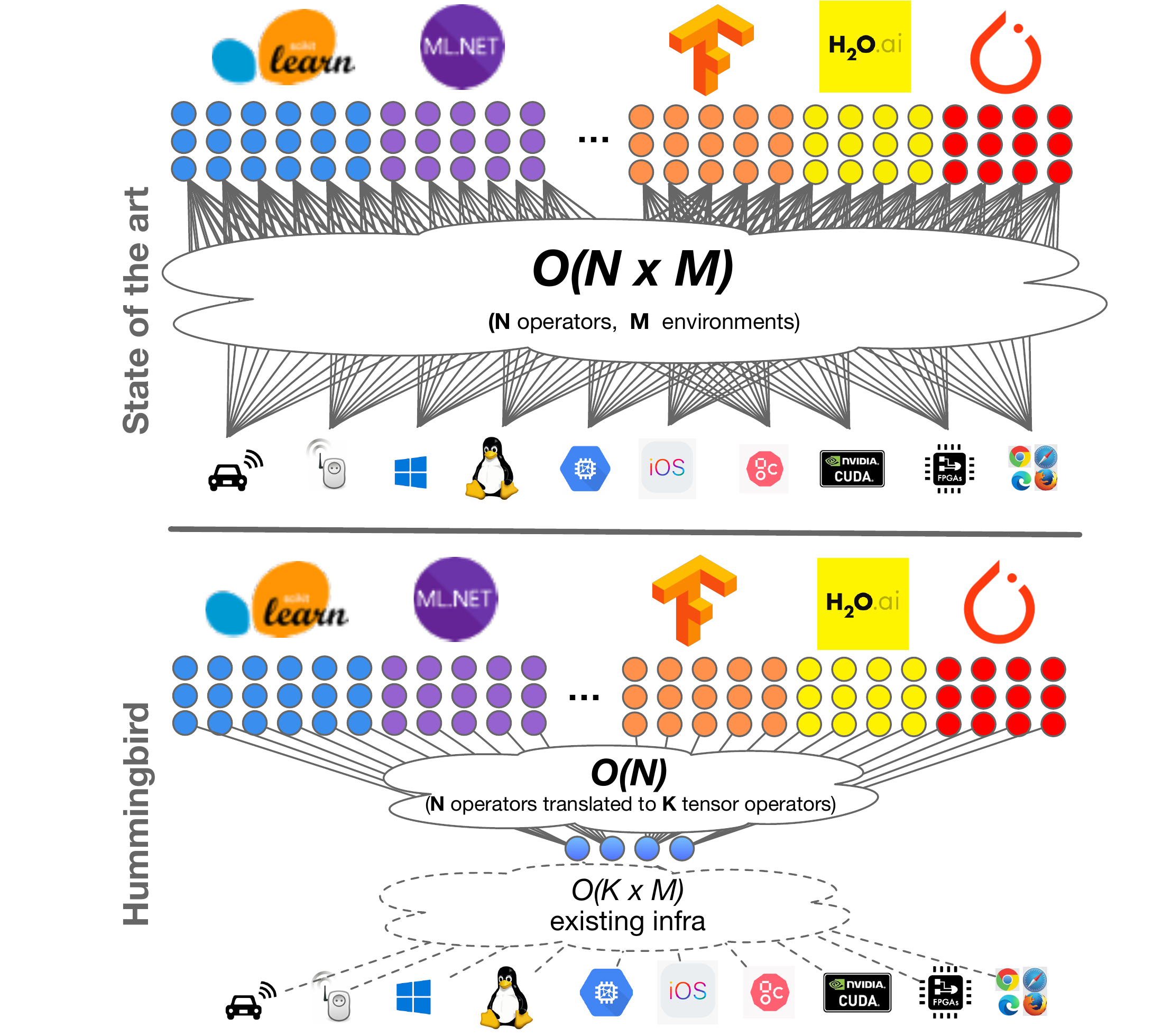}
\vspace{-2ex}
\caption{Prediction serving complexity: state-of-the-art (top) vs. \lsystem (bottom).
}
\label{fig:motivation}
\vspace{-3ex}
\end{figure}

\stitle{A Missing Abstraction.}
Today's featurizers and model implementations {\em are not expressed in a shared logical abstraction, but rather in an ad-hoc fashion} using programming languages such as R, Python, Java, C++, or C\#.
This hints to the core problem with today's approaches to model scoring: {\em the combinatorial explosion of supporting many operators (and frameworks) across multiple target environments}. 
Figure~\ref{fig:motivation} (top) highlights this visually by showing how existing solutions lead to an $O(N \times M)$ explosion to support $N$ operators from various ML frameworks against $M$ deployment environments (e.g., how to run a scikit-learn model on an embedded device?).
Furthermore, \cite{dsonds} shows that the number of libraries used in data science (a metric correlated to $N$) increased by roughly $4\times$ in the last 2 years.
Our expectation is that $M$ is also destined to grow as ML is applied more widely across a broad range of enterprise applications and hardware (e.g., \cite{graphcore,celebras,fpga,tpu,sambanova}). 
From the vantage point of implementing runtimes for model scoring, this is a daunting proposition. 
We argue that any brute-force approach directly tackling all combinations would dilute engineering focus leading to costly and less optimized solutions.
In fact, today, with very few exceptions (e.g., NVIDIA RAPIDS~\cite{rapids} for GPU), traditional ML operators are only implemented for CPUs.



This state of affairs is in contrast with the DNN space, where neural networks are authored using tensor transformations (e.g., multiplications, convolutions), providing an algebraic abstraction over computations.
Using such abstractions rather than imperative code not only enables evolved optimizations~\cite{xla,tvm}
but also facilitates support for diverse environments (such as mobile devices~\cite{mobile}, web browsers~\cite{tensorflowjs}, and hardware accelerators~\cite{graphcore,tpu,fpga}), unlocking new levels of performance and portability.


\stitle{Our Solution.}
To bypass this $N\times M$ explosion in implementing traditional ML operators, we built \lsystem (HB for short).
\system leverages compilation and optimization techniques to translate a broad set of traditional ML operators into a small set of $K$ core operators, thereby reducing the cost to $O(N) + O(K \times M)$, as shown in Figure~\ref{fig:motivation} (bottom). This is also the key intuition behind the ONNX model format~\cite{onnx} and its various runtimes~\cite{onnx-supported}. However, with \system we take one further bold step: we demonstrate that this set of core operators can be reduced to tensor computations and therefore be executed over DNN frameworks. This allows us to piggyback on existing investments in DNN compilers, runtimes, and specialized-hardware, and reduce the challenge of ``running $K$ operators across $M$ environments'' for traditional ML to just $O(N)$ operator translations.
This leads to improved performance and portability, and reduced infrastructure complexity. 

\stitle{Contributions.} 
In this paper we answer three main questions:
\begin{CompactEnumerate}
    \item Can traditional ML operators (both linear algebra-based such as linear models, and algorithmic ones such as decision trees) be translated to tensor computations?
    \item Can the resulting computations in tensor space be competitive with the imperative alternatives we get as input (e.g., traversing a tree)?
    \item Can \system help in reducing software complexity and improving model portability? 
\end{CompactEnumerate}

Concretely, we: (1)~port thousands of benchmark predictive pipelines to two DNN backends (PyTorch and TVM); (2)~show that we can seamlessly leverage hardware accelerators and deliver speedups of up to $3\times$ against hand-crafted GPU kernels, and up to $1200\times$ for predictive pipelines against state-of-the-art frameworks; 
and (3)~qualitatively confirm improvements in software complexity and portability by enabling scikit-learn pipelines to run across CPUs and GPUs. 

\system is open source under the MIT license~\footnote{\url{https://github.com/microsoft/hummingbird}}, and is part of the PyTorch ecosystem~\cite{pytorch-ecosystem}. We are integrating HB with other systems, such as the ONNX converters~\cite{blog}.


\stitle{Organization.} The remainder of the paper is organized as follows. Section~\ref{sec:background} provides some background, and Section~\ref{s:overview} presents an overview of \system.
Section~\ref{sec:compilation} describes the compilation from traditional ML to tensor computations, whereas Section~\ref{sec:optimization} discusses various optimizations. Section~\ref{s:experiments} presents our evaluation. Section~\ref{s:related} is related work, then we conclude.

\eat{\system is subject to active development so our performance results are likely to change as we further optimize our use of DNN backends as well as our translation steps. Furthermore, we are working on open-sourcing \lsystem. Both of these aspects are subject of ongoing engagements with ONNX-Runtime, ML.NET, scikit-learn and TVM OSS communities, and we will report on them in greater details in the camera ready version, if the paper is accepted.}


\eat{

Machine Learning had phenomenal success in ``unicorn'' applications, such as search ranking, spam detection, recommender systems, speech recognition, natural language processing, and computer vision. This class of applications is characterized by: 1) a small number of applications, 2) each worth billions of dollars, and 3) frequently built around large Deep Neural Networks (DNN). This has justified complex and costly custom infrastructures, and is driving huge investments in frameworks and specialized hardware for DNNs \cite{ipu,tf-serving,sambanova,tpu}.
  
This success did not go unnoticed, and nearly every large enterprise is applying ML to many aspects of its business. Enterprise applications of ML however differ from the early kind in few fundamental ways: 1) they are very numerous (millions of applications) 2) each is likely worth (only) millions of dollars, and 3) they are primarily built on modest amounts of tabular and text data, an area dominated by classical ML: feature extraction, trees and linear models~\cite{flock,kaggle-survey,dsonds}.

\mw{Here and elsewhere, we talk about applications. I think that maybe we should talk about models instead. The same model (e.g. an OCR model) can be used in many applications. That in itself doesn't create the challenge we aim to solve: many models are used in many applications, each model only once.}

This makes enterprises more sensitive to operational costs \cite{ai-cost}, and prevents them from piggybacking on the investment made for unicorn applications\footnote{By contrast, in the world of BigData, enterprises were able to leverage web companies investments in frameworks such as Hadoop, HDFS, YARN, Spark, and Kubernetes.}. Furthermore enterprise applications have more challenging portability requirements due to their sheer number and longer lifespan\cite{enterprise-app-lifespan}. This creates intense pressure towards reducing software complexity and operational costs, as DNN-based bespoke infrastructures are simply untenable in this new setting. 

\mw{I am not sure this follows as strictly as we'd like. There is plenty of evidence that e.g. TF has usage in enterprise applications. And plenty of infra for DNN scoring exists. I believe we might have to unfold this and bring the whole "classical ML implies Python containers" argument in here. That then is untennable for enterprise ML, and we can help free-ride on the investments into DNN scoring infra.}

One of the worst offenders in terms of infrastructure complexity and resource cost is {\em model scoring}---the process of presenting a model with new data to get a prediction. The reason is simple: models are trained infrequently, often in resource-rich and uniform cloud environments, but they are used for prediction many times, often across a variety of environments (e.g., scale-out batch or interactive serving, personal computers, mobile and IoT devices). Looking just at resource costs: {\em model scoring contributes 45-65\% of the total cost of ownership} of a data science solution. This is true even when including support personnel costs, and for a rather simple cloud-native solution, while {\em comparing infrastructure costs model scoring is $7\times$ to $24\times$ more expensive than training}~\cite{sagemaker-tco}.
}

\vspace{-2ex}
\section{Background and Challenges}
\label{sec:background}
\vspace{-2ex}

We first provide background on traditional ML and DNNs. We then explain the challenges of compiling traditional ML operators and predictive pipelines into tensor computations. 




\eat{
\begin{figure}[t]
\includegraphics[width=\columnwidth]{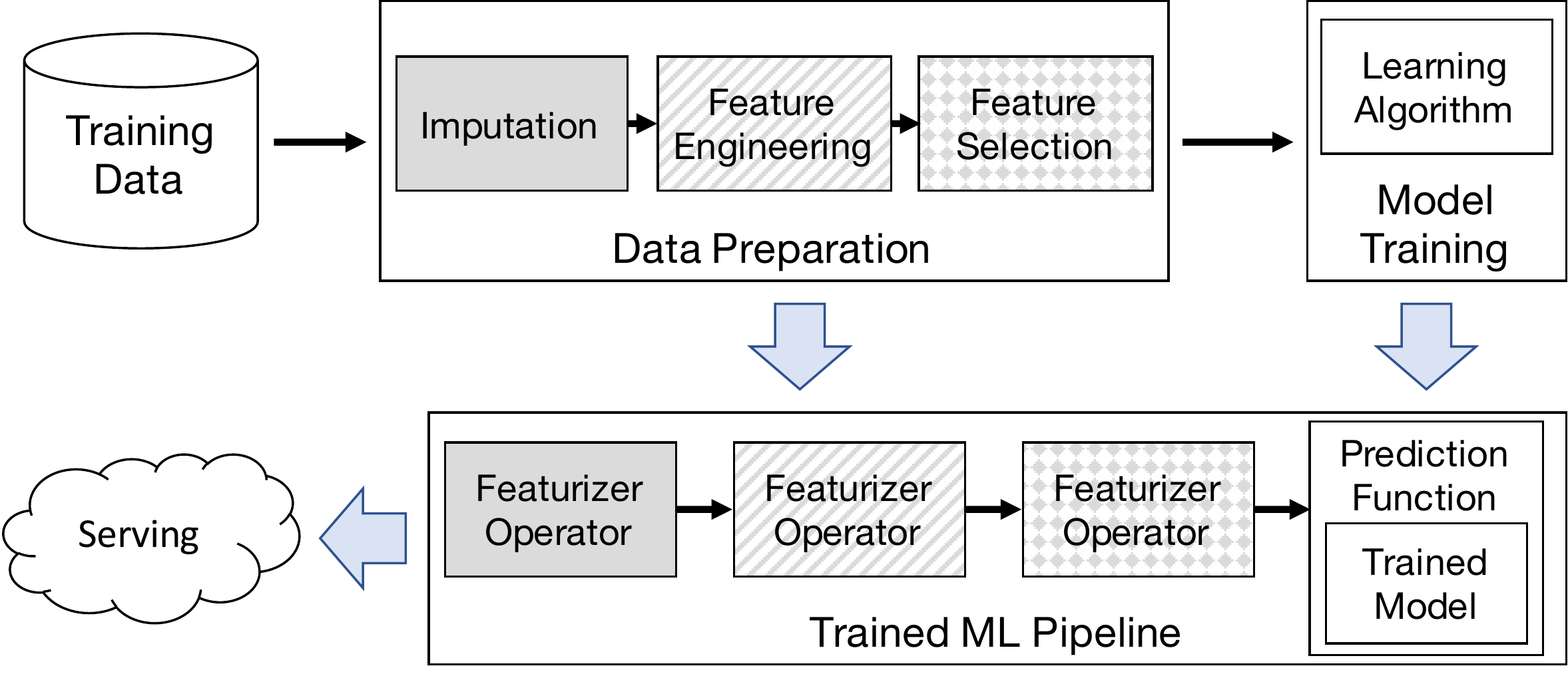}
\caption{Typical traditional ML workflow.}
\vspace{-4mm}
\label{fig:workflow}
\end{figure}

\mw{I did not like the figure and text. The above is my attempt at an alternative paragraph}
\stitle{Traditional ML.} Figure \ref{fig:workflow} depicts the traditional ML workflow. Starting with some \emph{input data}, a \emph{data preparation} step is used for data cleaning (e.g., completion of missing values through imputers~\cite{impute}), feature generation~\cite{preprocessing} and selection.
Output of the data preparation step is then passed to the \emph{training} step, where an ML model is trained using some learning procedure.
The trained ML model, along with the data featurizer operators constitute the \emph{predictive pipeline}, which essentially transforms input features into a prediction score (e.g., 1 or 0 for binary classification).
Wrapping a trained pipeline into a single artifact is common practice~\cite{mldotnet}.
These artifacts are then embedded inside host applications or containerized and deployed in the cloud for \emph{serving} prediction queries~\cite{DBLP:journals/sigmod/PolyzotisRWZ18, clipper2}.\footnote{Note that this is an oversimplification of actual ML workflow, and it does not cover for example hyper-parameter tuning, model selection, or cases in which ML models are used as featurizers. It is however a fair summary of most common use-cases.}
scikit-learn (Python-based), {\sc ml.net} ({\sc .net}-based), and H$_2$O (Java-based) are popular software tools that one could use to train and generate pipelines. However, it is important to note that they are primarily optimized for training and not for serving.
}

\vspace{-1.5ex}
\subsection{Traditional ML and DNNs}
\vspace{-1ex}
\stitle{Traditional Predictive Pipelines.} 
The result of the data science workflow over traditional ML are predictive pipelines, i.e., DAG of operators such as trained models, preprocessors, featurizers, and  missing-value imputers. The process of presenting a trained predictive pipeline with new data to obtain a prediction is referred to in literature interchangeably as: model scoring/inference/serving, pipeline evaluation, or prediction serving. We favor model scoring in our writing.

Packaging a trained pipeline into a single artifact is common practice~\cite{mldotnet}. These artifacts are then embedded inside host applications or containerized and deployed in the cloud to perform model scoring~\cite{DBLP:journals/sigmod/PolyzotisRWZ18, clipper2}. ML.NET~\cite{mldotnet} (.NET-based), scikit-learn~\cite{scikit} (Python-based), and H$_2$O~\cite{h2o} (Java-based) are popular toolkits to generate pipelines. However, they are primarily optimized for training. Scoring predictive pipelines is challenging, as their operators are implemented in imperative code 
and do not follow a shared abstraction. Supporting every operator in all target environments requires a huge effort, which is why these frameworks have limited portability. 


\stitle{DNNs.} Deep Neural Networks (DNNs) are a family of ML models that are based on artificial neurons~\cite{dl-book}. They take raw features as input and perform a series of transformation operations. Unlike traditional ML, transformations in DNNs are drawn from a common abstraction based on tensor operators (e.g., generic matrix multiplication, element-wise operations). 
In recent years, DNNs have been extremely successful in vision and natural language processing tasks~\cite{alexnet,bert}. Common frameworks used to author and train DNNs are TensorFlow~\cite{tensorflow}, PyTorch~\cite{pytorch}, CNTK~\cite{cntk}, and MXNet~\cite{mxnet}. While these frameworks can also be used to perform model scoring, next we discuss systems specifically designed for that. 

\stitle{Runtimes for DNN Model Scoring.} To cater to the demand for DNN model inference, a new class of systems has emerged. ONNX Runtime (ORT)~\cite{onnx-runtime} and TVM~\cite{tvm} are popular examples of such systems. These capitalize on the relative simplicity of neural networks: they accept a DAG of tensor operations as input, which they execute by implementing a small set of highly optimized operator kernels on multiple hardwares. Focusing on just the prediction serving scenario also enables these systems to perform additional inference-specific optimizations, which are not applicable for training. 
\system is currently compatible with all such systems. 
\eat{
\begin{CompactItemize}
    \vspace{-1mm}
    \item ORT executes ONNX~\cite{onnx} models.
    ONNX is an exchange format both for DNN models and traditional ML models (ONNX-ML~\cite{onnx-ml}).
    While for DNN models the ORT provides hardware acceleration, for ONNX-ML models currently only CPU execution is supported. This is because each ONNX-ML operator is implement by its own algorithmic logic which makes no use of tensor operations. ORT provides a set of graph optimizations such as constant folding and operator fusion~\cite{onnx-opt}.
    
    \vspace{-2mm}
    \item TorchScript allows to compile PyTorch models into pure C++ kernels which can then be executed without having to rely on the Python runtime.
    When PyTorch models are compiled into TorchScript using the JIT infrastructure, a set of algebraic and graph rewrites are applied to the input model~\cite{ts-jit}.
    
    \vspace{-2mm}
    \item Apache TVM executes models implemented its IR called Relay~\cite{relay}. TVM supports computational graph optimizations such as operator fusion and layout transformation, as well as low-level optimizations such as memory management, automatic generation of optimized kernels, and target hardware-specific tuning
\end{CompactItemize}
}

\vspace{-2ex}
\subsection{Challenges}\label{ss:challenges}
\vspace{-1.5ex}
\system combines the strength of traditional ML pipelines on structured data~\cite{tradMLtraining} with the computational and operational simplicity of DNN runtimes for model scoring. 
To do so, it relies on a simple yet key observation: once a model is trained,  it can be represented as a \emph{prediction function} transforming input features into a prediction score (e.g., 0 or 1 for binary classification), regardless of the training algorithm used. The same observation naturally applies to featurizers fit to the data.
Therefore, \system only needs to compile the prediction functions (not the training logic) for each operator in a pipeline into tensor computations and stitch them appropriately.
Towards this goal, 
we identify two challenges. 
\vspace{-1mm}
\begin{enumerate}[label={},wide, labelindent=1pt]
    \vspace{-1mm}
    \item {\bf Challenge 1}: \emph{How can we map traditional predictive pipelines into tensor computations?}
    Pipelines are generally composed of operators (with predictive functions) of two classes: \emph{algebraic} (e.g., scalers or linear models) and \emph{algorithmic} (e.g., one-hot encoder and tree-based models).
    While translating algebraic operators into tensor computations is straightforward, the key challenge for \system is the translation of algorithmic operators.
    Algorithmic operators perform arbitrary \emph{data accesses and control flow decisions}.
    For example, in a decision tree ensemble potentially every tree is different from each other, not only with respect to the structure, but also the decision variables and the threshold values.
    Conversely, tensor operators perform \emph{bulk operations} over the entire set of input elements.
    
    \vspace{-2mm}
    \item {\bf Challenge 2}: \emph{How can we achieve efficient execution for tensor-compiled traditional ML operators?}
    The ability to compile predictive pipelines into DAGs of tensor operations does not imply adequate performance of the resulting DAGs. In fact, common wisdom would suggest the opposite: even though tensor runtimes naturally support execution on hardware accelerators, 
    tree-based methods and commonly used data transformations are well known to be difficult to accelerate~\cite{inferline}, even using custom-developed implementations.
\end{enumerate}

\vspace{-2mm}
\vspace{-2ex}
\section{System Overview}
\label{s:overview}
\vspace{-2ex}
In this section we explain our approach to overcome the challenges outlined in Section~\ref{ss:challenges}, and present \system's architecture and implementation details. We conclude this section by explaining assumptions and limitations.

\vspace{-2ex}
\subsection{High-level Approach}
\vspace{-1ex}

In \system, we cast algorithmic operators into tensor computations.
You will notice that this transformation \emph{introduces redundancies}, both in terms of \emph{computation} (we perform more computations than the original traditional ML operators) and \emph{storage} (we create data structures that store more than what we actually need).
Although these redundancies might sound counter-intuitive at first, we are able to transform the arbitrary data accesses and control flow of the original operators into tensor operations that lead to efficient computations by leveraging state-of-the-art DNN runtimes.

For a given traditional ML operator, there exist different strategies for compiling it to tensor computations, each introducing a different degree of redundancy. We discuss such strategies for representative operators in Section~\ref{sec:compilation}.
The optimal tensor implementation to be used varies and is informed by model characteristics (e.g., tree-structure for tree-based models, or sparsity for linear models) and runtime statistics (e.g., batch size of the inputs).
\emph{Heuristics at the operator level}, \emph{runtime-independent optimizations at the pipeline level}, and \emph{runtime-specific optimizations at the execution level} enable \system to further improve predictive pipelines performance end-to-end.
The dichotomy between runtime-independent and runtime-specific optimizations allow us to both (1) apply optimizations unique to traditional ML and not captured by the DNN runtimes; and (2) exploit DNN runtime optimizations once the traditional ML is lowered into tensor computations.
Finally, \system is able to run end-to-end pipelines on the hardware platforms supported by the target DNN runtimes.

\begin{figure}[t!]
\centering
\includegraphics[clip, trim=3.5cm 11.5cm 4cm 5.8cm, width=\columnwidth]{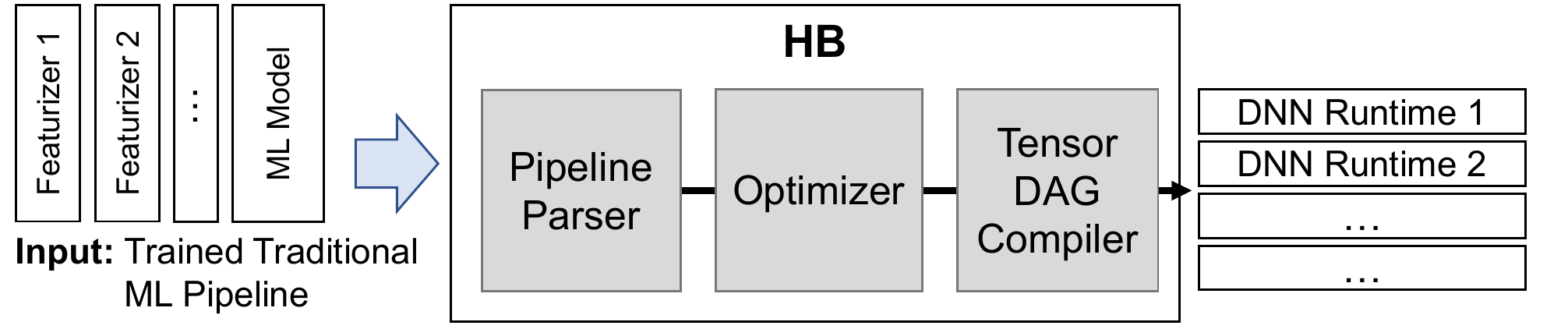}
\vspace{-3.0ex}
\caption{High-level architecture of \system.}
\label{fig:system_architecture}
\vspace{-3ex}
\end{figure}


\vspace{-2ex}
\subsection{System Architecture and Implementation}
\vspace{-1ex}
The high-level architecture of \system is shown in Figure~\ref{fig:system_architecture}.
\system has three main components: (1) \emph{Pipeline Parser}, (2) \emph{Optimizer}, and (3) \emph{Tensor DAG Compiler}.

\stitle{Pipeline Parser.}
In this phase, input pipelines are parsed one operator at a time, and each operator is \emph{wrapped} into a \emph{container} object.
Each operator's container maintains (1)~the inputs and outputs of the operator, and (2) the \emph{operator signature} that codifies the operator type (e.g., ``scikit-learn decision tree'').
\system parser also introduces a set of \emph{extractor functions} that are used to extract the parameters of each operator (e.g., weights of a linear regression, thresholds of a decision tree). 
Operator signatures dictate which extractor function should be used for each operator.
At startup time, extractor functions are registered into a hash table, mapping operator signatures to the related extractor function. 
\system parser is extensible, allowing users to easily add new extractor functions.
\system currently supports over 40 scikit-learn operators (listed in Table~\ref{tab:scikit_operators}), as well as parsers for XGBoost~\cite{xgboost}, LightGBM~\cite{lgbm}, and ONNX-ML~\cite{onnx-ml}.
At the end of the parsing phase, the input pipeline is ``logically'' represented in \system~as a DAG of containers storing all the information required for the successive phases.
\system parser is based on skl2onnx~\cite{skl2onnx}.




\stitle{Optimizer.}
In this phase, the DAG of containers generated in the parsing phase is traversed in topological order in two passes. During the first traversal pass, the Optimizer extracts the parameters of each operator via the referenced extractor function and stores them in the container.
Furthermore, since \system supports different operator implementations based on the extracted parameters, the Optimizer annotates the container with the compilation strategy to be used for that specific operator (\ref{sec:heuristics}).
During the second pass, \system tries to apply runtime-independent optimizations (\ref{sec:runtime-independent}) over the DAG. 

\stitle{Tensor DAG Compiler.}
In this last phase, the DAG of containers is again traversed in topological order and a \emph{conversion-to-tensors function} is triggered based on each operator signatures. Each conversion function receives as input the extracted parameters and generates a PyTorch's \emph{neural network module} composed of a small set of tensor operators (listed in Table~\ref{tab:tensor_operators}). The generated module is then exported into the target runtime format. The current version of \system supports PyTorch/TorchScript, ONNX, and TVM output formats.
The runtime-specific optimizations are triggered at this level.

\setcounter{table}{1}
\begin{table}[!h]
    \vspace{-2ex}
    \renewcommand\thetable{2} 
    \centering
    \caption{PyTorch tensor operators used by the Tensor DAG Compiler.}
    \label{tab:tensor_operators}
    \vspace{-2ex}
    {
    \footnotesize
    \begin{tabular}{|p{7.4cm}|}
        \hline
        \texttt{matmul, add, mul, div, lt, le, eq, gt, ge, 
        $\&$, $|$, $\ll$, $\gg$, bitwise\_xor, gather, 
        index\_select, cat, reshape, cast, abs, pow, 
        exp, arxmax, max, sum, relu, tanh, sigmoid, logsumexp, isnan, where}\\
        \hline
    \end{tabular}
    }
    \vspace{-4mm}
\end{table}

\setcounter{table}{0}
\begin{table}[ht]
    \centering
    \caption{Scikit-learn operators currently supported in \system.}
    \label{tab:scikit_operators}
    \vspace{-1ex}
    {
    \notsotiny
    \begin{tabular}{p{8cm}}
         \toprule
         {\bf Supported ML Models}\\
         \midrule
         LogisticRegression, SVC, NuSVC, LinearSVC, SGDClassifier, LogisticRegressionCV, DecisionTreeClassifier/Regression, RandomForestClassifier/Regression, ExtraTreesClassifier/Regressor, GradientBoostingClassifier/Regression,  HistGradientBoostingClassifier/Regressor, IsoltationForest, MLPClassifier, BernoulliNB, GaussianNB, MultinomialNB\\
         \midrule
                  {\bf Supported Featurizers}\\
         \midrule
          SelectKBest, VarianceThreshold, SelectPercentile, PCA, KernelPCA, TruncatedSVD, FastICA, SimpleImputer, Imputer, MissingIndicator, RobustScaler, MaxAbsScaler, MinMaxScaler, StandardScaler, Binarizer, KBinsDiscretizer, Normalizer, PolynomialFeatures, OneHotEncoder, LabelEncoder, FeatureHasher\\
         \bottomrule
    \end{tabular}
    }
    \vspace{-2ex}
\end{table}
\setcounter{table}{2}

\subsection{Assumptions and Limitations}
\vspace{-1.5ex}
In this paper, we make a few simplifying assumptions. 
First, we assume that predictive pipelines are ``pure'', i.e., they do not contain arbitrary user-defined operators.
There has been recent work~\cite{froid} on compiling imperative UDFs (user-defined functions) into relational algebra, and we plan to make use of such techniques in \system in the future.
Second, we do not support sparse data well.
We found that current support for sparse computations on DNN runtimes is primitive and not well optimized. 
We expect advances in DNN frameworks to improve on this aspect---TACO~\cite{taco} is a notable such example.
Third, although we support string operators, we currently do not support text feature extraction (e.g., \texttt{TfidfVectorizer}).
The problem in this case is twofold: (1)~compiling regex-based tokenizers into tensor computations is not trivial, and (2)~representing arbitrarily long text documents in tensors is still an open challenge.
Finally, \system is currently limited by single GPU memory execution. Given that several DNN runtimes nowadays support distributed processing~\cite{horovod,distributed-pytorch}, we plan to investigate distributed inference as future work.

\vspace{-2mm}
\vspace{-0.5ex}
\section{Compilation}\label{sec:compilation}
\vspace{-2ex}
\system supports compiling several algorithmic operators into tensor computations.
Given their popularity~\cite{dsonds}, in Section~\ref{ss:tree_compilation}  we explain our approach for tree-based models. 
Section~\ref{ss:other_techniques} gives a summary of other techniques that we use for both algorithmic and arithmetic operators.

\eat{
\begin{table}[ht]
\caption{Notation used in Section~\ref{ss:tree_compilation}}
\vspace{-2ex}
\label{table:gemm_notation}
{
\footnotesize
\begin{tabular}{c p{5.9cm}}
\toprule
Symbol & Description \\
\midrule
\midrule
$N, I, L, F, C$ & Ordered lists with all nodes, internal nodes, leaf nodes, features, and classes, respectively.\\
\midrule
$X \in \mathbb{R}^{n\times |F|}$ & Input records ($n$ is the number of records).\\
\midrule
$A \in \mathbb{R}^{|F|\times |I|}$ & 
    $A_{i,j}=
        \begin{cases}
            1, ~\text{$I_j$ evaluates  $F_i$}\\
            0, ~\text{Otherwise}\\
        \end{cases}
    $\\
\midrule
$B \in \mathbb{R}^{|I|}$ & $B_{i} =$ \texttt{ThresholdValue}($I_i$)\\
\midrule
$C \in \mathbb{R}^{|I|\times |L|}$ & 
    $C_{i,j}=
        \begin{cases}
            -1, ~\text{$L_j$ $\in$ \texttt{\small{RightSubTree}}($I_i$)}\\
            \hspace{3mm}1, ~\text{$L_j$ $\in$ \texttt{\small{LeftSubTree}}($I_i$)}\\
            \hspace{3mm}0, ~\text{Otherwise}\\
        \end{cases}
    $\\
\midrule
$D \in \mathbb{R}^{|L|}$ & $D_{k}=\hspace{-4mm}\sum\limits_{k \in L \xrightarrow{path} \texttt{Root}} \hspace{-4mm}\textbf{1}(k == \texttt{\small{LeftChild}}(\texttt{\small{Parent}}(k)))$ \\  
\midrule
$E \in \mathbb{R}^{|L|\times |C|}$ & 
    $E_{i,j}=
        \begin{cases}
            1, ~L_i\xrightarrow{map~to} C_j\\
            0, ~\text{Otherwise}\\
        \end{cases}
    $\\
\bottomrule
\end{tabular}
}
\vspace{-3mm}
\end{table}
}


\eat{
\begin{table}[ht]
    \centering
    \begin{tabular}{p{2.67cm} p{2.3cm}p{2.3cm}}
         \toprule
         Strategy &  Memory & Runtime\\
         \hline
         \hline
         \multirow{2}{3}{\texttt{GEMM}} & $O(|F||N|+$ &$O(|F||N| +$ &
         &$|N|^2 + |C||N|)$ &  $|N|^2 + |C||N|)$\\
         \hline
         \at{TreeTraversal} & $O(|N|)$ & $O(|N|)$ \\
         \hline
         \at{PerfectTreeTraversal} & $O(2^{|N|})$ & $O(|N|)$\\
         \bottomrule
    \end{tabular}
    \vspace{-2mm}
    \caption{Worst-case memory and runtime analysis of different tree compilation strategies, assuming the number of input records and number of trees are fixed. The notation is explained in Table~\ref{table:gemm_notation}. 
    }
    \label{tab:runtime_memory}
    \vspace{-0ex}
\end{table}
}

\eat{
\begin{figure*}[t]
\centering
\begin{subfigure}{0.3\textwidth}
    \centering
	\includegraphics[width=1.0\textwidth]{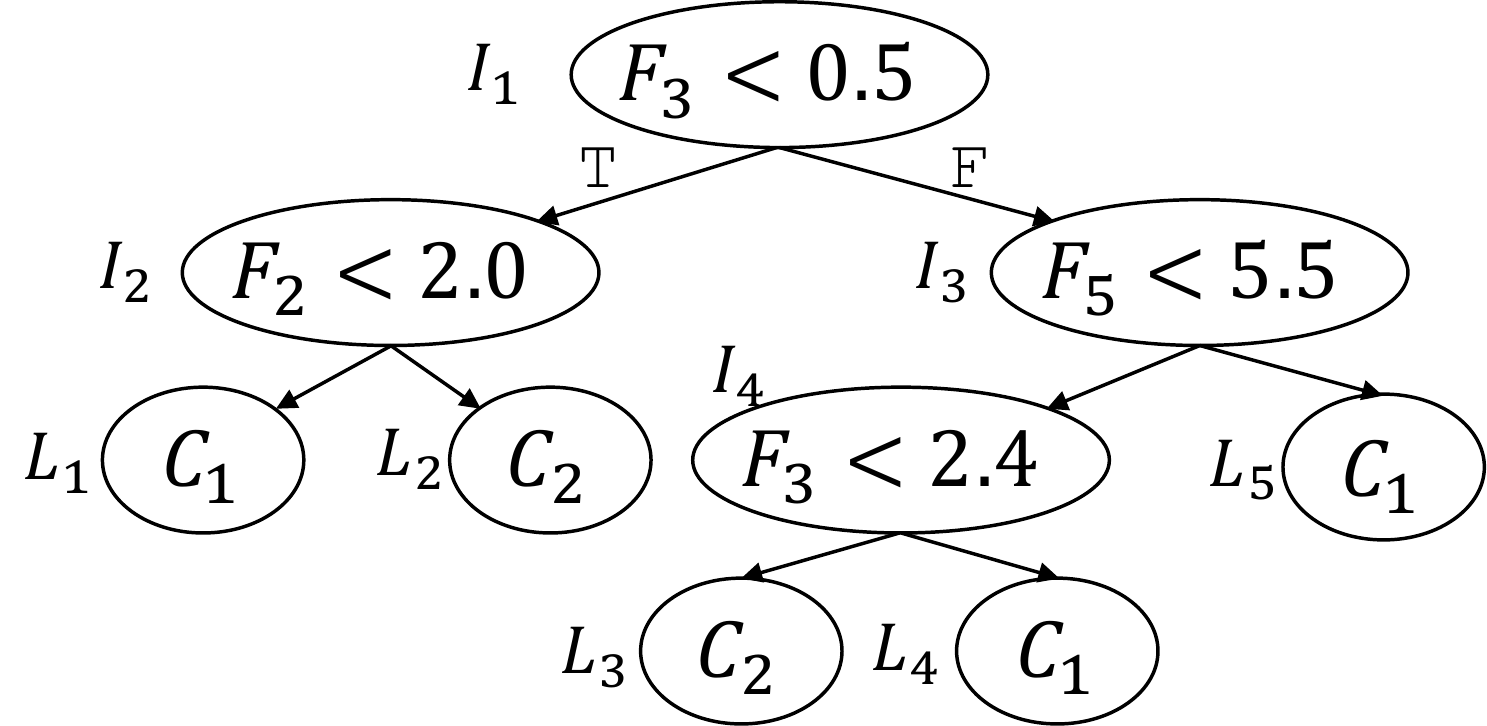}
	\centering
\end{subfigure}
\hspace{0.01\textwidth}
\begin{subfigure}{0.68\textwidth}
    \centering
	\includegraphics[width=1.0\textwidth]{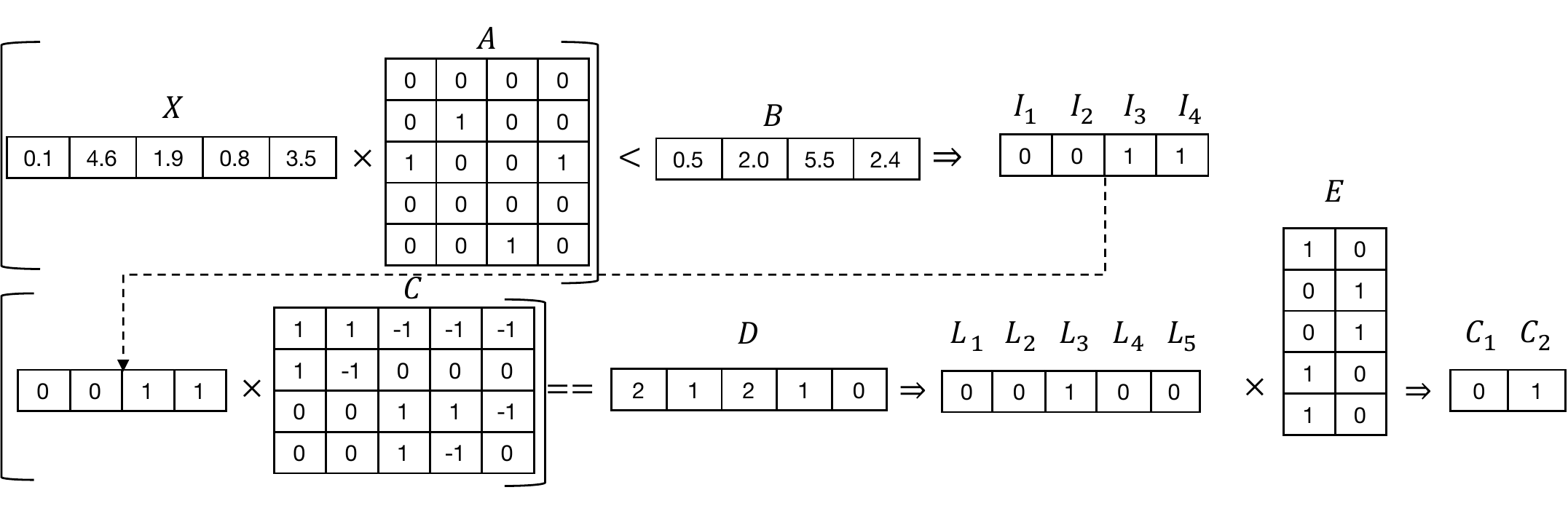}
	\centering
\end{subfigure}
\vspace{-3ex}
\caption{Compiling an example decision tree using the \texttt{GEMM} strategy.}
\label{fig:gemm}
\vspace{-4ex}
\end{figure*}
}

\vspace{-2ex}
\subsection{Compiling Tree-based Models}\label{ss:tree_compilation}
\vspace{-1.5ex}
\system has three different strategies for compiling tree-based models. Strategies differ based on the degree of redundancy introduced.
Table~\ref{table:gemm_notation} explains the notation used in this section.
We summarize the worst-case runtime and memory footprints of each strategy in Table~\ref{tab:runtime_memory}.
\system currently supports only trees built over numerical values: support for missing and categorical values is under development.
For the sake of presentation, we assume all decision nodes perform $<$ comparisons.

\begin{table}[ht]
\vspace{-1ex}
\caption{Notation used in Section~\ref{ss:tree_compilation}}
\vspace{-2ex}
\centering
\label{table:gemm_notation}
{
\footnotesize
\begin{tabular}{c p{5.9cm}}
\toprule
Symbol & Description \\
\midrule
\midrule
$N, I, L, F, C$ & Ordered lists with all nodes, internal nodes, leaf nodes, features, and classes, respectively.\\
\midrule
$X \in \mathbb{R}^{n\times |F|}$ & Input records ($n$ is the number of records).\\
\midrule
$A \in \mathbb{R}^{|F|\times |I|}$ & 
    $A_{i,j}=
        \begin{cases}
            1, ~\text{$I_j$ evaluates  $F_i$}\\
            0, ~\text{Otherwise}\\
        \end{cases}
    $\\
\midrule
$B \in \mathbb{R}^{|I|}$ & $B_{i} =$ \texttt{ThresholdValue}($I_i$)\\
\midrule
$C \in \mathbb{R}^{|I|\times |L|}$ & 
    $C_{i,j}=
        \begin{cases}
            -1, ~\text{$L_j$ $\in$ \texttt{\small{RightSubTree}}($I_i$)}\\
            \hspace{3mm}1, ~\text{$L_j$ $\in$ \texttt{\small{LeftSubTree}}($I_i$)}\\
            \hspace{3mm}0, ~\text{Otherwise}\\
        \end{cases}
    $\\
\midrule
$D \in \mathbb{R}^{|L|}$ & $D_{k}=\hspace{-4mm}\sum\limits_{k \in L \xrightarrow{path} \texttt{Root}} \hspace{-4mm}\textbf{1}(k == \texttt{\small{LeftChild}}(\texttt{\small{Parent}}(k)))$ \\  
\midrule
$E \in \mathbb{R}^{|L|\times |C|}$ & 
    $E_{i,j}=
        \begin{cases}
            1, ~L_i\xrightarrow{map~to} C_j\\
            0, ~\text{Otherwise}\\
        \end{cases}
    $\\
\bottomrule
\end{tabular}
}
\vspace{-4ex}
\end{table}


\begin{table}[ht]
    \centering
    \footnotesize
    \vspace{2ex}
    \caption{Worst-case memory and runtime analysis of different tree compilation strategies, assuming the number of input records and number of trees are fixed. The notation is explained in Table~\ref{table:gemm_notation}. 
    }
    \vspace{-2ex}
    \label{tab:runtime_memory}
    \begin{tabular}{p{1.2cm} p{3cm}p{3cm}}
         \toprule
         Strategy &  Memory & Runtime\\
         \hline
         \hline
         \texttt{GEMM} & $O(|F||N|+|N|^2 + |C||N|)$ & $O(|F||N| +|N|^2 + |C||N|)$\\
         \hline
         \at{TT} & $O(|N|)$ & $O(|N|)$ \\
         \hline
         \at{PTT} & $O(2^{|N|})$ & $O(|N|)$\\
         \bottomrule
    \end{tabular}

    \vspace{1ex}
\end{table}

\begin{figure*}[t]
\centering
\begin{subfigure}{0.3\textwidth}
    \centering
	\includegraphics[width=1.0\textwidth]{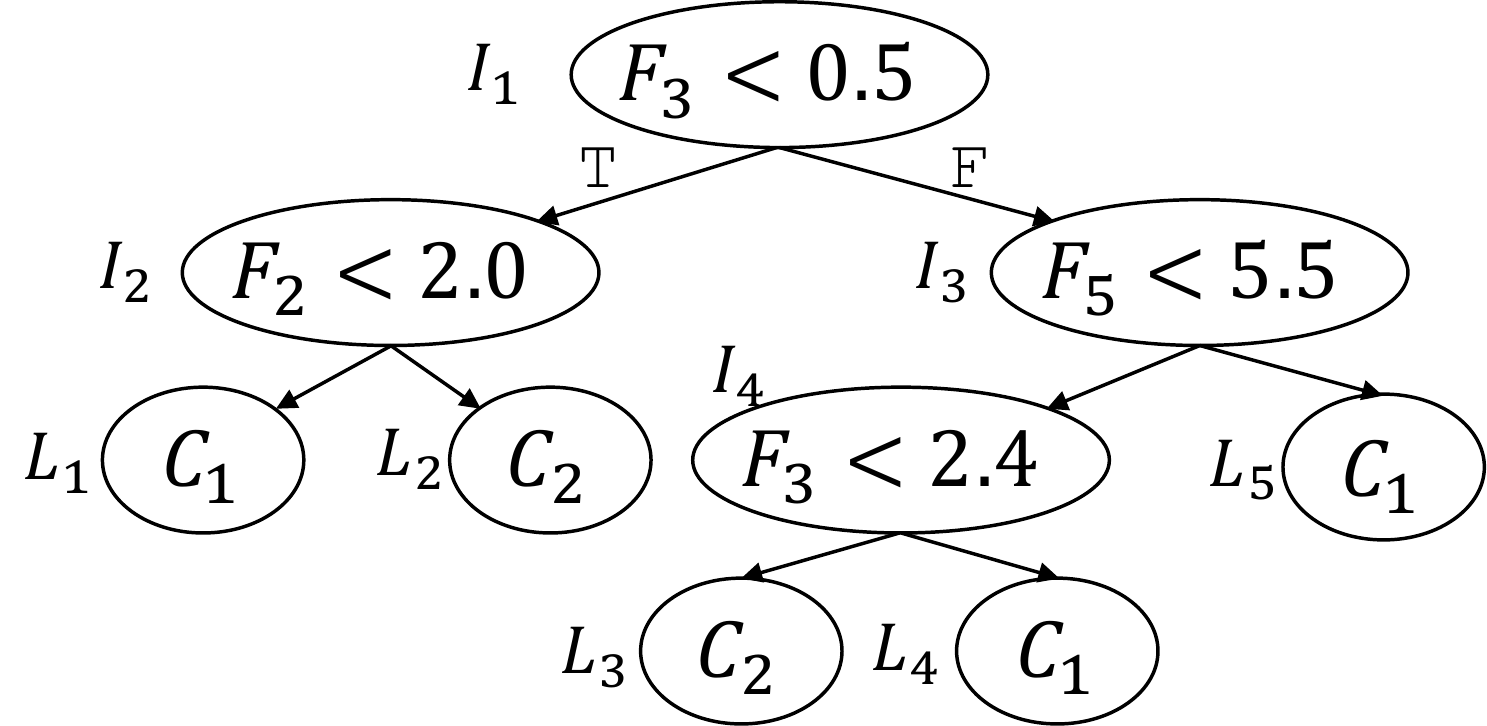}
	\centering
\end{subfigure}
\hspace{0.01\textwidth}
\begin{subfigure}{0.68\textwidth}
    \centering
	\includegraphics[width=0.91\textwidth]{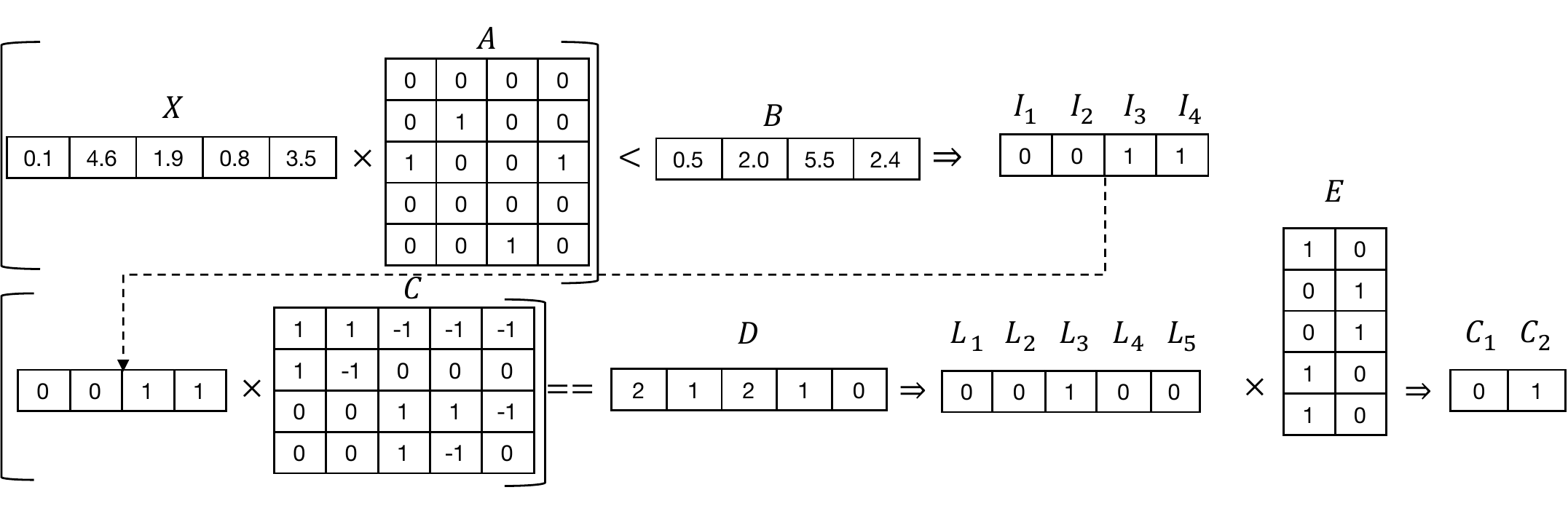}
	\centering
\end{subfigure}
\vspace{-2ex}
\caption{Compiling an example decision tree using the \texttt{GEMM} strategy.}
\label{fig:gemm}
\vspace{-2ex}
\end{figure*}

\stitle{Strategy 1: \texttt{GEMM}.}
We cast the evaluation of a tree as a series of three GEneric Matrix Multiplication (\texttt{GEMM}) operations interleaved by two element-wise logical operations.
Given a tree, we create five tensors which collectively capture the tree structure: $A, B, C, D,$ and $E$.
$A$ captures the relationship between input features and internal nodes.
$B$ is set to the threshold value of each internal node.
For any leaf node and internal node pair, $C$ captures whether the internal node is a parent of that internal node, and if so, whether it is in the left or right sub-tree.
$D$ captures the count of the internal nodes in the path from a leaf node to the tree root, for which the internal node is the left child of its parent.
Finally, $E$ captures the mapping between leaf nodes and the class labels.
Given these tensors, Algorithm~\ref{alg:gemm} presents how we perform tree scoring for a batch of input records $X$. A graphical representation of an execution of the \texttt{GEMM} strategy is depicted in Figure~\ref{fig:gemm}.

\setlength{\textfloatsep}{10pt}
\begin{algorithm}[ht]
{
\footnotesize
\DontPrintSemicolon
\SetAlgoLined
\SetKwInOut{Input}{Input}
\SetKwInOut{Output}{Output}
\Input{$X\in \mathbb{R}^{n\times |F|}$, Input records}
\Output{$R\in \{0,1\}^{n\times |C|}$, Predicted class labels}
\vspace{1mm}
\tcc{Evaluate all internal nodes}
$T \gets $ \texttt{GEMM}(X, A) \tcp*{$T \in \mathbb{R}^{n \times |I|}$}
$T \gets T < B$ \tcp*{$T \in \mathbb{R}^{n \times |I|}$}

\vspace{2mm}
\tcc{Find the leaf node which gets selected}
$T \gets $ \texttt{GEMM}(T, C) \tcp*{$T \in \mathbb{R}^{n \times |L|}$}
$T \gets T == D$ \tcp*{$T \in \mathbb{R}^{n \times |L|}$}

\vspace{2mm}
\tcc{Map selected leaf node to class label}
$R \gets $ \texttt{GEMM}(T, E) \tcp*{$R \in \mathbb{R}^{n \times |C|}$}
}
\caption{\texttt{GEMM} Strategy (Notation explained in Table~\ref{table:gemm_notation})}
\label{alg:gemm}
\end{algorithm}

The first \texttt{GEMM} is used to match each input feature with the internal node(s) using it. The following $<$ operations is used to evaluate all the internal decision nodes and produces a tensor of 0s and 1s based on the false/true outcome of the conditions.
The second \texttt{GEMM} operation generates an encoding for the path composed by the true internal nodes, while the successive $==$ operation returns the leaf node selected by the encoded path.
Finally, the third \texttt{GEMM} operation maps the selected leaf node to the class label.

This strategy can be easily applied to support tree ensembles and regression tasks too.
For tree ensembles, we create the above 2-dimensional tensors for each tree and batch them together.
As the number of leaf nodes and internal nodes can vary among trees, we pick the maximum number of leaf nodes and internal nodes for any tree as the tensor dimensions and pad the smaller tensor slices with zeros.
During scoring, we invoke the batched variants of \texttt{GEMM} and logical operations and perform a final \texttt{ReduceMean} operation over the batched dimension to generate the ensemble output.
For regression tasks, we initialize $E$ with label values.

\begin{table}[ht]
\caption{Additional notation used in Strategy 2: \at{TreeTraversal}}
\vspace{-2ex}
\label{table:tree_trav_notation}
\centering
{
\footnotesize
\begin{tabular}{c p{5.6cm}}
\toprule
Symbol & Description \\
\midrule
\midrule
$N_L\in \mathbb{Z}^{|N|}$ & 
    $N_{L_i}=
        \begin{cases}
            \texttt{\small{LeftChild}}(N_i), N_i \in I\\
            i, \text{Otherwise}\\
        \end{cases}
    $\\
\midrule
$N_R\in \mathbb{Z}^{|N|}$ & 
    $N_{R_i}=
        \begin{cases}
            \texttt{\small{RightChild}}(N_i), N_i \in I\\
            i, \text{Otherwise}\\
        \end{cases}
    $\\
\midrule
$N_F\in \mathbb{Z}^{|N|}$ &
    $N_{F_i}=
        \begin{cases}
            k, (N_i \in I) \wedge (N_i ~\text{evaluates}~ F_k)\\
            1, \text{Otherwise}\\
        \end{cases}
    $\\
\midrule
$N_T\in \mathbb{R}^{|N|}$ &
    $N_{T_i}=
        \begin{cases}
            \texttt{\small{ThresholdValue}}(N_i), N_i \in I\\
            0, \text{Otherwise}\\
        \end{cases}
    $\\
\midrule
$N_C\in \mathbb{Z}^{|N|\times |C|}$ &
    $N_{C_{i,k}}\hspace{-1mm}=\hspace{-1mm}
        \begin{cases}
            1, (N_i \in L) \wedge (N_i \text{
                    $\xrightarrow{map~to}$
                } C_k)\\
            0, \text{Otherwise}\\
        \end{cases}
    $\\
\bottomrule
\end{tabular}
}
\vspace{-1ex}
\end{table}


\stitle{Strategy 2: TreeTraversal (TT).} In the \texttt{GEMM} strategy, we incorporated a high degree of computational redundancy by evaluating all internal nodes and leaf nodes. 
Here, we try to reduce the computational redundancy by mimicking the typical tree traversal---but implemented using tensor operations.
In this strategy, the tree structure is captured by five tensors: $N_L, N_R, N_F, N_T,$ and $N_C$.
We formally define these tensors in Table~\ref{table:tree_trav_notation}.
The same column index (last dimension) across all tensors corresponds to the same tree node.
$N_L$ and $N_R$ capture the indices of the left and right nodes for a given node.
If the node is a leaf node, we set these to the index of the given node.
Similarly, $N_F$ and $N_T$ capture the feature index and threshold value for each node, respectively. For leaf nodes, we set $N_F$ to 1 and $N_T$ to 0.
Finally, $N_C$ captures the class label of each leaf node.
For internal nodes this can be any value; we set it to 0.

\begin{algorithm}[t]
{
\footnotesize
\DontPrintSemicolon
\SetAlgoLined
\SetArgSty{textup}
\SetKwInOut{Input}{Input}
\SetKwInOut{Output}{Output}
\Input{$X\in \mathbb{R}^{n\times |F|}$, Input records}
\Output{$R\in \{0,1\}^{n\times |C|}$, Predicted class labels}
\vspace{1mm}
\tcc{Initialize all records to point to $k$, with $k$ the index of \texttt{Root} node.}
$T_I \gets \{k\}^n$ \tcp*{$T_I \in \mathbb{Z}^{n}$}

\vspace{1mm}
\For{$i \gets 1$ \KwTo \texttt{TREE\_DEPTH}}{
\tcc{Find the index of the feature evaluated by the current node. Then find its value.}
$T_F \gets $\texttt{Gather}$(N_F, T_I)$ \tcp*{$T_F \in \mathbb{Z}^{n}$}
$T_V \gets $\texttt{Gather}$(X, T_f)$ \tcp*{$T_V \in \mathbb{R}^{n}$}

\vspace{1mm}
\tcc{Find the threshold, left child and right child}
$T_T \gets $\texttt{Gather}$(N_T, T_I)$ \tcp*{$T_T \in \mathbb{R}^{n}$}
$T_L \gets $\texttt{Gather}$(N_L, T_I)$ \tcp*{$T_L \in \mathbb{Z}^{n}$}
$T_R \gets $\texttt{Gather}$(N_R, T_I)$ \tcp*{$T_R \in \mathbb{Z}^{n}$}

\vspace{1mm}
\tcc{Perform logical evaluation. If true pick from $T_L$; else from $T_R$.}
$T_I \gets $\texttt{Where}$(T_V < T_T, T_L, T_R)$ \tcp*{$I \in \mathbb{Z}^{n}$}
}
\tcc{Find label for each leaf node}
$R \gets $\texttt{Gather}$(N_C, T_I)$ \tcp*{$R \in \mathbb{Z}^{n}$}
}
\caption{\at{TreeTraversal} Strategy (Notation in Tables~\ref{table:tree_trav_notation})}
\label{alg:tree_trav}
\end{algorithm}

Given these tensors, Algorithm~\ref{alg:tree_trav} presents how we perform scoring for a batch of input records $X$.
We use \texttt{Gather} and \texttt{Where} operations which can be used to perform index-based slicing and conditional value selection.
We first initialize an index tensor $T_I$ corresponding to all records in $X$, which points to the root node.
Using $T_I$, we \texttt{Gather} the corresponding feature indices and use them to \texttt{Gather} the corresponding feature values from $X$.
Similarly, we also \texttt{Gather} left node indices, right node indices, and node thresholds.
Using these gathered tensors, we then invoke a \texttt{Where} operation which checks for the tree node decisions. 
Based on the evaluation, for each record the \texttt{Where} operator either returns the left child index or right child index.
To perform full tree scoring, the above steps have to be repeated until we reach a leaf node for all records in $X$.
We exploit the fact that (1) \texttt{TREE\_DEPTH} is a known property of the input model at compilation time, and (2) all leaf nodes are at a depth $\le$ \texttt{TREE\_DEPTH}, to iterate for that fixed number of iterations to ensure that all records have found their corresponding leaf node. 
Tensors are created in such a way that if one of the indices reaches a leaf node before running for \texttt{TREE\_DEPTH} iterations, the same class label will keep getting selected.
At compile time, we unroll all iterations and remove the \texttt{for} loop to improve efficiency. For ensembles, we create tensors for each tree and batch them together.
However, between trees the number of nodes and dimensions may differ, 
so we use the maximum node count for any tree as the dimension and pad the remaining elements.


\stitle{Strategy 3: PerfectTreeTraversal (PTT).} Similar to the previous one, this strategy also mimics the tree traversal.
However, here we assume the tree is a \textit{perfect binary tree}.
In a perfect binary tree, all internal nodes have exactly two children and all leaf nodes are at the same depth level.
Assume we are given a non-perfect binary tree with a \texttt{TREE\_DEPTH} of $D$, and  $L_k$ is a leaf node which is at a depth of $D_k < D$.
To push $L_k$ to a depth $D$, we replace $L_k$ with a perfect sub-tree of depth $D-D_k$ and map all the leaf nodes of the sub-tree to $C_k$: the label of the original leaf node.
The decision nodes in the introduced sub-tree are free to perform arbitrary comparisons as the outcome is the same along any path.
By pushing all leaf nodes at depth $< D$ to a depth of $D$, we transform the original tree to a perfect tree with the same functionality.

\begin{table}[ht]
\vspace{-1ex}
\caption{Additional notation used in Strategy 3}
\vspace{-2ex}
\label{table:perfect_tree_trav_notation}
\centering
\scalebox{0.9}{
\begin{tabular}{c p{4.8cm}}
\toprule
Symbol & Description \\
\midrule
\midrule
$I'\in \mathbb{Z}^{2^{D-1}}, L'\in \mathbb{Z}^{2^D}$ & Internal and leaf nodes of the perfect tree ordered by level.\\
\midrule
$N'_F\in \mathbb{Z}^{|I'|}$ & $N'_{F_i} = k \iff I'_i$ evaluates $F_k$\\
\midrule
$N'_T\in \mathbb{R}^{|I'|}$ & $N'_{T_i} =$ \texttt{ThresholdValue}($I'_i$)\\
\midrule
$N'_C\in \mathbb{Z}^{|L'|\times |C|}$ &
        $N'_{C_{i,k}}=
        \begin{cases}
            1, N_i \text{$\xrightarrow{map~to}$} C_k\\
            0, \text{Otherwise}\\
        \end{cases}
    $\\
\bottomrule
\end{tabular}
}
\vspace{-1ex}
\end{table}

Working on perfect trees enables us to get rid of $N_L$ and $N_R$ tensors as we can now calculate them analytically, which also reduces memory lookup overheads during scoring.
Thus we create only three tensors to capture the tree structure: $N'_F, N'_T$, and $N'_C$ (Table~\ref{table:perfect_tree_trav_notation}).
They capture the same information as $N_F, N_T, N_C$ but have different dimensions and have a strict condition on the node order.
Both $N'_F$ and $N'_T$ have $2^{D-1}$ elements and the values correspond to internal nodes generated by level order tree traversal.
$N'_C$ has $2^{D}$ elements with each corresponding to an actual leaf node from left to right order.

Given these tensors, in Algorithm~\ref{alg:perfect_tree_trav} we present how \at{PTT} works.
From a high-level point of view, it is very similar to the \at{TT} strategy with only a few changes.
First, the index tensor $T_I$ is initialized to all ones as the root node is always the first node.
Second, we get rid of finding the left index and right index of a node and using them in the \texttt{Where} operation.
Instead, the \texttt{Where} operation returns $0$ for true case and $1$ for the false case.
By adding this to $2\times T_I$ we get the index of the child for the next iteration.
For ensembles, we use the maximum \texttt{TREE\_DEPTH} of any tree as $D$ for transforming trees to perfect trees.
We create tensors separate for each tree and batch them together for $N'_C$.
But for $N'_F$ and $N'_T$ instead of batching, we interleave them together in some order such that values corresponding to level $i$ for all trees appear before values corresponding to level $i+1$ of any tree.

\setlength{\textfloatsep}{0ex}
\begin{algorithm}[t]
\footnotesize
\DontPrintSemicolon
\SetAlgoLined
\SetArgSty{textup}
\SetKwInOut{Input}{Input}
\SetKwInOut{Output}{Output}
\Input{$X\in \mathbb{R}^{n\times |F|}$, Input records}
\Output{$R\in \{0,1\}^{n\times |C|}$, Predicted class labels}
\vspace{1mm}
\tcc{Initialize all records to point to the root node.}
$T_I \gets \{1\}^n$ \tcp*{$T_I \in \mathbb{Z}^{n}$}

\vspace{1mm}
\For{$i \gets 1$ \KwTo \texttt{TREE\_DEPTH}}{
\tcc{Find the index of the feature evaluated by the current node. Then find its value.}
$T_F \gets $\texttt{Gather}$(N_F, T_I)$ \tcp*{$T_F \in \mathbb{Z}^{n}$}
$T_V \gets $\texttt{Gather}$(X, T_f)$ \tcp*{$T_V \in \mathbb{R}^{n}$}

\vspace{1mm}
\tcc{Find the threshold}
$T_T \gets $\texttt{Gather}$(N_T, T_I)$ \tcp*{$T_T \in \mathbb{R}^{n}$}

\vspace{1mm}
\tcc{Perform logical evaluation. If true pick left child; else right child.}
$T_I \gets 2\times T_I$ + \texttt{Where}$(T_V < T_T, 0, 1)$ \tcp*{$I \in \mathbb{Z}^{n}$}
}
\tcc{Find label for each leaf node}
$R \gets $\texttt{Gather}$(N'_C, T_I)$ \tcp*{$R \in \mathbb{Z}^{n}$}
\caption{\at{PTT} Strategy (Notation in Tables~\ref{table:perfect_tree_trav_notation})}
\label{alg:perfect_tree_trav}
\end{algorithm}

\setlength{\textfloatsep}{10pt}

\vspace{-1ex}
\subsection{Summary of Other Techniques}\label{ss:other_techniques}
\vspace{-0.5ex}
Next, we discuss the other techniques used across ML operators to efficiently compile them into tensor computations.

\stitle{Exploiting Automatic Broadcasting.} Broadcasting~\cite{broadcasting} is the process of making two tensors shape compatible for element-wise operations.
Two tensors are said to be shape compatible if each dimension pair is the same, or one of them is 1.
At execution time, tensor operations implicitly repeat the size 1 dimensions to match the size of the other tensor, without allocating memory. 
In \system, we heavily use this feature to execute some computation over multiple inputs.
For example, consider performing an one-hot encoding operation over column $X_i\in \mathbb{R}^{n}$ with a vocabulary $V\in \mathbb{Z}^m$.
In order to implement this using tensor computations, we \texttt{Reshape} $X_i$ to $[n, 1]$ and $V$ to $[1, m]$ and calculate $R=$ \texttt{Equal}($X$, $V$), $R\in \{0, 1\}^{n\times m}$.
The \texttt{Reshape} operations are for free because they only modify the metadata of the tensor.
However, this approach performs redundant comparisons as it checks the feature values from all records against all vocabulary values. 

\stitle{Minimize Operator Invocations.} Given two approaches to implement an ML operator, we found that often picking the one which invokes fewer operators outperforms the other---even if it performs extra computations.
Consider a featurizer that generates feature interactions.
Given an input $X\in \mathbb{R}^{n\times d}$, with $d=|F|$, it generates a transformed output $R\in \mathbb{R}^{n\times \frac{d\cdot (d + 1)}{2}}$ with $R_{i} = [X_{i,1}^2, ..., X_{i,d}^2, X_{i,1}X_{i,2}, ...X_{i, d-1}X_{i,d}]$.
One way to implement this operator is to compute each new feature separately by first \texttt{Gather}ing the corresponding input feature columns, perform an element-wise \texttt{Mul}tiplication, and con\texttt{Cat}enate all new features.
However, this approach requires performing $d^2+d+1$ operations and hence is highly inefficient due to high operator scheduling overheads.
Alternatively, one could implement the same operator as follows.
First, \texttt{Reshape} $X$ into $X'\in \mathbb{R}^{n\times d\times 1}$ and $X''\in \mathbb{R}^{n\times 1\times d}$.
Then perform a batched \texttt{GEMM} using these inputs, which will create $R'\in \mathbb{R}^{n\times d\times d}$.
Finally, \texttt{Reshape} $R'$ to $R''\in \mathbb{R}^{n\times d^2}$.
Notice that each row in $R''$ has all the values of the corresponding row in $R$, but in a different order.
It also has some redundant values due to commutativity of multiplication (i.e., $x_ix_j = x_jx_i$).
Hence, we perform a final \texttt{Gather} to extract the features in the required order, and generate $R$.
Compared to the previous one, this approach increases both the computation and the memory footprint roughly by a factor of two.
However, we can implement  feature interaction in just two tensor operations.

\stitle{Avoid Generating Large Intermediate Results.} Automatic broadcasting in certain cases can become extremely inefficient due to the materialization of large intermediate tensors.
Consider the Euclidean distance matrix calculation, which is popular in many ML operators (e.g., SVMs, KNN).
Given two tensors $X\in \mathbb{R}^{n\times d}$ and $Y\in \mathbb{R}^{m\times d}$, the objective is to calculate a tensor $D\in \mathbb{R}^{n\times m}$, where $D_{i,j}=||X_i - Y_j||_2^2$.
Implementing this using broadcasting requires first \text{reshap}ing $X$ to $X'\in \mathbb{R}^{n\times 1 \times d}$, $Y$ to $Y'\in \mathbb{R}^{1\times m \times d}$, calculate $(X'-Y') \in \mathbb{R}^{n\times m \times d}$, and perform a final \texttt{Sum} over the last dimension. This approach causes a size blowup by a factor of $d$ in intermediate tensors.
Alternatively, a popular trick~\cite{euclidean_distance_trick} is to use the quadratic expansion of $D_{i,j}=||X_i||_2^2 + ||Y_j||_2^2 - 2\cdot X_i Y_j^T$ and calculate the individual terms separately.
This avoids generating intermediate tensors.

\stitle{Fixed Length Restriction on String Features.} Features with strings of arbitrary lengths pose a challenge for \system.
Strings are commonly used in categorical features, and operators like one-hot encoding and feature hashing natively support strings.
To support string features, \system imposes a fixed length restriction, with the length being determined by the max size of any string in the vocabulary.
Vocabularies are generated during training and can be accessed at compile time by \system.
Fixed length strings are then encoded into an \texttt{int8}.
\vspace{-2mm}
\vspace{-1.5ex}
\section{Optimizations}
\label{sec:optimization}
\setlength{\textfloatsep}{20pt}
\vspace{-2ex}
In this section we discuss the key optimizations performed by the \system's Optimizer: heuristics for picking operator strategies (Section~\ref{sec:heuristics}) and runtime-independent optimizations (Section~\ref{sec:runtime-independent}).
Recall that our approach also leverages runtime-specific optimizations at the Tensor Compiler level.
We refer to ~\cite{torchscript,tvm} for runtime-specific optimizations.

\vspace{-1ex}
\subsection{Heuristics-based Strategy Selection}
\label{sec:heuristics}
\vspace{-1.5ex}
For a given classical ML operator, there can be more than one compilation strategy available. In the previous section we explained three such strategies for tree-based models.
In practice, no strategy consistently dominates the others, but each is preferable in different situations based on the input and model structure.
For instance, the \at{GEMM} strategy gets significantly inefficient as the size of the decision trees gets bigger because of the large number of redundant computations.
This strategy performs $O(2^D)$ ($D$ is the depth of the tree) computations whereas the original algorithmic operator needs to perform only $O(D)$ comparisons.
Nevertheless, with small batch sizes or a large number of smaller trees, this strategy can be performance-wise optimal on modern hardware, where \texttt{GEMM} operations can run efficiently.
With large batch sizes and taller trees,  \at{TT} techniques typically outperform the \at{GEMM} strategy and \at{PTT} is slightly faster than vanilla \at{TT} due to the reduced number of memory accesses.
But if the trees are too deep, we cannot implement \at{PTT} because the $O(2^D)$ memory footprint of the associated data structures will be  prohibitive.
In such cases, we resort to \at{TT}. 
The exact crossover point where \at{GEMM} strategy outperforms other strategies is determined by the characteristics of the tree model (e.g., number of trees, maximum depth of the trees), runtime statistics (e.g., batch size), and the underlying hardware (e.g., CPUs, GPUs).
For instance, from our experiments (see Figure~\ref{fig:opt-tree}) we found that the \at{GEMM} strategy performs better for shallow trees ($D\leq 3$ on CPU, $\leq10$ on GPU) or for scoring with smaller batch sizes.
For tall trees, using  \at{PTT} when $D \le 10$ give a reasonable trade-off between memory footprint and runtime, which leaves vanilla \at{TreeTraversal} the only option for very tall trees ($D>10$).
These heuristics are currently hard-coded. 

\vspace{-2ex}
\subsection{Runtime-independent Optimizations}
\label{sec:runtime-independent}
\vspace{-1ex}
We discuss two novel optimizations, which are unique to \system. 
\system's approach of separating the prediction pipeline from training pipeline, and representing them in a logical DAG before compilation into tensor computations facilitate the optimization of end-to-end pipelines.

\stitle{Feature Selection Push-Down.} Feature selection is a popular operation that is often used as the \textit{final featurization step} as it reduces over-fitting and improves the accuracy of the ML model~\cite{feature_selection}. 
However, during scoring, it can be pushed down in the pipeline to avoid redundant computations such as scaling and one-hot encoding for discarded features or even reading the feature at all.
This idea is similar to the concept of projection push-down in relation query processing but through user-defined table functions, which in our case are the ML operators.
For operators such as feature scaling, which performs 1-to-1 feature transformations, selection push-down can be easily implemented.
However, for operators such as one-hot encoding and polynomial featurization, which perform 1-to-m or m-to-1 feature transformations, the operator will have to absorb the feature selection and stop generating those features.
For example, say one-hot encoding is applied on a categorical feature column which has a vocabulary size of 10, but 4 of those features are discarded by the feature selector. In such cases, we can remove such features from the vocabulary. 
Note that for some ``blocking'' operators~\cite{pretzel}, such as normalizers, it is not possible to push-down the feature selection.

\stitle{Feature Selection Injection.} Even if the original pipeline doesn't have a feature selection operator, it is possible to inject one and then push it down. Linear models with L1 regularization (Lasso) is a typical example where feature selection is implicitly performed. 
The same idea can be extended to tree-based models to prune the features that are not used as decision variables.
In both of these examples, the ML model also has to be updated to take into account the pruned features. For linear models we prune the zero weights; for tree models, we update the indices of the decision variables.

\eat{
\stitle{Algebraic Rewrites.} We found opportunities to rewrite several operators that perform linear algebra operations into a single \texttt{GEMM} operation.
Consider a pipeline that trains a logistic regression model and has feature scaling and matrix decomposition (e.g., PCA) as the featurization steps.
It is algebraically represented in Eq.~\ref{eq:algebraic_rewrite}-LHS. 

\vspace{-2mm}
\begin{multline}\label{eq:algebraic_rewrite}
    \texttt{sigmoid}\bigg(\Big(\big(\frac{X - \alpha}{\beta}\big) \cdot W_{PCA} \Big) \cdot W_{LR} + B_{LR} \bigg) \\
    = \texttt{sigmoid}(X\cdot W + B)
\end{multline}

Notice that parentheses in Eq.~\ref{eq:algebraic_rewrite}-LHS capture the order in which the operators were trained and it requires performing 5 tensor operations: 2 element-wise ops for scaling; two \texttt{GEMM} ops for matrix decomposition and logistic regression; and a final \texttt{sigmoid} op for logistic regression.
It is possible to use linear algebra properties and represent the same pipeline using two ops as shown in RHS, where tensor $W$ and $B$ can be pre-computed and used during scoring.
These kinds of patterns are in fact very common in classical ML; any subset of scaling, matrix decomposition, and linear models constitutes such patterns.
However, they do not appear in DNNs due to the use of non-linear transformations and hence most tensor runtime optimizers are oblivious of these opportunities.
\system's optimizer has a roster of such patterns and checks for potential rewrites during optimization.

\stitle{Batching Stacked Models.} Recall that in Section~\ref{ss:tree_compilation}, we batched the tensors from all trees into single tensors and performed batched tensor operations.
Alternatively, we could have stored them separately and invoke tensor operations on each tree.
Though this would yield the same result it will be highly inefficient due to two reasons: (1) high operator invocation overhead and (2) high memory access overhead (due to multiple reading of the input tensor $X$).
It is possible to apply the same batching optimization across multiple ML operators.
Consider a stacked ML model that is composed of logistic regression, linear SVM, and Bernoulli Naive Bayes models.
While these models are conceptually different during scoring all three of them are essentially performing a \texttt{GEMM} operation.
Thus, it is possible to batch them together into one \texttt{GEMM} operation to reduce overheads.
Efficiently finding this type of graph substitutions in an arbitrary tensor computation DAG is still an active area of research~\cite{taso}; hence not supported by many DNN runtimes.
On the other hand, \system implements few patterns as the one above, directly over traditional ML operators and triggers batching rewrites.
}
\vspace{-1mm}
\vspace{-1ex}
\section{Experimental Evaluation}
\label{s:experiments}
\vspace{-1.5ex}




In our experimental evaluation 
we report two micro-benchmark experiments showing how \system performs compared to current state-of-the-art for inference over (1) tree ensembles (Section~\ref{sec:mini-trees}); (2) other featurization operators and ML models (Section~\ref{sec:mini-ops}). Then we evaluate the optimizations by showing: (1) the need for heuristics for picking the best tree-model implementation (Section \ref{sec:optimizations}); and (2) the benefits introduced by the runtime-independent optimizations (Section~\ref{sec:opt-runtime-independent}).
Finally, we conduct an end-to-end evaluation using pipelines (Section~\ref{sec:end-to-end}).
We evaluate both CPUs and hardware accelerators (GPUs).

\eat{\stitle{Key Takeaways.} Although \system targets the high-level tensor APIs provided by DNN frameworks, still it is able to outperform custom C++ and CUDA implementations, both for batch and single tuple-at-the-time scenarios. These results also hold for end-to-end pipelines where \system is able to provide up to 1200$\times$ speedup compared to scikit-learn.
}

\stitle{Hardware and Software Setup.} For all the experiments (except when stated otherwise) we use an Azure NC6 v2 machine equipped with 112 GB of RAM, an Intel Xeon CPU E5-2690 v4 @ 2.6GHz (6 virtual cores), and an NVIDIA P100 GPU. The machine runs Ubuntu 18.04 with PyTorch 1.3.1, TVM 0.6, scikit-learn 0.21.3, XGBoost 0.9, LightGBM 2.3.1, ONNX runtime 1.0, RAPIDS 0.9, and CUDA 10. We run TVM with \texttt{opt\_level} 3 when not failing; 0 otherwise. 

\stitle{Experimental Setup.}
We run all the experiments 5 times and report the truncated mean (by averaging the middle values) of the processor time.
In the following, we use ONNX-ML to indicate running an ONNX-ML model (i.e., traditional ML part of the standard) on the ONNX runtime.
Additionally, we use \textbf{bold numbers} to highlight the best performance for the specific setup (CPU or GPU). \emph{Note that both scikit-learn and ONNX-ML do not natively support hardware acceleration.}

\vspace{-2ex}
\subsection{Micro-benchmarks}
\vspace{-1ex}
\subsubsection{Tree Ensembles}
\label{sec:mini-trees}
\vspace{-1.5ex}





\stitle{Setup.} This experiment is run over a set of popular datasets used for benchmarking gradient boosting frameworks~\cite{gbm-bench}. We first do a 80\%/20\% train/test split over each dataset. Successively, we train a scikit-learn  \emph{random forest}, \emph{XGBoost}~\cite{xgboost}, and \emph{LightGBM}~\cite{lgbm} models using the default parameters of the benchmark. Specifically, we set the number of trees to 500 and maximum depth to 8.
For XGBoost and LightGBM we use the scikit-learn API. Note that each algorithm generates trees with different structures, and this experiment helps with understanding how \system behaves with various tree types and dataset scales. For example, XGBoost generates balanced trees, LightGBM mostly generates skinny tall trees, while random forest is a mix between the two.
Finally, we score the trained models over the test dataset using different batch sizes. 
We compare the results against \system with different runtime backends and an ONNX-ML version of the model generated using ONNXMLTools~\cite{onnxmltools}.
When evaluating over GPU, we also compared against NVIDIA RAPIDS Forest Inference Library (FIL)~\cite{fil}. We don't compare against GPU implementations for XGBoost or LightGBM because we consider FIL as state-of-the-art~\cite{fil-blog}.
For the CPU experiments, we use all six cores in the machine, while for request/response 
experiments we use one core. 
We set a timeout of 1 hour for each experiment. 

\eat{
\begin{table}[h]
\centering
\caption{Datasets used from the tree ensembles micro-benchmark}
\vspace{-2ex}
\label{tb:tree-data}
\begin{tabular}{|c|c|c|c|c|}
\hline
\cellcolor{black}\textcolor{white}{\sf Datasets} & \cellcolor{black}\textcolor{white}{\sf \#Rows} & \cellcolor{black}\textcolor{white}{\sf \#Columns} & \cellcolor{black}\textcolor{white}{\sf Task} \\ \hline
Fraud              & $285K$ & 28 & Binary \\ \hline
Epsilon              & $500K$ & 2000 & Binary \\ \hline
Year              & $515K$ & 90 & Regression \\ \hline
CovType              & $581K$ & 54 & Multiclass \\
\hline
Higgs              & $11M$ & 28 & Binary \\
\hline
Airline              & $115M$ & 13 & Binary \\
\hline
\end{tabular}
\vspace{-1ex}
\end{table}
}

\stitle{Datasets.} We use 6 datasets from NVIDIA's gbm-bench~\cite{gbm-bench}. 
The datasets cover a wide spectrum of use-cases: from regression to multiclass classification, from 285$K$ rows to 100$M$, and from few 10s of columns to 2$K$.

\begin{table*}[t!]
\centering
\caption{Batch Experiments ($10K$ records at-a-time) for both CPU (6 cores) and GPU. Reported numbers are in seconds.}
\vspace{-2ex}
\label{tbl:scenario-1-data-size}
\scalebox{0.95}{
\notsotiny
\tablecolumnmarginsmall
\begin{tabular}{ccclllllllc}
\toprule
\multirow{2}[3]{*}{Algorithm} & \multirow{2}[3]{*}{Dataset}  & \multicolumn{2}{c}{Baselines (CPU)}   & 
\multicolumn{3}{c}{\system CPU}  &
\multicolumn{1}{c}{Baselines (GPU)} & \multicolumn{2}{c}{\system GPU} \\
\cmidrule(lr){3-4} \cmidrule(lr){5-7} \cmidrule(lr){8-8} \cmidrule(lr){9-10}
& & \multicolumn{1}{c}{Sklearn} & \multicolumn{1}{c}{ONNX-ML} & \multicolumn{1}{c}{PyTorch} & \multicolumn{1}{c}{TorchScript} &
\multicolumn{1}{c}{TVM} & \multicolumn{1}{c}{RAPIDS FIL} & \multicolumn{1}{c}{TorchScript} & \multicolumn{1}{c}{TVM} \\ \midrule

\multirow{6}{*}{Rand. Forest} & Fraud & \multicolumn{1}{c}{\textbf{2.5}} & \multicolumn{1}{c}{7.1} & \multicolumn{1}{c}{8.0} & \multicolumn{1}{c}{7.8} &
\multicolumn{1}{c}{3.0} & \multicolumn{1}{c}{not supported} & \multicolumn{1}{c}{0.044} & \multicolumn{1}{c}{\textbf{0.015}} \\

& Epsilon & \multicolumn{1}{c}{9.8} & \multicolumn{1}{c}{18.7} & \multicolumn{1}{c}{14.7} & \multicolumn{1}{c}{13.9} &
\multicolumn{1}{c}{\textbf{6.6}} & \multicolumn{1}{c}{not supported} & \multicolumn{1}{c}{\textbf{0.13}} & \multicolumn{1}{c}{\textbf{0.13}} \\

& Year & \multicolumn{1}{c}{1.9} & \multicolumn{1}{c}{6.6} & \multicolumn{1}{c}{7.8} & \multicolumn{1}{c}{7.7} &
\multicolumn{1}{c}{\textbf{1.4}} & \multicolumn{1}{c}{not supported} & \multicolumn{1}{c}{0.045} & \multicolumn{1}{c}{\textbf{0.026}} \\

& Covtype & \multicolumn{1}{c}{\textbf{5.9}} & \multicolumn{1}{c}{18.1} & \multicolumn{1}{c}{17.22} & \multicolumn{1}{c}{16.5} &
\multicolumn{1}{c}{6.8} & \multicolumn{1}{c}{not supported} & \multicolumn{1}{c}{0.11} & \multicolumn{1}{c}{\textbf{0.047}} \\

& Higgs & \multicolumn{1}{c}{\textbf{102.4}} & \multicolumn{1}{c}{257.6} & \multicolumn{1}{c}{314.4} & \multicolumn{1}{c}{314.5} &
\multicolumn{1}{c}{118.0} & \multicolumn{1}{c}{not supported} & \multicolumn{1}{c}{1.84} & \multicolumn{1}{c}{\textbf{0.55}} \\

& Airline & \multicolumn{1}{c}{1320.1} & \multicolumn{1}{c}{timeout} & \multicolumn{1}{c}{timeout} & \multicolumn{1}{c}{timeout} &
\multicolumn{1}{c}{\textbf{1216.7}} & \multicolumn{1}{c}{not supported} & \multicolumn{1}{c}{18.83} & \multicolumn{1}{c}{\textbf{5.23}} \\ \midrule

\multirow{6}{*}{LightGBM} & Fraud & \multicolumn{1}{c}{3.4} & \multicolumn{1}{c}{5.9} & \multicolumn{1}{c}{7.9} & \multicolumn{1}{c}{7.6} &
\multicolumn{1}{c}{\textbf{1.7}} & \multicolumn{1}{c}{\textbf{0.014}} & \multicolumn{1}{c}{0.044} & \multicolumn{1}{c}{\textbf{0.014}} \\

& Epsilon & \multicolumn{1}{c}{10.5} & \multicolumn{1}{c}{18.9} & \multicolumn{1}{c}{14.9} & \multicolumn{1}{c}{14.5} &
\multicolumn{1}{c}{\textbf{4.0}} & \multicolumn{1}{c}{0.15} & \multicolumn{1}{c}{0.13} & \multicolumn{1}{c}{\textbf{0.12}} \\

& Year & \multicolumn{1}{c}{5.0} & \multicolumn{1}{c}{7.4} & \multicolumn{1}{c}{7.7} & \multicolumn{1}{c}{7.6} &
\multicolumn{1}{c}{\textbf{1.6}} & \multicolumn{1}{c}{\textbf{0.023}} & \multicolumn{1}{c}{0.045} & \multicolumn{1}{c}{0.025} \\

& Covtype & \multicolumn{1}{c}{51.06} & \multicolumn{1}{c}{126.6} & \multicolumn{1}{c}{79.5} & \multicolumn{1}{c}{79.5} &
\multicolumn{1}{c}{\textbf{27.2}} & \multicolumn{1}{c}{not supported} & \multicolumn{1}{c}{0.62} & \multicolumn{1}{c}{\textbf{0.25}} \\

& Higgs & \multicolumn{1}{c}{198.2} & \multicolumn{1}{c}{271.2} & \multicolumn{1}{c}{304.0} & \multicolumn{1}{c}{292.2} &
\multicolumn{1}{c}{\textbf{69.3}} & \multicolumn{1}{c}{0.59} & \multicolumn{1}{c}{1.72} & \multicolumn{1}{c}{\textbf{0.52}} \\

& Airline & \multicolumn{1}{c}{1696.0} & \multicolumn{1}{c}{timeout} & \multicolumn{1}{c}{timeout} & \multicolumn{1}{c}{timeout} &
\multicolumn{1}{c}{\textbf{702.4}} & \multicolumn{1}{c}{5.55} & \multicolumn{1}{c}{17.65} & \multicolumn{1}{c}{\textbf{4.83}} \\\midrule

\multirow{6}{*}{XGBoost} & Fraud & \multicolumn{1}{c}{1.9} & \multicolumn{1}{c}{5.5} & \multicolumn{1}{c}{7.7} & \multicolumn{1}{c}{7.6} &
\multicolumn{1}{c}{\textbf{1.6}} & \multicolumn{1}{c}{\textbf{0.013}} & \multicolumn{1}{c}{0.44} & \multicolumn{1}{c}{0.015} \\

& Epsilon & \multicolumn{1}{c}{7.6} & \multicolumn{1}{c}{18.9} & \multicolumn{1}{c}{14.8} & \multicolumn{1}{c}{14.8} &
\multicolumn{1}{c}{\textbf{4.2}} & \multicolumn{1}{c}{0.15} & \multicolumn{1}{c}{0.13} & \multicolumn{1}{c}{\textbf{0.12}} \\

& Year & \multicolumn{1}{c}{3.1} & \multicolumn{1}{c}{8.6} & \multicolumn{1}{c}{7.6} & \multicolumn{1}{c}{7.6} &
\multicolumn{1}{c}{\textbf{1.6}} & \multicolumn{1}{c}{\textbf{0.022}} & \multicolumn{1}{c}{0.045} & \multicolumn{1}{c}{0.026} \\

& Covtype & \multicolumn{1}{c}{42.3} & \multicolumn{1}{c}{121.7} & \multicolumn{1}{c}{79.2} & \multicolumn{1}{c}{79.0} &
\multicolumn{1}{c}{\textbf{26.4}} & \multicolumn{1}{c}{not supported} & \multicolumn{1}{c}{0.62} & \multicolumn{1}{c}{\textbf{0.25}} \\

& Higgs & \multicolumn{1}{c}{126.4} & \multicolumn{1}{c}{309.7} & \multicolumn{1}{c}{301.0} & \multicolumn{1}{c}{301.7} &
\multicolumn{1}{c}{\textbf{66.0}} & \multicolumn{1}{c}{0.59} & \multicolumn{1}{c}{1.73} & \multicolumn{1}{c}{\textbf{0.53}} \\

& Airline & \multicolumn{1}{c}{1316.0} & \multicolumn{1}{c}{timeout} & \multicolumn{1}{c}{timeout} & \multicolumn{1}{c}{timeout} &
\multicolumn{1}{c}{\textbf{663.3}} & \multicolumn{1}{c}{5.43} & \multicolumn{1}{c}{17.16} & \multicolumn{1}{c}{\textbf{4.83}}\\
\bottomrule
\end{tabular}
}
\vspace{-2.5ex}
\end{table*}

\stitle{List of Experiments.} We run the following set of experiments: (1) batch inference, both on CPU and GPU; (2) request/response where one single record is scored at a time; 
(3) scaling experiments by varying batch sizes, both over CPU and GPU; (4) evaluation on how \system behaves on different GPU generations; (5) dollar cost per prediction; (6)  memory consumption; (7) validation of the produced output wrt scikit-learn; and finally (8) time spent on compiling the models. 

\eat{
\begin{table*}[t!]

\centering
\caption{Batch Experiments ($10K$ records at-a-time) for both CPU (6 cores) and GPU. Reported numbers are in seconds.}
\label{tbl:scenario-1-data-size}
{
\notsotiny
\tablecolumnmarginsmall
\begin{tabular}{cccllllllllc}

\toprule
\multirow{2}{*}{Algorithm} & \multirow{2}{*}{Dataset}  & \multicolumn{2}{c}{Baselines (CPU)}   & \multicolumn{4}{c}{Hummingbird CPU}  & \multicolumn{1}{c}{Baselines (GPU)} & \multicolumn{2}{c}{Hummingbird GPU} \\
\cmidrule(lr){3-4} \cmidrule(lr){5-8} \cmidrule(lr){9-9} \cmidrule(lr){10-11}
& & \multicolumn{1}{c}{Sklearn} & \multicolumn{1}{c}{ONNX-ML} & \multicolumn{1}{c}{PyTorch} & \multicolumn{1}{c}{TorchScript} & \multicolumn{1}{c}{ONNX} & \multicolumn{1}{c}{TVM} & \multicolumn{1}{c}{RAPIDS FIL} & \multicolumn{1}{c}{TorchScript} & \multicolumn{1}{c}{TVM} \\ \midrule
\multirow{6}{*}{Rand. Forest} & Fraud & \multicolumn{1}{c}{\textbf{2.51}} & \multicolumn{1}{c}{8.1} & \multicolumn{1}{c}{8.59} & \multicolumn{1}{c}{8.58} & \multicolumn{1}{c}{92.58} & \multicolumn{1}{c}{3.83} & \multicolumn{1}{c}{not supported} & \multicolumn{1}{c}{0.044} & \multicolumn{1}{c}{\textbf{0.015}} \\
& Epsilon & \multicolumn{1}{c}{11.22} & \multicolumn{1}{c}{27.98} & \multicolumn{1}{c}{24.26} & \multicolumn{1}{c}{22.94} &
\multicolumn{1}{c}{error} & \multicolumn{1}{c}{\textbf{8.17}} & \multicolumn{1}{c}{not supported} & \multicolumn{1}{c}{\textbf{0.13}} & \multicolumn{1}{c}{\textbf{0.13}} \\
& Year & \multicolumn{1}{c}{2.32} & \multicolumn{1}{c}{17.22} & \multicolumn{1}{c}{8.97} & \multicolumn{1}{c}{8.35} & \multicolumn{1}{c}{154.72} & \multicolumn{1}{c}{\textbf{1.44}} & \multicolumn{1}{c}{not supported} & \multicolumn{1}{c}{0.045} & \multicolumn{1}{c}{\textbf{0.026}} \\
& Covtype & \multicolumn{1}{c}{47.63} & \multicolumn{1}{c}{24.78} & \multicolumn{1}{c}{19.13} & \multicolumn{1}{c}{\textbf{18.98}} & \multicolumn{1}{c}{260.35} & \multicolumn{1}{c}{20.19} & \multicolumn{1}{c}{not supported} & \multicolumn{1}{c}{0.11} & \multicolumn{1}{c}{\textbf{0.047}} \\
& Higgs & \multicolumn{1}{c}{776.79} & \multicolumn{1}{c}{337.55} & \multicolumn{1}{c}{339.15} & \multicolumn{1}{c}{337.8} & \multicolumn{1}{c}{3436.34} & \multicolumn{1}{c}{\textbf{152.24}} & \multicolumn{1}{c}{not supported} & \multicolumn{1}{c}{1.84} & \multicolumn{1}{c}{\textbf{0.55}} \\
& Airline & \multicolumn{1}{c}{timeout} & \multicolumn{1}{c}{timeout} & \multicolumn{1}{c}{timeout} & \multicolumn{1}{c}{timeout} & \multicolumn{1}{c}{timeout} & \multicolumn{1}{c}{\textbf{1589.34}} & \multicolumn{1}{c}{not supported} & \multicolumn{1}{c}{18.83} & \multicolumn{1}{c}{\textbf{5.23}} \\ \midrule
\multirow{6}{*}{LightGBM} & Fraud & \multicolumn{1}{c}{3.75} & \multicolumn{1}{c}{6.59} & \multicolumn{1}{c}{8.71} & \multicolumn{1}{c}{8.21} & \multicolumn{1}{c}{89.9} & \multicolumn{1}{c}{\textbf{1.97}} & \multicolumn{1}{c}{\textbf{0.014}} & \multicolumn{1}{c}{0.044} & \multicolumn{1}{c}{\textbf{0.014}} \\
& Epsilon & \multicolumn{1}{c}{14.14} & \multicolumn{1}{c}{26.03} & \multicolumn{1}{c}{23.60} & \multicolumn{1}{c}{22.95} & \multicolumn{1}{c}{error} & \multicolumn{1}{c}{\textbf{4.42}} & \multicolumn{1}{c}{0.15} & \multicolumn{1}{c}{0.13} & \multicolumn{1}{c}{\textbf{0.12}} \\
& Year & \multicolumn{1}{c}{6.19} & \multicolumn{1}{c}{10.14} & \multicolumn{1}{c}{9.15} & \multicolumn{1}{c}{9.01} & \multicolumn{1}{c}{153.66} & \multicolumn{1}{c}{\textbf{1.77}} & \multicolumn{1}{c}{\textbf{0.023}} & \multicolumn{1}{c}{0.045} & \multicolumn{1}{c}{0.025} \\
& Covtype & \multicolumn{1}{c}{67.12} & \multicolumn{1}{c}{158.3} & \multicolumn{1}{c}{146.6} & \multicolumn{1}{c}{146.8} & \multicolumn{1}{c}{1256.09} & \multicolumn{1}{c}{\textbf{29.19}} & \multicolumn{1}{c}{not supported} & \multicolumn{1}{c}{0.62} & \multicolumn{1}{c}{\textbf{0.25}} \\
& Higgs & \multicolumn{1}{c}{324} & \multicolumn{1}{c}{351.95} & \multicolumn{1}{c}{325.70} & \multicolumn{1}{c}{324.25} & \multicolumn{1}{c}{3303.76} & \multicolumn{1}{c}{\textbf{75.93}} & \multicolumn{1}{c}{0.59} & \multicolumn{1}{c}{1.72} & \multicolumn{1}{c}{\textbf{0.52}} \\
& Airline & \multicolumn{1}{c}{2047.13} & \multicolumn{1}{c}{timeout} & \multicolumn{1}{c}{timeout} & \multicolumn{1}{c}{timeout} & \multicolumn{1}{c}{timeout} & \multicolumn{1}{c}{\textbf{780.44}} & \multicolumn{1}{c}{5.55} & \multicolumn{1}{c}{17.65} & \multicolumn{1}{c}{\textbf{4.83}} \\\midrule
\multirow{6}{*}{XGBoost} & Fraud & \multicolumn{1}{c}{2.01} & \multicolumn{1}{c}{6.4} & \multicolumn{1}{c}{8.61} & \multicolumn{1}{c}{8.18} & \multicolumn{1}{c}{89.72} & \multicolumn{1}{c}{\textbf{1.94}} & \multicolumn{1}{c}{\textbf{0.013}} & \multicolumn{1}{c}{0.44} & \multicolumn{1}{c}{0.015} \\
& Epsilon & \multicolumn{1}{c}{14.85} & \multicolumn{1}{c}{29.01} & \multicolumn{1}{c}{23.69} & \multicolumn{1}{c}{24.38} &
\multicolumn{1}{c}{error} & \multicolumn{1}{c}{\textbf{4.42}} & \multicolumn{1}{c}{0.15} & \multicolumn{1}{c}{0.13} & \multicolumn{1}{c}{\textbf{0.12}} \\
& Year & \multicolumn{1}{c}{5.78} & \multicolumn{1}{c}{15.75} & \multicolumn{1}{c}{8.63} & \multicolumn{1}{c}{7.87} & \multicolumn{1}{c}{153.97} & \multicolumn{1}{c}{\textbf{1.77}} & \multicolumn{1}{c}{\textbf{0.022}} & \multicolumn{1}{c}{0.045} & \multicolumn{1}{c}{0.026} \\
& Covtype & \multicolumn{1}{c}{63.45} & \multicolumn{1}{c}{173.91} & \multicolumn{1}{c}{147.69} & \multicolumn{1}{c}{147.6} &
\multicolumn{1}{c}{1255.89} &
\multicolumn{1}{c}{\textbf{28.53}} & \multicolumn{1}{c}{not supported} & \multicolumn{1}{c}{0.62} & \multicolumn{1}{c}{\textbf{0.25}} \\
& Higgs & \multicolumn{1}{c}{\textbf{74.76}} & \multicolumn{1}{c}{293.52} & \multicolumn{1}{c}{323.94} & \multicolumn{1}{c}{322.56} & \multicolumn{1}{c}{3285.98} & \multicolumn{1}{c}{77.01} & \multicolumn{1}{c}{0.59} & \multicolumn{1}{c}{1.73} & \multicolumn{1}{c}{\textbf{0.53}} \\
& Airline & \multicolumn{1}{c}{1577.50} & \multicolumn{1}{c}{timeout} & \multicolumn{1}{c}{timeout} & \multicolumn{1}{c}{timeout} & \multicolumn{1}{c}{timeout} & \multicolumn{1}{c}{\textbf{764.61}} & \multicolumn{1}{c}{5.43} & \multicolumn{1}{c}{17.16} & \multicolumn{1}{c}{\textbf{4.83}}\\
\bottomrule
\end{tabular}
}

\end{table*}
}


\stitle{Batch Inference.}
\label{sec:tree-batch}
Table~\ref{tbl:scenario-1-data-size} reports the inference time 
for random forest, XGBoost and LightGBM models run over the 6 datasets. The batch size is set to $10K$ records. 
Looking at the CPU numbers from the table, we can see that:
\vspace{-1ex}
\begin{enumerate}
\item Among the baselines, scikit-learn models outperform ONNX-ML implementations by 2 to 3$\times$. This is because ONNX-ML v1.0 is not optimized for batch inference.\vspace{-1ex}
\item Looking at the \system's backends, there is not a large difference between PyTorch and TorchScript, and in general these backends perform comparable to ONNX-ML.\vspace{-1ex}
\item The TVM backend provides the best performance on 15 experiments out of 18. In the worst case TVM is 20\% slower (than scikit-learn); in the best cases it is up to 2$\times$ faster compared to the baseline solutions.
\end{enumerate}\vspace{-1ex}

Let us look now at the GPU numbers of Table~\ref{tbl:scenario-1-data-size}:
\vspace{-1ex}
\begin{enumerate}
    \item Baseline RAPIDS does not support random forest nor multiclass classification tasks.
    For the remaining experiments, GPU acceleration is able to provide speedups of up to 300$\times$ compared to CPU baselines.\footnote{The original FIL blog post~\cite{fil-blog} claims GPU acceleration to be in the order of 28$\times$ for XGBoost, versus close to 300$\times$ in our case (Airline). We think that the difference is in the hardware: in fact, they use 5 E5-2698 CPUs for a total of 100 physical cores, while we use a E5-2690 CPU with 6 (virtual) physical cores. Additionally, they use a V100 GPU versus a P100 in our case.}\vspace{-1ex}
    \item Looking at \system backends, TorchScript is about 2 to 3$\times$ slower compared to RAPIDS. TVM is instead the faster solution on 14 experiments out of 18, with a 10\% to 20\% improvement wrt RAPIDS. 
\end{enumerate}\vspace{-1ex}

The results are somehow surprising: \system targets the high-level tensor APIs provided by PyTorch and TVM, and still it is able to outperform custom C++ and CUDA implementations.

\eat{
\begin{itemize}
    \item \stitle{Setup.} Baselines: lightgbm, xgboost, rf, FIL. Default num trees. Backends: ONNX, PyTorch, TorchScript, TVM.
    \item \st{Batch experiments}
    \item \st{GPU experiments}
    \item \st{request/response}
    \item \st{scaling experiments}
    \item \st{conversion times. p100, 1 cpu, batch of 1. Parallel execution increase the conversion by up to 5X for some combination. Even more for GPU.}
    \item \st{memory consumption. Tested with p100, 1 cpu, batch of 1k}.
    \item \st{Same output or not. Tested with p100, 6 cpu, batch of $10K$.} 
    \item Comparison TVM vs PyTorch. For a dataset and algo show that TVM wihout operator fusion and scheduling is equivalent to PyTorch. Not sure if we can show this. Just removing operation fusion is not enough.
     \item K80, V100, P100 (IPU?). For this experiment we compared performance across different Nvidia GPU models. Figure~\ref{fig:k80p100v100} shows the performance of ...
\end{itemize}
}

\stitle{Request/response.}
\begin{figure}  
\vspace{-1ex}
	\centering
	\begin{subfigure}{0.5\textwidth}
		\centering
		\includegraphics[trim={0 0 0 0.5cm},clip,width=0.7\textwidth]{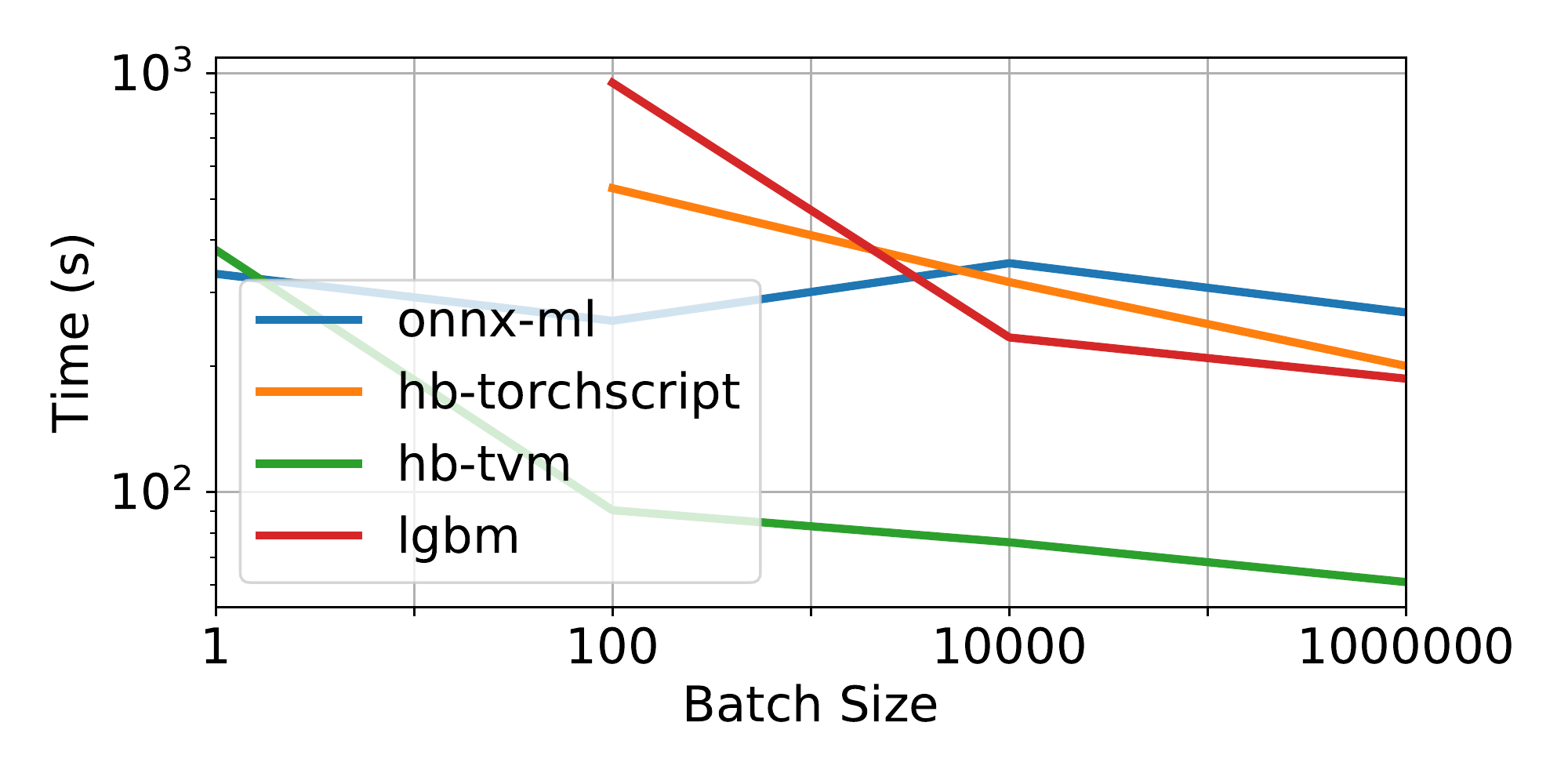}
		\vspace{-2.5ex}
		\caption{CPU (Higgs, LightGBM), 6 cores}
        \label{fig:scale-cpu}
	\end{subfigure}\vspace{-0.4ex}
	\begin{subfigure}{0.5\textwidth}
		\centering
		\includegraphics[trim={0 0 0 0.5cm},clip,width=0.7\textwidth]{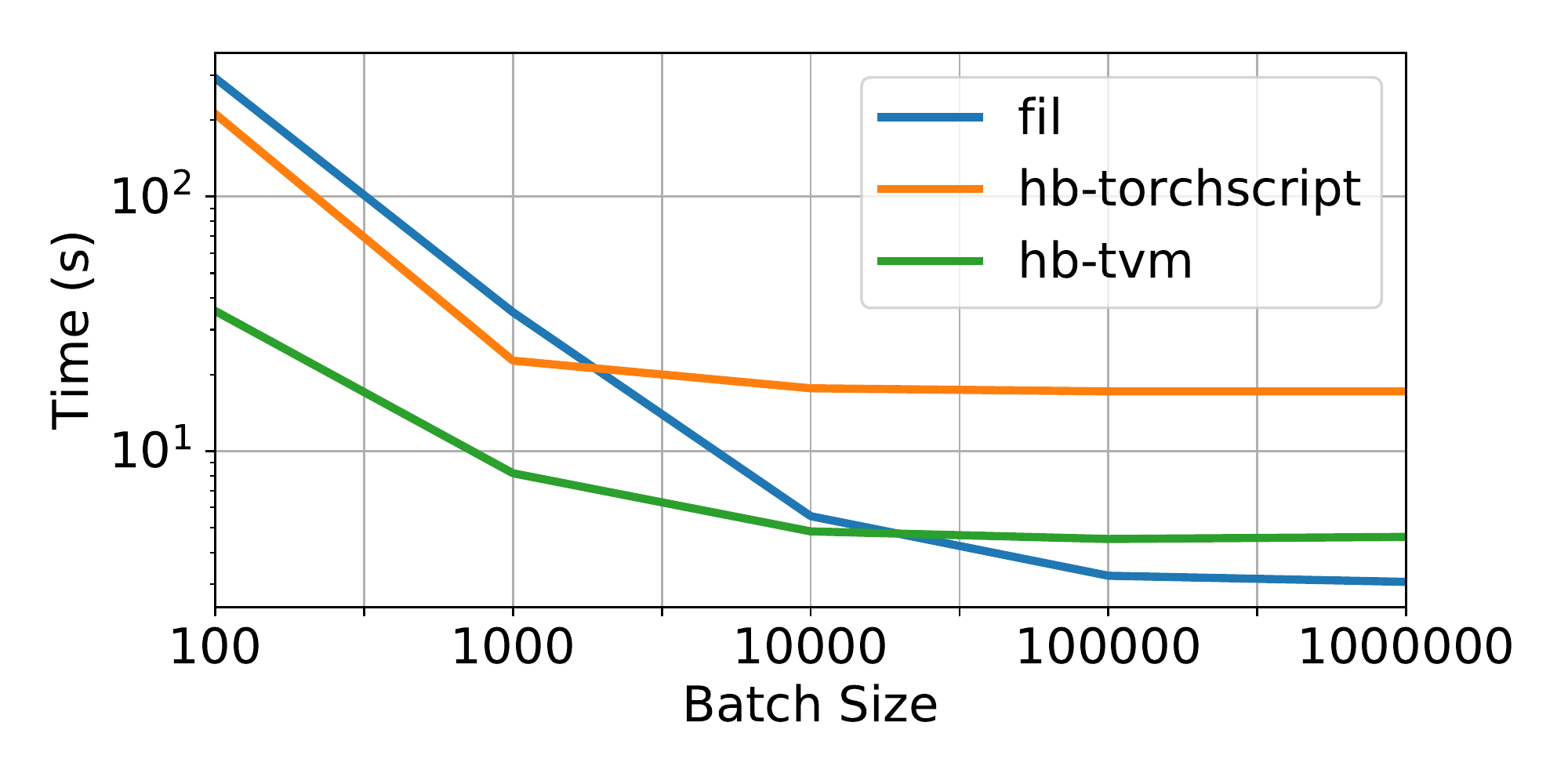}
		\vspace{-2.5ex}
		\caption{GPU (Airline, LightGBM)}
		\label{fig:scale-gpu}
	\end{subfigure}
	\vspace{-2ex}
 	\caption{Performance wrt scaling the batch size.}
 	\vspace{-3.5ex}
\end{figure}
In this scenario, one single record is scored at a time. For this experiment we run inference over the entire test datasets, but with batch size equal to 1. 
We used the same datasets and setup of Section~\ref{sec:tree-batch}, except that (1)  we removed the Airline dataset since no system was able to complete within the 1 hour timeout; and (2) we only use one single core.
The results are depicted in Table~\ref{tbl:tree-req-res}:
\begin{enumerate}\vspace{-1ex}
    \item 
    Unlike the batch scenario, ONNX-ML is much faster compared to scikit-learn, in some cases even more than 100$\times$. The reason is that ONNX-ML is currently optimized  for single record, single core inference, whereas scikit-learn design is more towards batch inference.
    \item PyTorch and TorchScript, again, behave very similarly. For random forest they are faster than scikit-learn but up to 5$\times$ slower compared to ONNX-ML. For LightGBM and XGBoost they are sometimes on par with scikit-learn, sometime slower.
    \item TVM provides the best performance in 11 cases out of 15, with a best case of 3$\times$ compared to the baselines.
\end{enumerate}\vspace{-1ex}

\begin{table}[h]
\caption{Request/response times in seconds (one record at-a-time).} 
\vspace{-2ex}
\hspace{-1ex}
\label{tbl:tree-req-res}
{
\scalebox{0.9}{
\notsotiny
\tablecolumnmarginsmall
\begin{tabular}{ccclllll}
\toprule
\multirow{2}[3]{*}{Algorithm} & \multirow{2}[3]{*}{Dataset}  & \multicolumn{2}{c}{Baselines}   &
\multicolumn{3}{c}{\system} \\
\cmidrule(lr){3-4} \cmidrule(lr){5-7}
& & \multicolumn{1}{c}{Sklearn} & \multicolumn{1}{c}{ONNX-ML} & \multicolumn{1}{c}{PT} & \multicolumn{1}{c}{TS} &
\multicolumn{1}{c}{TVM} \\ \midrule
\multirow{5}{*}{Rand. Forest} & Fraud & \multicolumn{1}{c}{1688.22} & \multicolumn{1}{c}{\textbf{9.96}} & \multicolumn{1}{c}{84.95} & \multicolumn{1}{c}{75.5} &
\multicolumn{1}{c}{11.63}\\
& Epsilon & \multicolumn{1}{c}{2945.42} & \multicolumn{1}{c}{32.58} & \multicolumn{1}{c}{153.32} & \multicolumn{1}{c}{134.17} &
\multicolumn{1}{c}{\textbf{20.4}}\\
& Year & \multicolumn{1}{c}{1152.56} & \multicolumn{1}{c}{18.99} & \multicolumn{1}{c}{84.82} & \multicolumn{1}{c}{74.21} &
\multicolumn{1}{c}{\textbf{9.13}}\\
& Covtype & \multicolumn{1}{c}{3388.50} & \multicolumn{1}{c}{35.49} & \multicolumn{1}{c}{179.4} & \multicolumn{1}{c}{157.8} &
\multicolumn{1}{c}{\textbf{34.1}}\\
& Higgs & \multicolumn{1}{c}{timeout} & \multicolumn{1}{c}{\textbf{335.23}} & \multicolumn{1}{c}{timeout} & \multicolumn{1}{c}{timeout} &
\multicolumn{1}{c}{450.65}\\
\midrule
\multirow{5}{*}{LightGBM} & Fraud & \multicolumn{1}{c}{354.27} & \multicolumn{1}{c}{12.05} & \multicolumn{1}{c}{96.5} & \multicolumn{1}{c}{84.56} &
\multicolumn{1}{c}{\textbf{10.19}}\\
& Epsilon & \multicolumn{1}{c}{40.7} & \multicolumn{1}{c}{29.28} & \multicolumn{1}{c}{167.43} & \multicolumn{1}{c}{148.87} &
\multicolumn{1}{c}{\textbf{17.3}}\\
& Year & \multicolumn{1}{c}{770.11} & \multicolumn{1}{c}{16.51} & \multicolumn{1}{c}{84.55} & \multicolumn{1}{c}{74.05} &
\multicolumn{1}{c}{\textbf{9.27}}\\
& Covtype & \multicolumn{1}{c}{135.39} & \multicolumn{1}{c}{209.16} &
\multicolumn{1}{c}{854.07} & \multicolumn{1}{c}{822.93} &
\multicolumn{1}{c}{\textbf{42.86}}\\
& Higgs & \multicolumn{1}{c}{timeout} & \multicolumn{1}{c}{\textbf{374.64}} & \multicolumn{1}{c}{timeout} & \multicolumn{1}{c}{timeout} &
\multicolumn{1}{c}{391.7}\\
\midrule
\multirow{5}{*}{XGBoost}& Fraud & \multicolumn{1}{c}{79.99} & \multicolumn{1}{c}{\textbf{7.78}} & \multicolumn{1}{c}{96.84} & \multicolumn{1}{c}{84.61} &
\multicolumn{1}{c}{10.21}\\
& Epsilon & \multicolumn{1}{c}{121.21} & \multicolumn{1}{c}{27.51} & \multicolumn{1}{c}{169.03} & \multicolumn{1}{c}{148.76} &
\multicolumn{1}{c}{\textbf{17.4}}\\
& Year & \multicolumn{1}{c}{98.67} & \multicolumn{1}{c}{17.14} & \multicolumn{1}{c}{85.23} & \multicolumn{1}{c}{74.62} &
\multicolumn{1}{c}{\textbf{9.25}}\\
& Covtype & \multicolumn{1}{c}{135.3} & \multicolumn{1}{c}{197.09} & \multicolumn{1}{c}{883.64} & \multicolumn{1}{c}{818.39} &
\multicolumn{1}{c}{\textbf{43.65}}\\
& Higgs & \multicolumn{1}{c}{timeout} & \multicolumn{1}{c}{585.89} & \multicolumn{1}{c}{timeout} & \multicolumn{1}{c}{timeout} &
\multicolumn{1}{c}{\textbf{425.12}}\\
\bottomrule
\end{tabular}
}
\vspace{-2ex}
}
\end{table}

\begin{table}[h!]
\vspace{-0ex}
\caption{Peak memory consumption  (in MB) for Fraud.} 
\vspace{-2ex}
\centering
\label{tbl:tree-memory}
{
\scalebox{0.9}{
\notsotiny
\tablecolumnmarginsmall
\begin{tabular}{cccc}
\toprule
Framework   & Random Forest & LightGBM & XGBoost \\ \midrule
Sklearn & 180 & 182 & 392 \\
ONNX-ML & 265 & 258 & 432\\
TorchScript & 375 & 370 & 568\\
TVM & 568 & 620 & 811\\
\bottomrule
\end{tabular}
}
}
\vspace{-2ex}
\end{table}

\noindent These results are again surprising, considering that tensor operations should be more optimized for bulk workloads rather than request/response scenarios.

\stitle{Scaling the Batch Size.}
\label{sec:tree-scaling}
We study how the performance of baselines and \system's backends change with the batch size. Figures~\ref{fig:scale-cpu} and~\ref{fig:scale-gpu} depicts the performance variation over CPU and GPU, respectively. We report only a few combinations of dataset / algorithm, but all the other combinations behave similarly. Starting with the CPU experiment, we can see that ONNX-ML has the best runtime for batch size of 1, but then its performance remains flat as we increase the batch size. TorchScript and scikit-learn did not complete within the timeout for batch equal to 1, but, past 100, they both scale linearly as we increase the batch size. TVM is comparable to ONNX-ML for batch of 1; for batches of 100 records it gets about 5$\times$ faster, while it scales like TorchScript for batches greater than 100. This is likely due to the fact that TVM applies a set of optimizations (e.g., operator fusion) that introduce a constant-factor speedup compared to TorchScript.

Looking at the GPU numbers (Figure~\ref{fig:scale-gpu}), TorchScript and TVM again follow a similar trend, with TVM being around 3$\times$ faster than TorchScript. Both TVM and TorchScript plateau at about a batch size of $10K$. RAPIDS FIL is slower than TorchScript for small batch sizes, but it scales better than \system. This is because of its custom CUDA implementation that is able to better use hardware under higher utilization. Interestingly, FIL as well plateaus at around $100K$ records. The custom CUDA implementation introduces a 50\% gain over \system with TVM runtime over large batches.

\begin{figure}[t]
	\begin{subfigure}{0.5\textwidth}
		\centering
    	 \includegraphics[trim={0 0 0 0.7cm},clip,width=0.7\textwidth]{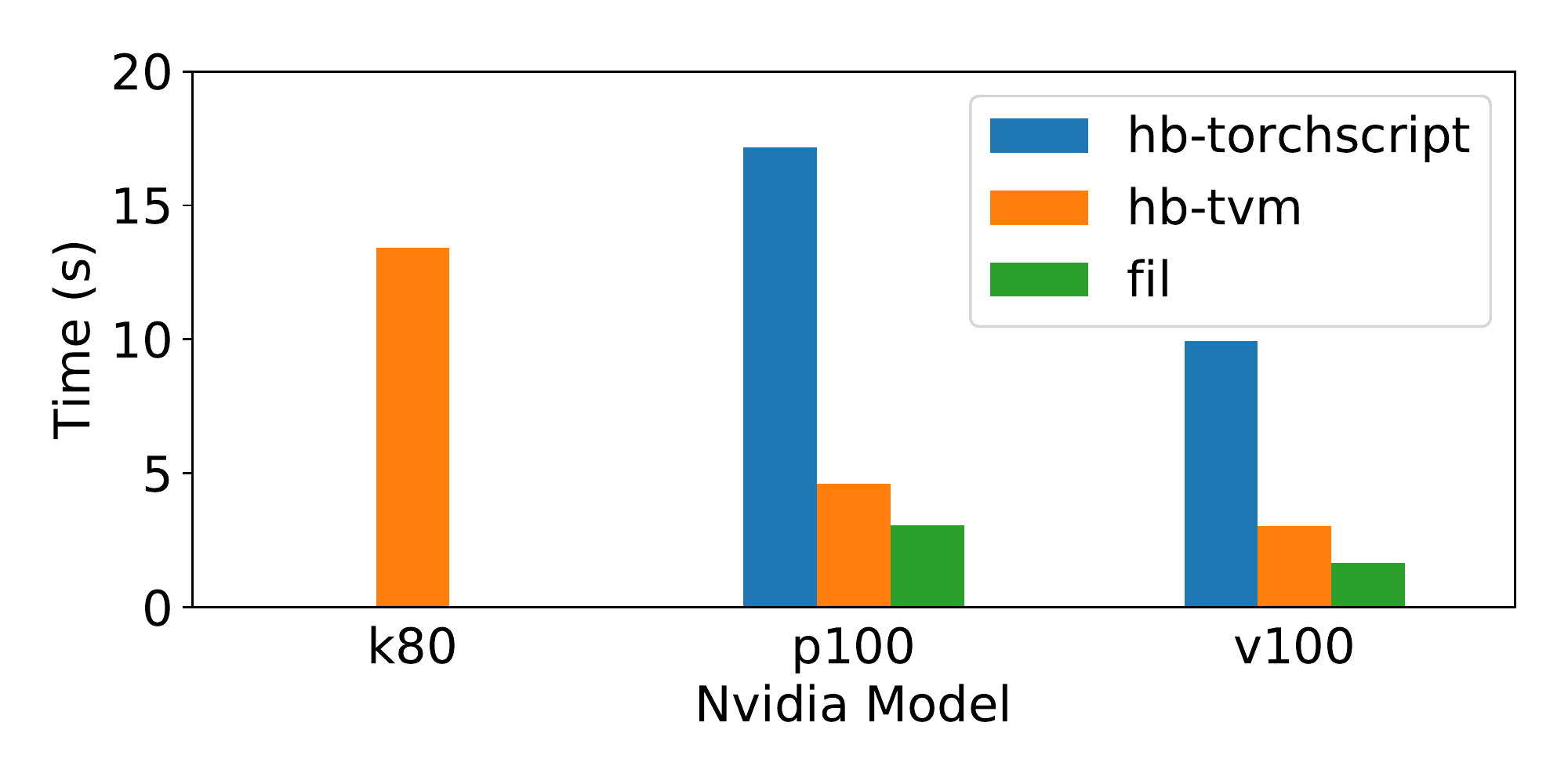}
    	\vspace{-2.5ex}
     	\caption{Batch size $1M$ }
     	\label{fig:k80p100v100-1M}
     	\vspace{-0ex}
	\end{subfigure}
	\vspace{1ex}
	\begin{subfigure}{0.5\textwidth}
		\centering
		\includegraphics[trim={0 0 0 0.7cm},clip,width=0.7\textwidth]{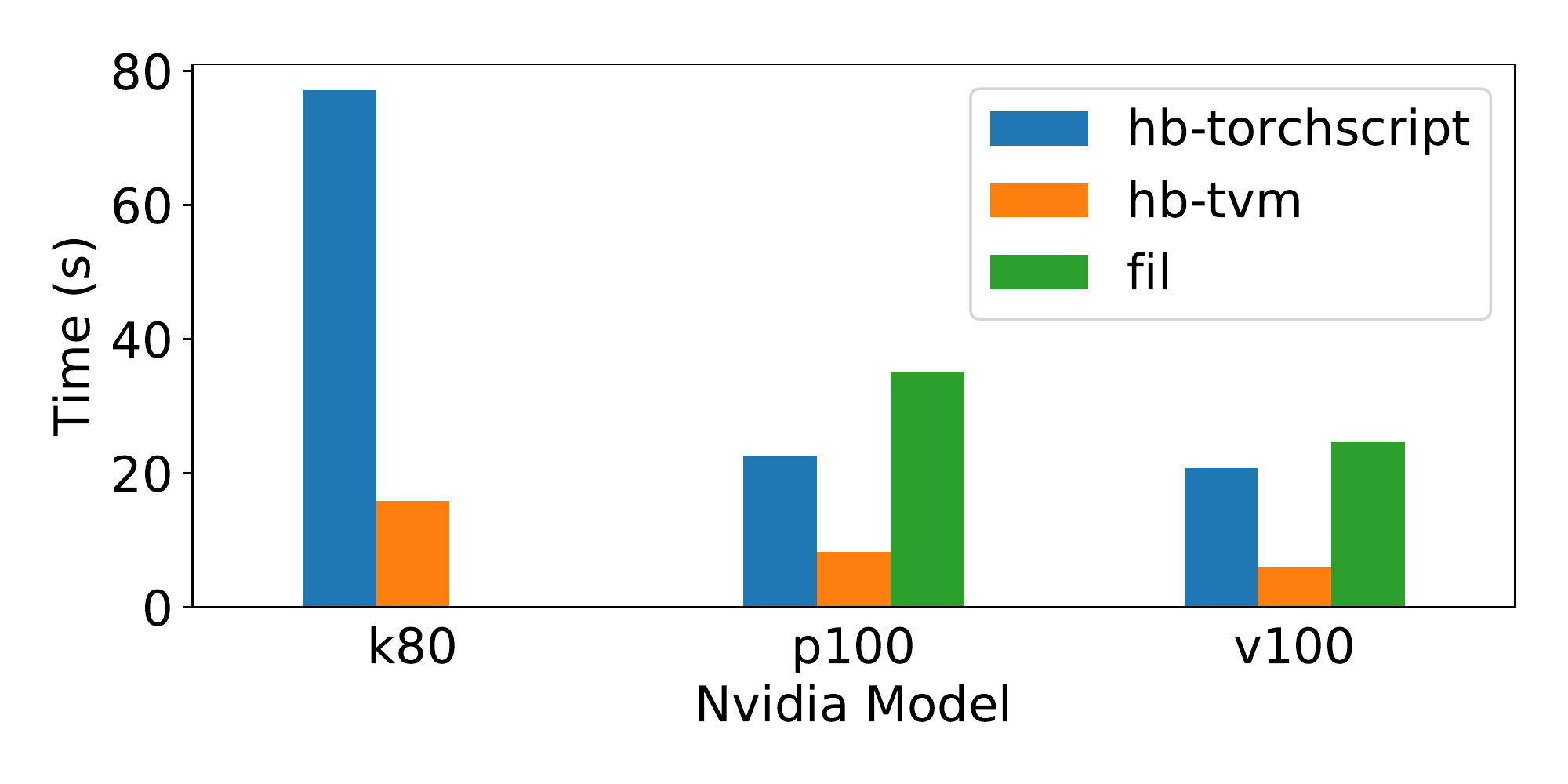}
		\vspace{-2.5ex}
		\caption{Batch size 1$K$}
    	\label{fig:k80p100v100-1k}
	\end{subfigure}
	\vspace{-3ex}
	\captionof{figure}{Performance across GPUs for Airline, LightGBM}
	\label{fig:k80p100v100}
 	\vspace{-2.5ex}
\end{figure}

\begin{table}[t]
\caption{
Conversion times (in seconds) over one core.}\vspace{-2ex}
\centering
\label{tbl:tree-conversion}
\scalebox{0.95}{
\notsotiny
\tablecolumnmarginsmall
\begin{tabular}{ccclll}
\toprule
\multirow{2}[3]{*}{Algorithm} & \multirow{2}[3]{*}{Dataset} & \multirow{2}[3]{*}{ONNX-ML} & \multicolumn{3}{c}{\system} \\
\cmidrule(lr){4-6}
& &  & \multicolumn{1}{c}{PyTorch} & \multicolumn{1}{c}{TorchScript} & \multicolumn{1}{c}{TVM} \\ \midrule \multirow{5}{*}{Rand.Forest} & Fraud & \multicolumn{1}{c}{1.28} & \multicolumn{1}{c}{0.55} & \multicolumn{1}{c}{0.58} &
\multicolumn{1}{c}{102.37}\\
& Epsilon & \multicolumn{1}{c}{7.53} & \multicolumn{1}{c}{2.63} & \multicolumn{1}{c}{2.67} &
\multicolumn{1}{c}{108.64}\\
& Year & \multicolumn{1}{c}{7.11} & \multicolumn{1}{c}{2.77} & \multicolumn{1}{c}{2.86} &
\multicolumn{1}{c}{69.99}\\
& Covtype & \multicolumn{1}{c}{9.87} & \multicolumn{1}{c}{2.16} & \multicolumn{1}{c}{2.2} &
\multicolumn{1}{c}{106.8}\\
& Higgs & \multicolumn{1}{c}{8.25} & \multicolumn{1}{c}{2.41} & \multicolumn{1}{c}{2.44} &
\multicolumn{1}{c}{103.77}\\
& Airline & \multicolumn{1}{c}{6.82} & \multicolumn{1}{c}{2.42} & \multicolumn{1}{c}{2.53} &
\multicolumn{1}{c}{391.07}\\
\midrule
\multirow{5}{*}{LightGBM} & Fraud & \multicolumn{1}{c}{1.34} & \multicolumn{1}{c}{0.98} & \multicolumn{1}{c}{1.06} &
\multicolumn{1}{c}{3.42}\\
& Epsilon & \multicolumn{1}{c}{11.71} & \multicolumn{1}{c}{7.55} & \multicolumn{1}{c}{7.60} &
\multicolumn{1}{c}{9.95}\\
& Year & \multicolumn{1}{c}{9.49} & \multicolumn{1}{c}{6.11} & \multicolumn{1}{c}{6.15} &
\multicolumn{1}{c}{8.35}\\
& Covtype & \multicolumn{1}{c}{32.46} & \multicolumn{1}{c}{22.57} & \multicolumn{1}{c}{23.12} &
\multicolumn{1}{c}{26.36}\\
& Higgs & \multicolumn{1}{c}{6.73} & \multicolumn{1}{c}{25.04} & \multicolumn{1}{c}{26.3} &
\multicolumn{1}{c}{109}\\
& Airline & \multicolumn{1}{c}{11.52} & \multicolumn{1}{c}{6.38} & \multicolumn{1}{c}{6.47} &
\multicolumn{1}{c}{8.19}\\
\midrule
\multirow{5}{*}{XGBoost}& Fraud & \multicolumn{1}{c}{0.55} & \multicolumn{1}{c}{0.65} & \multicolumn{1}{c}{0.7} &
\multicolumn{1}{c}{86.59}\\
& Epsilon & \multicolumn{1}{c}{6.86} & \multicolumn{1}{c}{25.89} & \multicolumn{1}{c}{25.94} &
\multicolumn{1}{c}{113.4}\\
& Year & \multicolumn{1}{c}{5.66} & \multicolumn{1}{c}{23.4} & \multicolumn{1}{c}{23.54} &
\multicolumn{1}{c}{110.24}\\
& Covtype & \multicolumn{1}{c}{9.87} & \multicolumn{1}{c}{2.16} & \multicolumn{1}{c}{2.20} &
\multicolumn{1}{c}{106.8}\\
& Higgs & \multicolumn{1}{c}{6.73} & \multicolumn{1}{c}{25.04} & \multicolumn{1}{c}{26.3} &
\multicolumn{1}{c}{109}\\
\bottomrule
\end{tabular}
}
\vspace{-2ex}
\end{table}

\stitle{Scaling Hardware.}
\label{sec:tree-hardware}
We tested how RAPIDS FIL and \system (TorchScript and TVM) scale as we change the GPU model.
For this experiment we tried both with a large batch size ($1M$ records, Figure~\ref{fig:k80p100v100} (a)) to maximize hardware utilization, and a smaller batch size ($1K$, Figure~\ref{fig:k80p100v100} (b)). 
 We ran this on all datasets across random forest, LightGBM, XGBoost with similar results, and present the Airline dataset (the largest) with LightGBM as a representative sample.
We tested on three NVIDIA devices: K80 (the oldest, 2014), P100 (2016), and V100 (2017). 
From the figures, in general we can see that: (1) RAPIDS FIL does not run on the K80 because it is an old generation; (2) with a batch size of 1K we get slower total inference time because we don't utilize the full hardware; (3) TorchScript and TVM runtimes for \system scale similarly on different hardware, although TVM is consistently 4 to 7$\times$ faster;  (4) FIL scales similarly to \system, although it is 50\% faster on large batches, 3$\times$ slower for smaller batches; (5) TorchScript is not optimal in memory management because for batches of $1M$ it fails on the K80 with an OOM exception. 
Finally, we also were able to run \system on the new Graphcore IPU~\cite{graphcore} over a single decision tree. 

\stitle{Cost.} 
Figure~\ref{fig:cost} shows the cost comparison between the Azure VM instance equipped with GPU, and a comparable one without GPU (E8 v3). The plot shows the cost of executing 100k samples with a batch size of 1K for random forest. The cost is calculated based on the hourly rate of each VM divided by the amortized cost of a single prediction. We executed scikit-learn on the CPU and TorchScript and TVM on the GPU for comparison. We found that the CPU cost was significantly higher (between 10$\times$-120$\times$) across all experiments.~\footnote{Note: airline times out for random forest for CPU with 1$K$ batch.} An interesting result was that the oldest GPU was the most cost effective, with the K80 and TVM having the lowest cost for 13 out of the 18 experiments (including LightGBM and XGBoost, not pictured). This result is explained by the fact that the K80 is readily available at significantly lower cost.

\eat{
\begin{figure}  
	\centering
	\begin{subfigure}{0.5\textwidth}
		\centering
    	\includegraphics[width=0.8\textwidth]{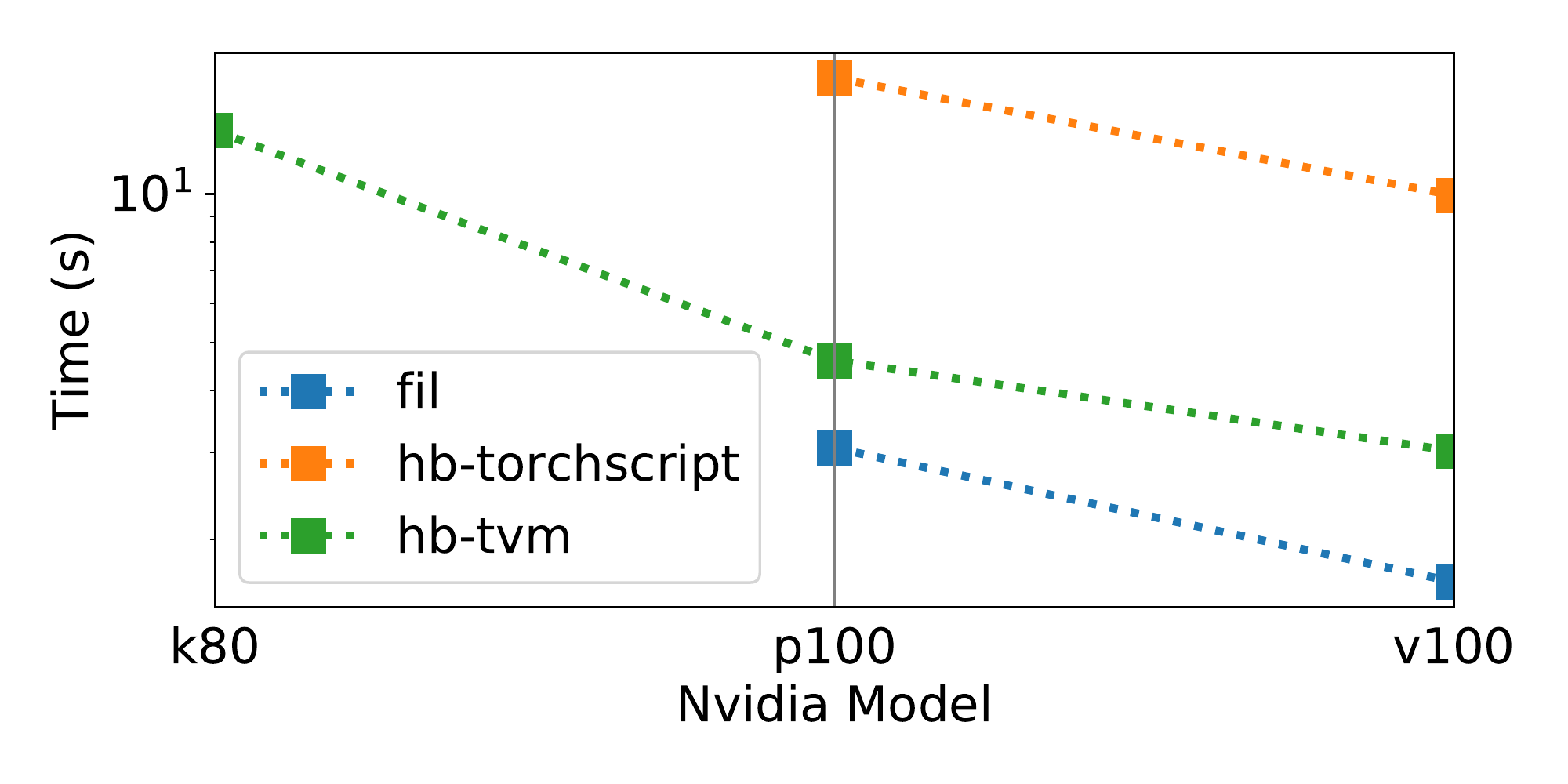}
    	\vspace{-2ex}
     	\caption{Batch size $1M$ }
     	\label{fig:k80p100v100-1M}
	\end{subfigure}
	\vspace{-2ex}
	\begin{subfigure}{0.5\textwidth}
		\centering
		\includegraphics[trim={0 0 0 0.5cm},clip,width=0.7\textwidth]{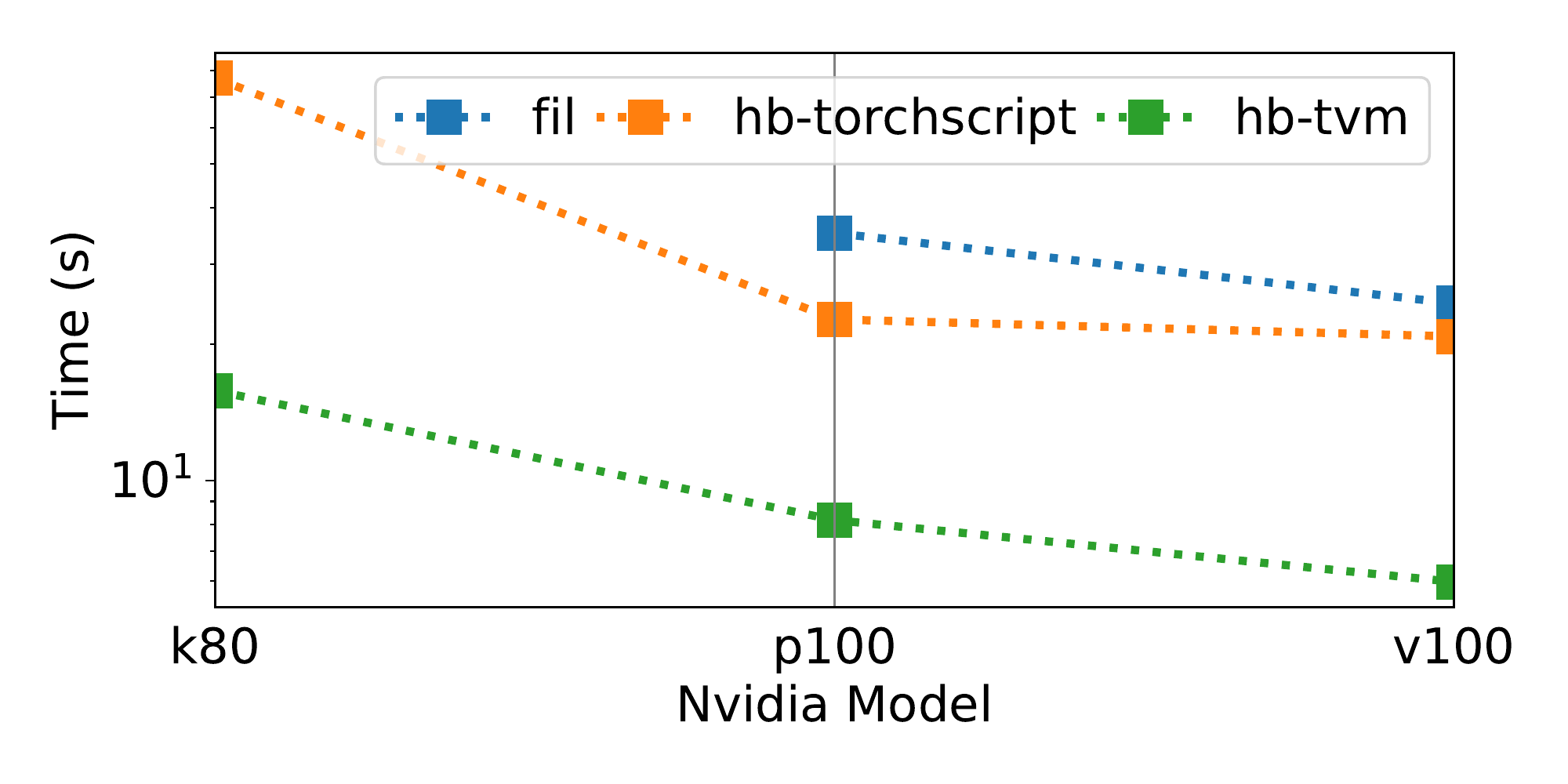}
		\vspace{-2ex}
		\caption{Batch size 1K}
    	\label{fig:k80p100v100-1k}
	\end{subfigure}
	\vspace{-1.5ex}
 	\caption{Performance across different GPU models for (Airline, LightGBM)}
 	\label{fig:k80p100v100}
 	\vspace{-4ex}
\end{figure}
}

\begin{figure}
\vspace{-1ex}
\hspace{-3ex}
		\centering
		\includegraphics[width=0.45\textwidth]{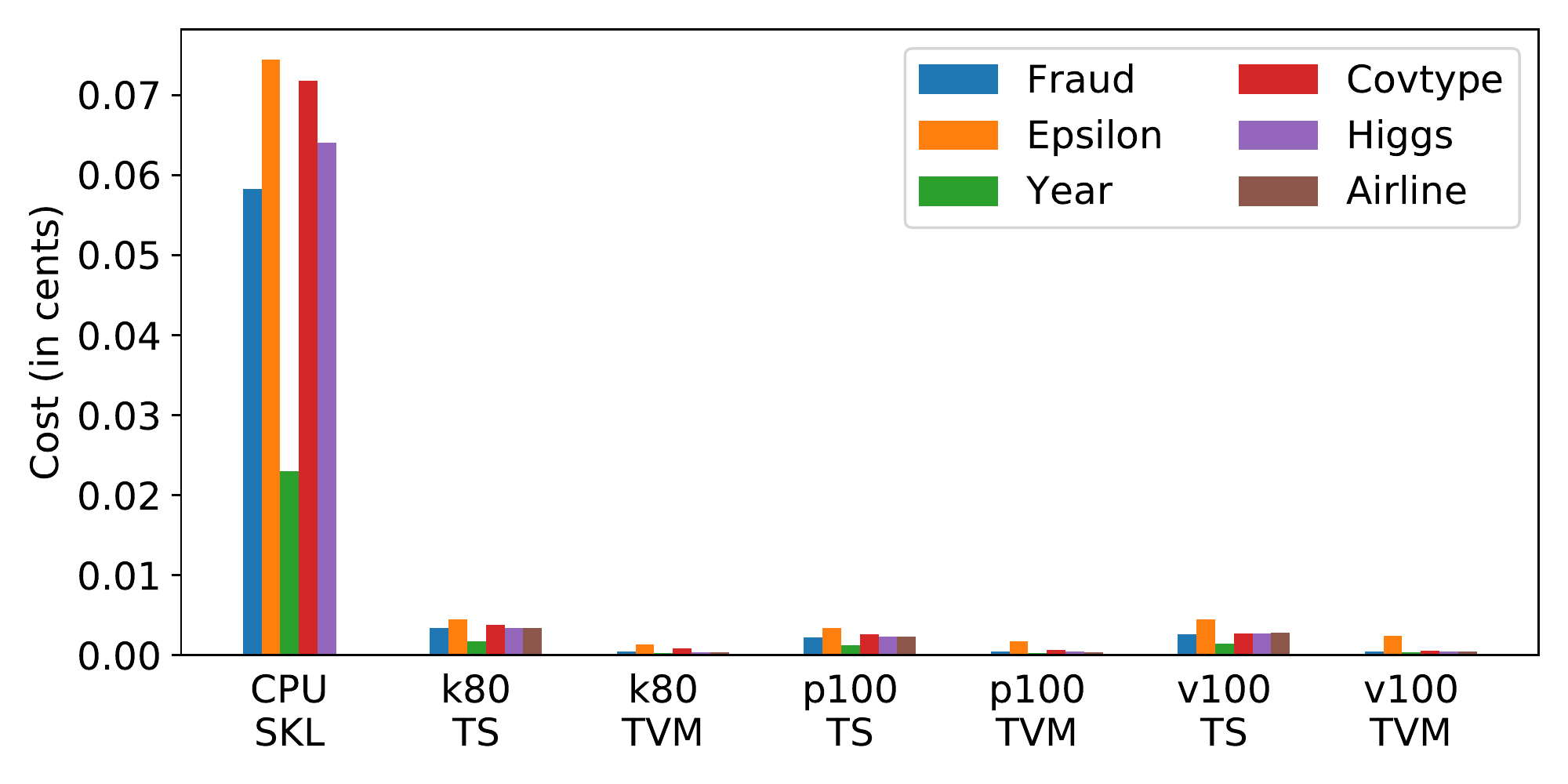}
		\vspace{-2ex}
		\caption{Cost for random forest 100k samples, batch size of 1$K$.}
        \label{fig:cost}
 	\vspace{-3ex}
\end{figure}


\stitle{Memory Consumption.}
\label{sec:tree-memory}
We measured the peak memory consumption over the Fraud dataset and for each algorithm. We used the \texttt{memory\_usage} function in the \texttt{memory\_profiler} library~\cite{mem_profiler}. The numbers are reported in Table~\ref{tbl:tree-memory}, and are the result of the execution over 1 core with a batch size of 1$K$. As we can see, scikit-learn is always the most memory efficient. ONNX-ML consumes from 10\% to 50\% more memory, while \system with TorchScript runtime consumes from 50\% to about 2$\times$ more memory than scikit-learn. Conversely, TVM consumes from 2$\times$ to 3$\times$ more memory wrt scikit-learn. We think that TVM is more memory hungry because it optimizes compute at the cost of memory requirements. 
Note that the batch size influences the total memory consumption.

\stitle{Output Validation.}
\label{sec:tree-validation}
Since we run tree ensemble models as tensor operations,
we could introduce rounding errors over floating point operations. Therefore, we need to validate that indeed the outputs produced match.  To evaluate this, we used the \texttt{numpy} \texttt{testing.assert\_allclose} function, and we set the relative and absolute errors to $10^{-5}$. We validate both the final scores and the probabilities (when available) for all combinations of datasets and algorithms. Out of the 18 experiments listed in Table~\ref{tbl:scenario-1-data-size}, 9 of them returned no mismatches for \system, 12 in the ONNX-ML case. 
Among the mismatches, the worst case for \system is random forest with Covtype where we have 0.8\% of records differing from the original scikit-learn output. For the Epsilon dataset, \system with random forest returns a mismatch on 0.1\% of records. All the remaining mismatches effect less than 0.1\% of records. 
Note that the differences are small. The biggest mismatch is of 0.086 (absolute difference) for Higgs using LightGBM. For the same experiment ONNX-ML has an absolute difference of 0.115.  

\stitle{Conversion Time.}
\label{sec:tree-conversion}
Table~\ref{tbl:tree-conversion} shows the time it takes to convert a trained model into a target framework.
The numbers are related to the generation of models running on a single core. This cost occurs only once per model and are not part of the inference cost.
As we can see, converting a model to ONNX-ML can take up to a few tens of seconds; \system with PyTorch backend is constantly about 2$\times$ to 3$\times$ faster wrt ONNX-ML in converting random forests models, while it varies for LightGBM and XGBModels. TorchScript models are generated starting from PyTorch models, and in general this further compilation step does not introduce any major overhead. Finally, conversion to TVM is much slower, and it might take more than 3 minutes.
This is due to code generation and optimizations introduced in TVM.

As a final note: parallel (i.e., more than 1 core) and GPU execution introduced further conversion time overheads, especially on TVM. For instance, TVM can take up to 40 minutes to convert a random forest model for execution on GPU.

\vspace{-1.5ex}
\subsubsection{Operators}
\label{sec:mini-ops}
\vspace{-1.5ex}

\eat{
\begin{itemize}
    \item \stitle{Setup.} Baselines: scikit-learn and onnx-ml. Hardware: P100 machine, one core. Backends: torchscript, TVM.
    \item \st{Scaling experiments}
    \item \st{gpu experiments}
    \item check experiments
    \item memory experiments
\end{itemize}
}

\stitle{Setup.} This micro-benchmark is a replication of the suite comparing scikit-learn and ONNX-ML operators~\cite{onnx-bench}.
We test all scikit-learn operators of the suite that are supported by both ONNX-ML and \system (minus tree ensembles models). The total number of tested operators is 13, and they are a mix of ML models (Logistic Regression, Support Vector Machines, etc.) and featurizers (e.g., Binarizer, Polynomial, etc.). For this micro-benchmark we score 1 million records.

\stitle{Datasets.} We use the Iris datasets~\cite{iris} with 20 features.

\stitle{List of Experiments.} We run the following experiments: (1) batch inference over $1M$ records, both on CPU and GPU;
(2) request/response over 1 record; (3) memory consumption and conversion time.
All the output results are correct.

\begin{table}[h!]
\vspace{-3ex}
\hspace{-2ex}
\caption{Batch experiments for operators on both CPU (1 core) and GPU. Numbers are in milliseconds. (TS is short for TorchScript)}
\vspace{-0ex}
\label{tbl:ops-results}
\hspace{-1ex}
\scalebox{0.98}{
\notsotiny
\tablecolumnmarginsmall
\begin{tabular}{ccccccccccc}
\toprule
\multirow{2}[2]{*}{Operator}
& \multicolumn{2}{c}{Baselines (CPU)} & \multicolumn{2}{c}{\system CPU} &
\multicolumn{2}{c}{\system GPU}\\
\cmidrule(lr){2-3} \cmidrule(lr){4-5} \cmidrule(lr){6-7}
& \multicolumn{1}{c}{Sklearn} & \multicolumn{1}{c}{ONNX-ML} & \multicolumn{1}{c}{TS} & \multicolumn{1}{c}{TVM} &
\multicolumn{1}{c}{TS} & \multicolumn{1}{c}{TVM}\\ 
\midrule
 Log. Regres. &  \multicolumn{1}{c}{970} & \multicolumn{1}{c}{1540} & \multicolumn{1}{c}{260} & \multicolumn{1}{c}{\textbf{47}} & \multicolumn{1}{c}{\textbf{13}} & 15\\
SGDClass. & 
\multicolumn{1}{c}{180} & \multicolumn{1}{c}{1540} & \multicolumn{1}{c}{270} & \multicolumn{1}{c}{\textbf{49}} & \multicolumn{1}{c}{\textbf{11}} & 15\\
LinearSVC & 
\multicolumn{1}{c}{110} & \multicolumn{1}{c}{69} & \multicolumn{1}{c}{260} & \multicolumn{1}{c}{\textbf{51}} & \multicolumn{1}{c}{\textbf{12}} & 18\\
NuSVC & 
\multicolumn{1}{c}{3240} & \multicolumn{1}{c}{4410} & \multicolumn{1}{c}{\textbf{2800}} & \multicolumn{1}{c}{3000} & \multicolumn{1}{c}{140} & \textbf{72} \\
SVC & 
\multicolumn{1}{c}{1690} & \multicolumn{1}{c}{2670} & \multicolumn{1}{c}{\textbf{1520}} & \multicolumn{1}{c}{1560} & \multicolumn{1}{c}{120} & \textbf{41} \\
BernoulliNB & 
\multicolumn{1}{c}{280} & \multicolumn{1}{c}{1670} & \multicolumn{1}{c}{290} & \multicolumn{1}{c}{\textbf{65}} & \multicolumn{1}{c}{\textbf{12}} & 14 \\
MLPClassifier & 
\multicolumn{1}{c}{930} & \multicolumn{1}{c}{1860} & \multicolumn{1}{c}{\textbf{910}} & \multicolumn{1}{c}{1430} & \multicolumn{1}{c}{\textbf{17}} & 31\\
Dec.TreeClass. & 
\multicolumn{1}{c}{59} & \multicolumn{1}{c}{1610} & \multicolumn{1}{c}{560} & \multicolumn{1}{c}{\textbf{35}} & \multicolumn{1}{c}{\textbf{13}} & 16 \\
Binarizer & 
\multicolumn{1}{c}{98} & \multicolumn{1}{c}{75} & \multicolumn{1}{c}{\textbf{39}} & \multicolumn{1}{c}{59} & \multicolumn{1}{c}{\textbf{38}} & \textbf{38}\\
MinMaxScaler & 
\multicolumn{1}{c}{92} & \multicolumn{1}{c}{200} & \multicolumn{1}{c}{78} & \multicolumn{1}{c}{\textbf{57}} & \multicolumn{1}{c}{\textbf{38}} & \textbf{38} \\
Normalizer & 
\multicolumn{1}{c}{94} & \multicolumn{1}{c}{140} & \multicolumn{1}{c}{\textbf{83}} & \multicolumn{1}{c}{97} & \multicolumn{1}{c}{\textbf{39}} & 40 \\
Poly.Features & 
\multicolumn{1}{c}{4030} & \multicolumn{1}{c}{29160} & \multicolumn{1}{c}{6380} & \multicolumn{1}{c}{\textbf{3130}} & \multicolumn{1}{c}{\textbf{340}} & error \\
StandardScaler & 
\multicolumn{1}{c}{150} & \multicolumn{1}{c}{200} & \multicolumn{1}{c}{77} & \multicolumn{1}{c}{\textbf{58}} & \multicolumn{1}{c}{\textbf{38}} & \textbf{38} \\

\bottomrule
\end{tabular}
}
\vspace{-2ex}
\end{table}

\stitle{Batch Inference.}
\label{sec:ops-batch}
The batch numbers are reported in Table~\ref{tbl:ops-results}. 
On CPU, scikit-learn is faster than ONNX-ML, up to 6$\times$ for polynomial featurizer, although in most of the cases the two systems are within a factor of 2. \system with TorchScript backend is competitive with scikit-learn, whereas with TVM backend \system is faster on 8 out of 13 operators, with in general a speedup of about 2$\times$ compared to scikit-learn.
If now we focus to the GPU numbers, we see that \system with TorchScript backend compares favorably against TVM on 11 operators out of 13. 
This is in contrast with the tree ensemble micro-benchmark where the TVM backend was faster than the TorchScript one. We suspect that this is because TVM optimizations are less effective on these ``simpler'' operators. For the same reason, GPU acceleration does not provide the speedup we instead saw for the tree ensemble models. In general, we see around 2$\times$ performance improvement over the CPU runtime: only polynomial featurizer runs faster, with almost a 10$\times$ improvement. 
TVM returns a runtime error when generating the polynomial featurizer model on GPU.


\stitle{Request/response.}
Table~\ref{tbl:operations-rr-results} contains the times to score 1 record.
The results are similar to the request/response scenario for the tree ensemble micro-benchmark. Namely, ONNX-ML outperform both scikit-learn and \system in 9 out of 13 cases. Note, however, that all frameworks are within a factor of 2. The only outlier is polynomial featurizer which is about 10$\times$ faster on \system with TVM backend. 

\begin{table}[h]
\centering
\vspace{-0mm}
\caption{Request/Response experiments for operators on CPU (single core). Reported numbers are in milliseconds.}
\vspace{-2ex}
\label{tbl:operations-rr-results}
{
\scalebox{0.9}{
\notsotiny
\tablecolumnmarginsmall
\begin{tabular}{ccccccccccc}
\toprule
\multirow{2}[2]{*}{Operator} 
&\multicolumn{2}{c}{Baselines}  & \multicolumn{2}{c}{\system} \\
\cmidrule(lr){2-3} \cmidrule(lr){4-5}
& \multicolumn{1}{c}{Sklearn} & \multicolumn{1}{c}{ONNX-ML} & \multicolumn{1}{c}{TS} &  \multicolumn{1}{c}{TVM}\\ 
\midrule
 LogisticRegression & \multicolumn{1}{c}{0.087} & \multicolumn{1}{c}{\textbf{0.076}} & \multicolumn{1}{c}{0.1} & \multicolumn{1}{c}{0.1} \\
SGDClassifier & \multicolumn{1}{c}{\textbf{0.098}} & \multicolumn{1}{c}{0.1} & \multicolumn{1}{c}{0.12} & \multicolumn{1}{c}{0.1} \\
LinearSVC & \multicolumn{1}{c}{0.077} & \multicolumn{1}{c}{\textbf{0.05}} & \multicolumn{1}{c}{0.11} & \multicolumn{1}{c}{0.1} \\
NuSVC & \multicolumn{1}{c}{0.086} & \multicolumn{1}{c}{\textbf{0.072}} & \multicolumn{1}{c}{4.1} & \multicolumn{1}{c}{0.14} \\
SVC & \multicolumn{1}{c}{0.086} & \multicolumn{1}{c}{\textbf{0.074}} & \multicolumn{1}{c}{2.3} & \multicolumn{1}{c}{0.12}\\
BernoulliNB & \multicolumn{1}{c}{0.26} & \multicolumn{1}{c}{\textbf{0.1}} & \multicolumn{1}{c}{0.07} & \multicolumn{1}{c}{0.11} \\
MLPClassifier & \multicolumn{1}{c}{0.15} & \multicolumn{1}{c}{0.11} & \multicolumn{1}{c}{\textbf{0.1}} & \multicolumn{1}{c}{0.12} \\
DecisionTreeClassifier & \multicolumn{1}{c}{0.087} & \multicolumn{1}{c}{\textbf{0.074}} & \multicolumn{1}{c}{0.44} & \multicolumn{1}{c}{0.12} \\
Binarizer & \multicolumn{1}{c}{0.064} & \multicolumn{1}{c}{\textbf{0.053}} & \multicolumn{1}{c}{0.063} & \multicolumn{1}{c}{0.1} \\
MinMaxScaler & \multicolumn{1}{c}{0.066} & \multicolumn{1}{c}{0.060} & \multicolumn{1}{c}{\textbf{0.058}} & \multicolumn{1}{c}{0.1} \\
Normalizer & \multicolumn{1}{c}{0.11} & \multicolumn{1}{c}{\textbf{0.063}} & \multicolumn{1}{c}{0.072} & \multicolumn{1}{c}{0.1} \\
PolynomialFeatures & \multicolumn{1}{c}{1.2} & \multicolumn{1}{c}{1} & \multicolumn{1}{c}{0.5} & \multicolumn{1}{c}{\textbf{0.1}} \\
StandardScaler & \multicolumn{1}{c}{0.069} & \multicolumn{1}{c}{\textbf{0.048}} & \multicolumn{1}{c}{0.059} & \multicolumn{1}{c}{0.1} \\

\bottomrule
\end{tabular}
}
}
\vspace{-2ex}
\end{table}

\stitle{Memory Consumption and Conversion Time.}
\label{sec:ops-memory}
We measured the peak memory consumed and conversion time for each operator on each framework. We used batch inference over 1K records. For memory consumption, the results are in line with what we already saw in Section~\ref{sec:mini-trees}. 
Regarding the conversion time, for ONNX-ML and \system with TorchScript, the conversion time is in the order of few milliseconds. The TVM backend is slightly slower but still in the order of few tens of milliseconds (exception for NuSVC and SVC which take up to 3.2 seconds). In comparison with the tree ensembles numbers (Table~\ref{tbl:tree-conversion}), we confirm that these operators are simpler, even from a compilation perspective.

\eat{
\stitle{Conversion Time.}
\label{sec:ops-conversion}
Table~\ref{tbl:ops-conversion} lists the conversion time for each operator and target framework. These conversion times were generated using the same batch configuration as for the memory measurements. In general, for ONNX-ML and \system with TorchScript backed, the conversion time is in the order of few milliseonds. The TVM backend is slightly slower but still in the order of few tens of milliseconds (exception for NuSVC and SVC which take few seconds). 
In comparison with the numbers reported for the tree ensembles models (Table~\ref{tbl:tree-conversion}), we confirm that these operators are simpler, even from a compilation perspective.
}

\eat{
\begin{table*}
\begin{minipage}{.4\linewidth}
\centering
\caption{Peak memory consumption for each operator (in MBs).}
\label{tbl:ops-memory}
{
{
\notsotiny
\tablecolumnmarginsmall
\begin{tabular}{ccccc}
\toprule
\multirow{1}[1]{*}{Operator} 
& \multicolumn{1}{c}{Sklearn} & \multicolumn{1}{c}{ONNX-ML} & \multicolumn{1}{c}{TorchScript} &  \multicolumn{1}{c}{TVM} \\ 
\midrule
 LogisticRegression & \multicolumn{1}{c}{132} & \multicolumn{1}{c}{161} & \multicolumn{1}{c}{296} & \multicolumn{1}{c}{336} \\
SGDClassifier & \multicolumn{1}{c}{129} & \multicolumn{1}{c}{164} & \multicolumn{1}{c}{300} & \multicolumn{1}{c}{333} \\
LinearSVC & \multicolumn{1}{c}{130} & \multicolumn{1}{c}{162} & \multicolumn{1}{c}{298} & \multicolumn{1}{c}{334} \\
NuSVC & \multicolumn{1}{c}{120} & \multicolumn{1}{c}{159} & \multicolumn{1}{c}{304} & \multicolumn{1}{c}{360} \\
SVC & \multicolumn{1}{c}{128} & \multicolumn{1}{c}{163} & \multicolumn{1}{c}{280} & \multicolumn{1}{c}{351} \\
BernoulliNB & \multicolumn{1}{c}{129} & \multicolumn{1}{c}{163} & \multicolumn{1}{c}{267} & \multicolumn{1}{c}{333} \\
MLPClassifier & \multicolumn{1}{c}{173} & \multicolumn{1}{c}{204} & \multicolumn{1}{c}{317} & \multicolumn{1}{c}{379} \\
DecisionTreeClassifier & \multicolumn{1}{c}{128} & \multicolumn{1}{c}{161} & \multicolumn{1}{c}{272} & \multicolumn{1}{c}{338} \\
Binarizer & \multicolumn{1}{c}{127} & \multicolumn{1}{c}{159} & \multicolumn{1}{c}{263} & \multicolumn{1}{c}{332} \\
MaxAbsScaler & \multicolumn{1}{c}{127} & \multicolumn{1}{c}{159} & \multicolumn{1}{c}{267} & \multicolumn{1}{c}{334} \\
MinMaxScaler & \multicolumn{1}{c}{127} & \multicolumn{1}{c}{159} & \multicolumn{1}{c}{266} & \multicolumn{1}{c}{331} \\
Normalizer & \multicolumn{1}{c}{127} & \multicolumn{1}{c}{159} & \multicolumn{1}{c}{264} & \multicolumn{1}{c}{330} \\
PolynomialFeatures & \multicolumn{1}{c}{130} & \multicolumn{1}{c}{166} & \multicolumn{1}{c}{272} & \multicolumn{1}{c}{337} \\
RobustScaler & \multicolumn{1}{c}{127} & \multicolumn{1}{c}{159} & \multicolumn{1}{c}{263} & \multicolumn{1}{c}{333} \\
StandardScaler & \multicolumn{1}{c}{128} & \multicolumn{1}{c}{159} & \multicolumn{1}{c}{265} & \multicolumn{1}{c}{331} \\
\bottomrule
\end{tabular}
}
}
\end{minipage}
\hspace{2ex}\begin{minipage}{.65\linewidth}
\caption{Conversion time for operators. Numbers are in seconds.}
\centering
\label{tbl:ops-conversion}
\notsotiny
\tablecolumnmarginsmall
\begin{tabular}{ccccc}
\toprule
{ONNX-ML} & \multicolumn{1}{c}{TorchScript} &  \multicolumn{1}{c}{TVM} \\ 
\midrule
\multicolumn{1}{c}{0.002} & \multicolumn{1}{c}{0.021} & \multicolumn{1}{c}{0.47} \\
\multicolumn{1}{c}{0.005} & \multicolumn{1}{c}{0.007} & \multicolumn{1}{c}{0.47} \\
\multicolumn{1}{c}{0.002} & \multicolumn{1}{c}{0.007} & \multicolumn{1}{c}{0.49} \\
\multicolumn{1}{c}{0.004} & \multicolumn{1}{c}{0.001} & \multicolumn{1}{c}{1.98} \\
\multicolumn{1}{c}{0.004} & \multicolumn{1}{c}{0.001} & \multicolumn{1}{c}{3.23} \\
\multicolumn{1}{c}{0.006} & \multicolumn{1}{c}{0.001} & \multicolumn{1}{c}{0.46} \\
\multicolumn{1}{c}{0.005} & \multicolumn{1}{c}{0.009} & \multicolumn{1}{c}{0.65} \\
\multicolumn{1}{c}{0.004} & \multicolumn{1}{c}{0.026} & \multicolumn{1}{c}{0.53} \\
\multicolumn{1}{c}{0.002} & \multicolumn{1}{c}{0.005} & \multicolumn{1}{c}{0.37} \\
\multicolumn{1}{c}{0.001} & \multicolumn{1}{c}{0.006} & \multicolumn{1}{c}{0.39} \\
\multicolumn{1}{c}{0.001} & \multicolumn{1}{c}{0.005} & \multicolumn{1}{c}{0.36} \\
\multicolumn{1}{c}{0.001} & \multicolumn{1}{c}{0.006} & \multicolumn{1}{c}{0.37} \\
\multicolumn{1}{c}{0.098} & \multicolumn{1}{c}{0.01} & \multicolumn{1}{c}{0.41} \\
\multicolumn{1}{c}{0.001} & \multicolumn{1}{c}{0.006} & \multicolumn{1}{c}{0.36} \\
\multicolumn{1}{c}{0.002} & \multicolumn{1}{c}{0.006} & \multicolumn{1}{c}{0.36} \\
\bottomrule
\end{tabular}
\end{minipage}
\end{table*}
}

\begin{figure*}[t!]  
    \centering
    \begin{minipage}{0.85\textwidth}
    	\centering
    	\begin{subfigure}{\textwidth}
    		\centering
    		\includegraphics[trim={0.5cm 2.9cm 0 0},clip,width=1.\textwidth]{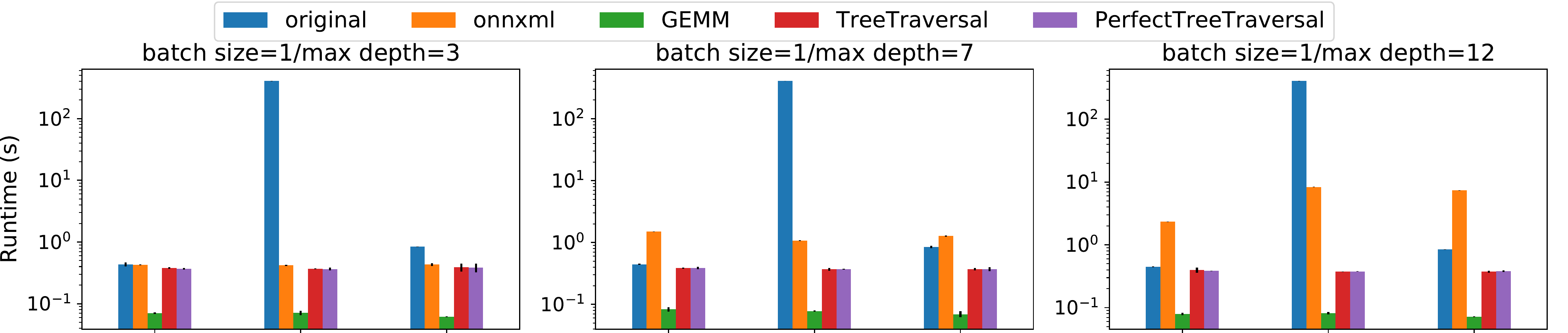}
            \label{fig:opt-tree-1}
    	\end{subfigure}
    	
    	\vspace{-1.5ex}
    	\begin{subfigure}{\textwidth}
    		\centering
    		\includegraphics[trim={0.5 0.9cm 0 1.1cm},clip,width=1.\textwidth]{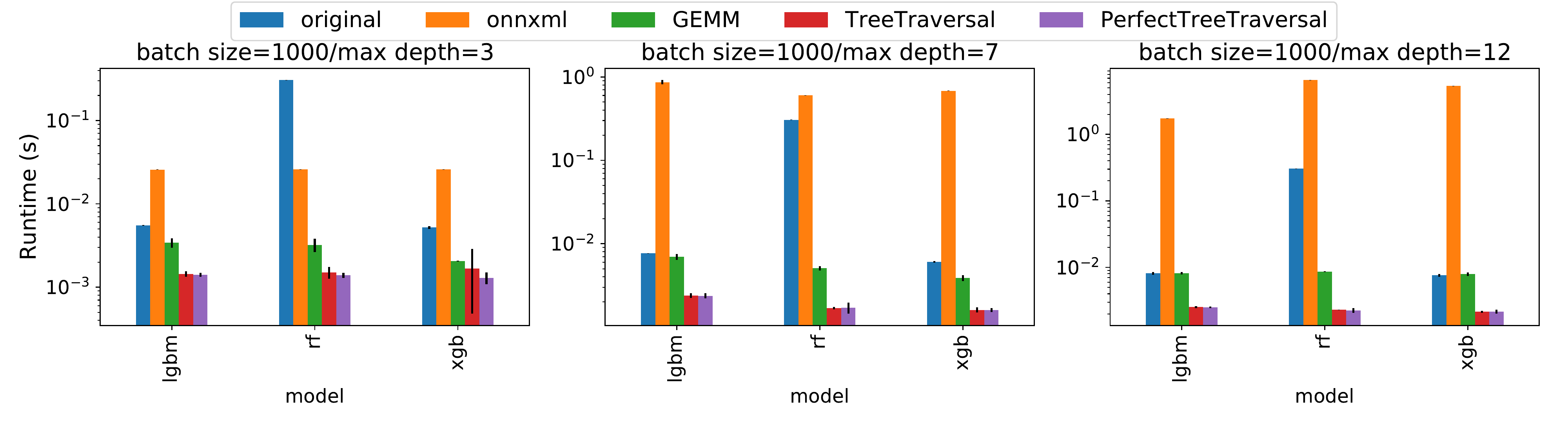}
    		\label{fig:opt-tree-1000}
    	\end{subfigure}
    	\vspace{-5ex}
     	\caption{Comparison between the different tree strategies as we vary the batch size and depth.}
     	\label{fig:opt-tree}
     \end{minipage}%
     \hspace{1ex}
        
        
 \vspace{-1.8ex}
\end{figure*}
 
 \begin{figure*}
 \begin{minipage}{.65\textwidth}
 \vspace{-0ex}
    \hspace{-1ex}
    \includegraphics[trim={0.5cm 0 8cm 0},clip,width=0.9\textwidth]{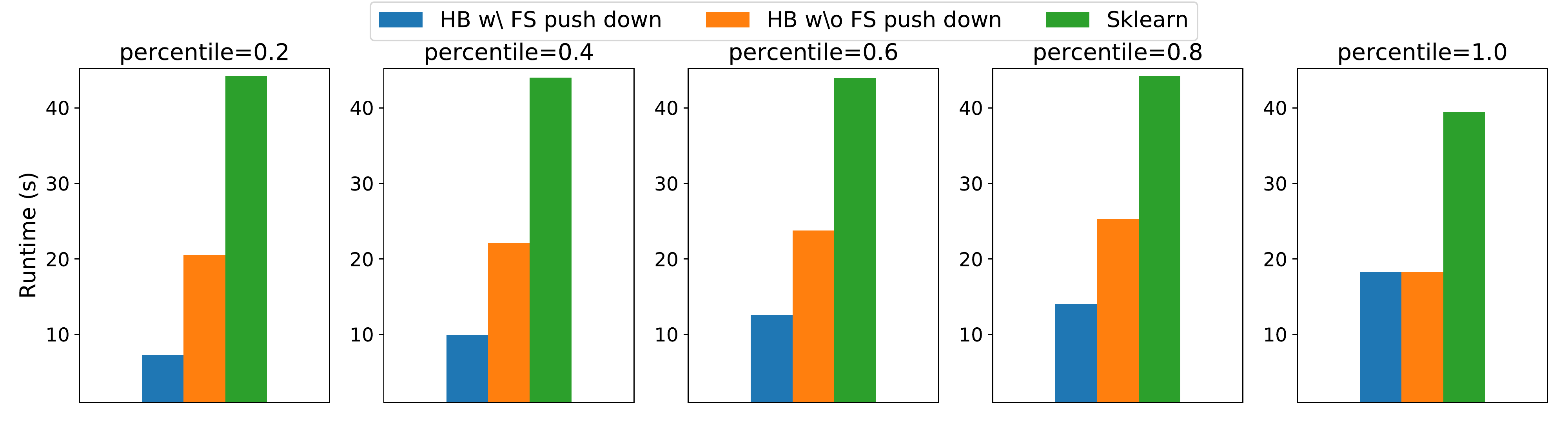}
	\vspace{-3.5ex}
	\caption{Feature selection push down.}
    \label{fig:opt-fs-pushdown}
    \vspace{0ex}
    \hspace{-1ex}
    \includegraphics[trim={0.5cm 0 8cm 0},clip,width=0.9\textwidth]{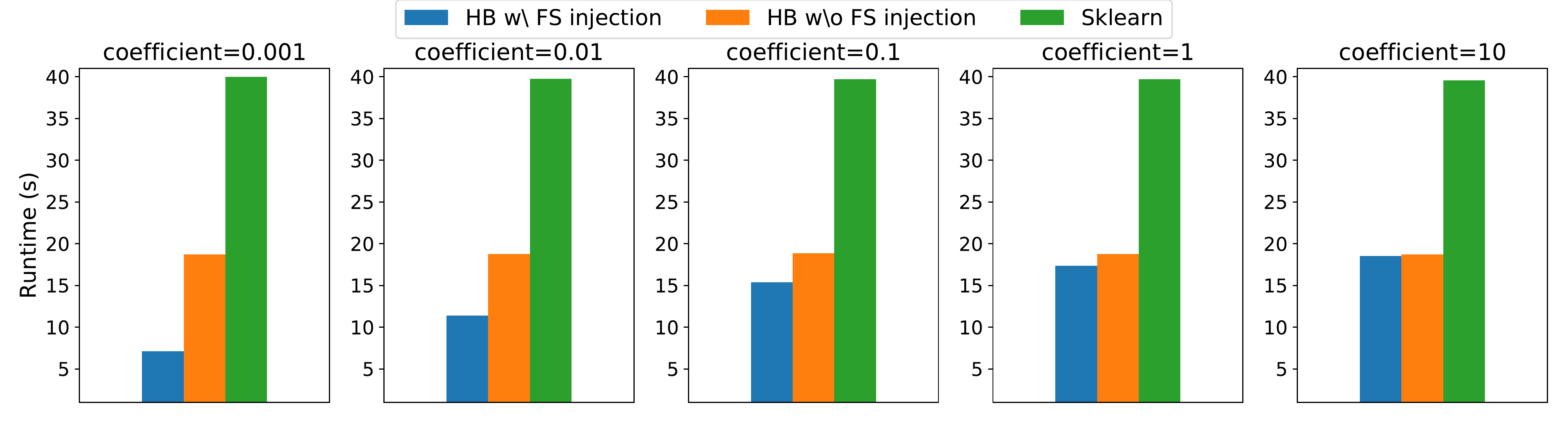}
	\vspace{-3.5ex}
	\caption{Feature selection injection.}
    \label{fig:opt-fs-injection}
    \vspace{-2ex}
    \end{minipage}
    \hspace{2ex}
    \begin{minipage}{.3\textwidth} 
    \hspace{-4.5ex}
    \vspace{1ex}
        \begin{subfigure}{1.18\textwidth}
        	\centering
        	\includegraphics[trim={0 0 0 0.5cm},clip,width=1.\textwidth]{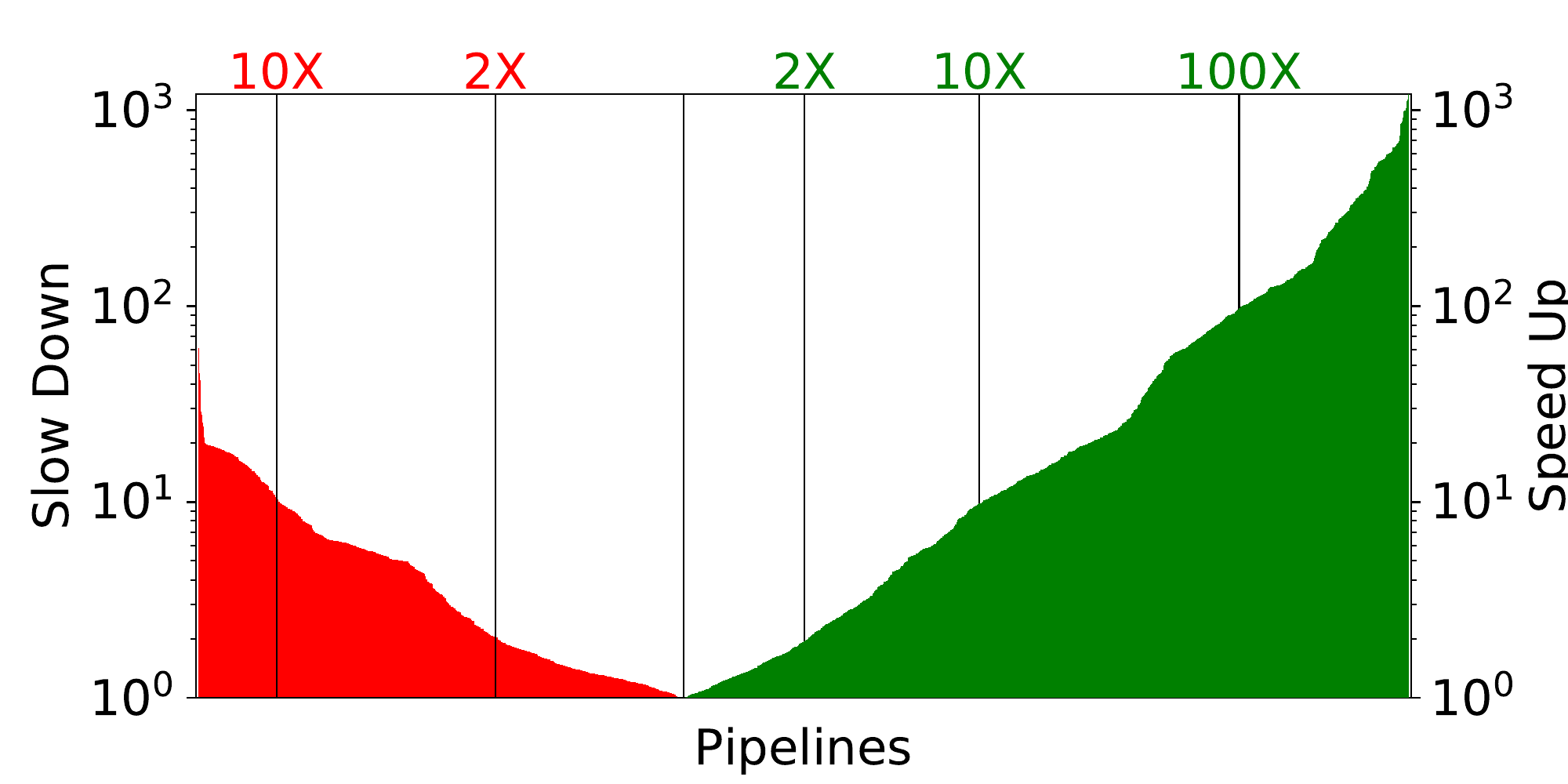}
        	\vspace{-4ex}
        	\caption{CPU}
            \label{fig:pip-ts-cpu}
        \end{subfigure}
        
        \vspace{-0.5ex}
        \hspace{-4.5ex}
        \begin{subfigure}{1.18\textwidth}
        	\centering
        	\includegraphics[trim={0 0 0 0.5cm},clip,width=1.\textwidth]{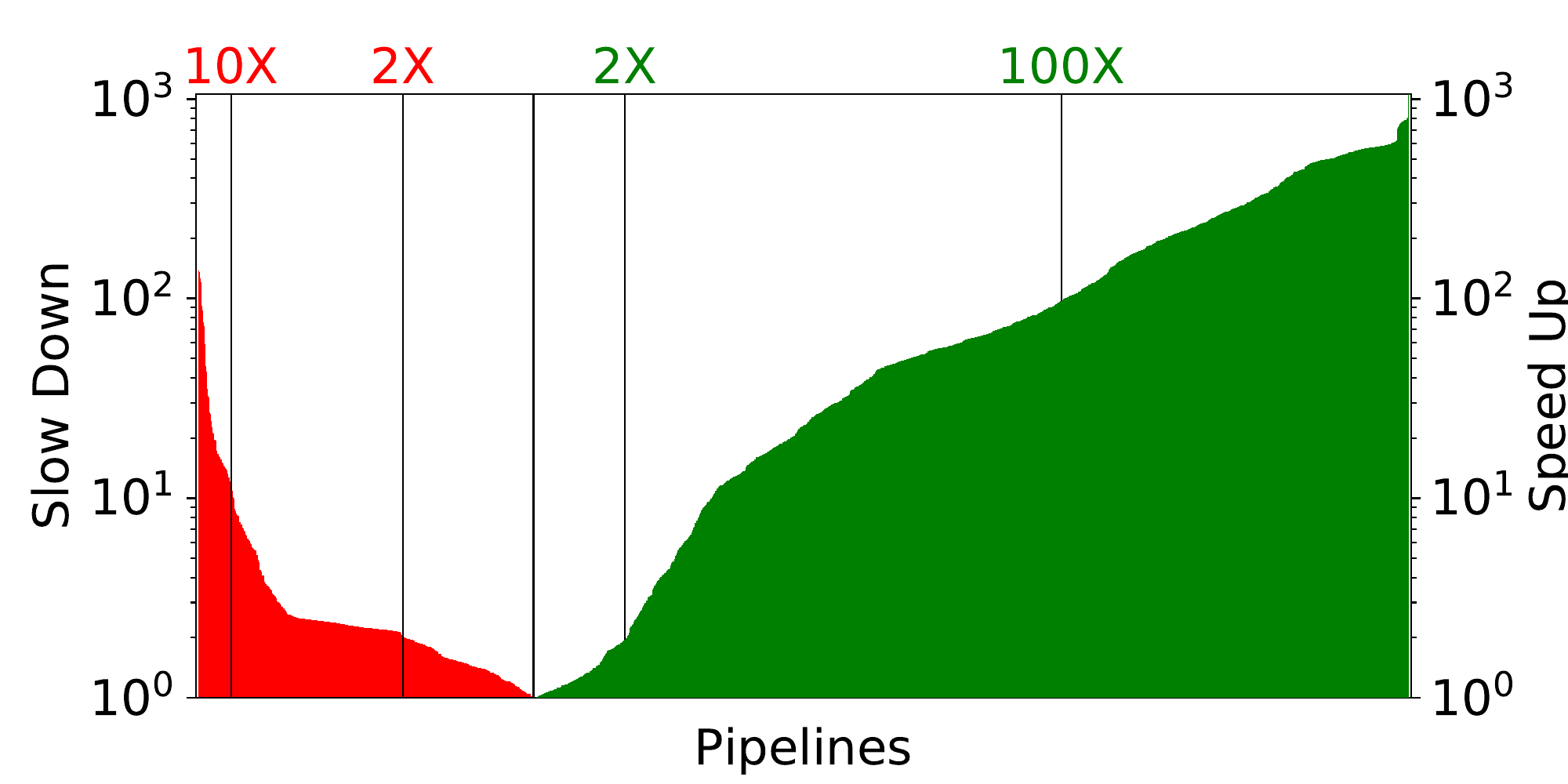}
        	\vspace{-4ex}
        	\caption{GPU}
        	\label{fig:pip-ts-gpu}
        \end{subfigure}
        
        \vspace{-2ex}
         \captionof{figure}{Speedup/slowdown of pipelines when using \system wrt baseline Sklearn.}
         \label{fig:pip-ts}
         \vspace{-3ex}
     \end{minipage}
\vspace{-2ex}
\end{figure*}
 
\vspace{-2ex}
\subsection{Optimizations}
\vspace{-1.5mm}
\subsubsection{Tree Models Implementation}
\label{sec:optimizations}
\vspace{-2mm}
Next we test the different tree-based models implementation to make the case for the heuristics.

\stitle{Datasets.} For this experiment we employ a synthetic dataset  randomly generated with 5000 rows and 200 features. \eat{The second is the {\sf nomao} dataset from OpenML~\cite{nomao}. This dataset contains about 35k rows and 119 features: 89 numerical and 30 categorical.}

\stitle{Experiments Setup.} 
We study the behavior of the tree implementations as we change the training algorithm, the batch size, and the tree depth. For each experiment we set the number of trees to 100.
We use the TVM runtime backend. 
\eat{
In the second and third experiments instead we evaluate the performance of the feature selection push down, and the feature injection optimizations. 
Both experiments will use the {\sf nomao} dataset, the TorchScript backend, and a similar pipeline composed of: (1) a  simple imputer (applied over the numerical features); (2) a one-hot encoder (applied to the categorical features); (3) a scaler normalizing the features output by both the imputer and the encoder; and (4) a final logistic regression. 
For the push-down experiment, the pipeline will additionally have a feature selection operator before the logistic regression, and we will plot the execution time as we vary the percentage of feature selection percentage. 
For the injection experiment, we will use a L1 regularizer for the logistic regression, and we vary the coefficient.}
Each experiment is run on 1 CPU core.

\stitle{Results.}
Figure~\ref{fig:opt-tree} shows the comparison between the different tree implementations, and the two scikit-learn and ONNX-ML baselines.
In the top part of the figure we run all experiments using a batch size of 1; on the bottom part we instead use a batch size of 1$K$. In the column on the left-hand side, we generate trees with a max depth of 3; 7 for the middle column, and 12 for column on the right-hand side.
In general, two things are apparent: (1) \system is as fast as or better than the baselines; and (2) no tree implementation is always better than the others.
The \at{GEMM} implementation outperforms the other two for small batch sizes, whereas {\sf TT} and {\sf PTT} are better over larger batch sizes.
Between {\sf TT} and {\sf PTT}, the latter is usually the best performant (although not by a large margin). {\sf PTT} however creates balanced trees, and fails for very deep trees.

\vspace{-2ex}
\subsubsection{Runtime-independent Optimizations.}
\label{sec:opt-runtime-independent}
\vspace{-1.5ex}

Next we test the optimizations described in Section~\ref{sec:runtime-independent}.

\stitle{Dataset.} We use the Nomao dataset~\cite{nomao} with 119 features.

\stitle{Feature Selection Push Down.}
In this experiment we measure the benefits of the feature selection push down. In Figure~\ref{fig:opt-fs-pushdown} we compare \system with and without feature selection push-down, and the baseline implementation of the pipelines in scikit-learn. We use a pipeline which trains a logistic regression model with L2 loss.
The featurization part contains one-hot encoding for categorical features, missing value imputation for numerical values, followed by feature scaling, and a final feature selection operator (scikit-learn's \texttt{SelectKBest}). We vary the percentile of features that are picked by the feature selection operator. 
In general, we can see that \system without optimization is about 2$\times$ faster than scikit-learn in evaluating the pipelines. For small percentiles, the feature selection push-down optimization delivers a further 3$\times$. As we increase the percentile of features that are selected, the runtime of \system both with and without optimizations increase, although with the optimization \system is still 2$\times$ faster than without.

\stitle{Feature Selection Injection.}
In this experiment we evaluate whether we can improve the performance of pipelines with sparse models by injecting (and then pushing down) feature selection operators. 
The pipeline is the same as in the previous case but without the feature selection operator.
Instead we train the logistic regression model with L1 regularization.
In Figure~\ref{fig:opt-fs-injection} we vary the L1 regularization coefficient and study how much performance we can gain. 
Also in this case, with very sparse models we can see up to 3$\times$ improvement wrt \system without optimization. Performance gains dissipate as we decrease the sparsity of the model.

\vspace{-2ex}
\subsection{End-to-end Pipelines}
\label{sec:end-to-end}
\vspace{-2ex}
\stitle{Setup.} In this experiment we test \system over end-to-end pipelines. We downloaded the 72 tasks composing the OpenML-CC18 suite~\cite{openml-cc18}. 
Among all the tasks, we discarded all the ``not pure scikit-learn'' ML pipelines 
(e.g., containing also arbitrary Python code).
We successively discarded all the pipelines returning a failure during training. 88\% of the remaining pipelines are exclusively composed by operators supported by \system, for a total of 2328 ML pipelines.
Among these, 11 failed during inference due to runtime errors in \system; we report the summary of executing 2317 pipelines.
These pipelines contain an average of 3.3 operators, which is in line with what was observed elsewhere~\cite{dsonds}.

\stitle{Datasets.} For this experiment we have 72 datasets in total~\cite{openml-cc18}. The datasets are a curated mix specifically designed for ML benchmarking. We did the typical 80\%/20\% split between training and inference. The smaller dataset has just 100 records, the bigger 19264, while the median value is 462. The minimum number of columns for a dataset is 4, the maximum 3072, with a median of 30. 


\stitle{Results.} 
Figure~\ref{fig:pip-ts} summarizes the speedup / slowdown introduced by \system when scoring all 2317 pipelines. As we can see, \system is able to accelerate about 60\% of the pipelines on CPU (\ref{fig:pip-ts-cpu}). In general, the slowest pipeline gets about 60$\times$ slower wrt scikit-learn, the fastest instead gets a 1200$\times$ speed up.
The slowdowns are due to a couple of factors: (a) the datasets used for these experiments are quite small; (b) some pipelines contain largely sparse operations (i.e., SVM on sparse inputs);
(c) several pipelines are small and do not require much computation (e.g., a simple inputer followed by a small decision tree). These three factors are highlighted also by the fact that even if we move computation to the GPU (\ref{fig:pip-ts-gpu}), still 27\% of the pipelines have some slowdown.
Note however that (1) both sparse and small pipelines can be detected at compile time, and therefore we can return a warning or an error; (2) DNN frameworks are continuously adding new sparse tensor operations (e.g.,~\cite{sparse-tensor-pytorch}); and (3) an option could be to add a specific runtime backend for sparse tensor operations (e.g., we have a prototype integration with TACO~\cite{taco}). 
In general, DNN frameworks are relatively young, and \system will exploit any future improvement with no additional costs.

With GPU acceleration (Figure~\ref{fig:pip-ts-gpu}), 73\% of the pipelines show some speedup. The slowest pipeline gets about 130$\times$ slower wrt scikit-learn, the fastest instead gets a speedup of 3 orders of magnitude.
Some of the pipelines get worse from CPU to GPU execution. This is due to (1) sparsity; (2) small compute; and (3) data movements between CPU and GPU memory. Indeed we run all pipelines on GPU, even the ones for which in practice would not make much sense (e.g., a decision tree with 3 nodes). We leave as future work an extension to our heuristics for picking the right hardware backend. 

\eat{
\begin{itemize}
    \item \stitle{Setup.} Baselines: sklearn OpenML-CC18. Hardware: P100 machine, all cores. Backends: TorchScript, TVM. 
    \item report speedup for batch
    \item report speed for gpu
\end{itemize}
}
    
\eat{
\begin{figure*}  
\vspace{-2ex}
	\centering
	\hspace{-4ex}
	\begin{subfigure}{0.35\textwidth}
		\centering
		\includegraphics[trim={0 0 1.3cm 1.1cm},clip,width=\textwidth]{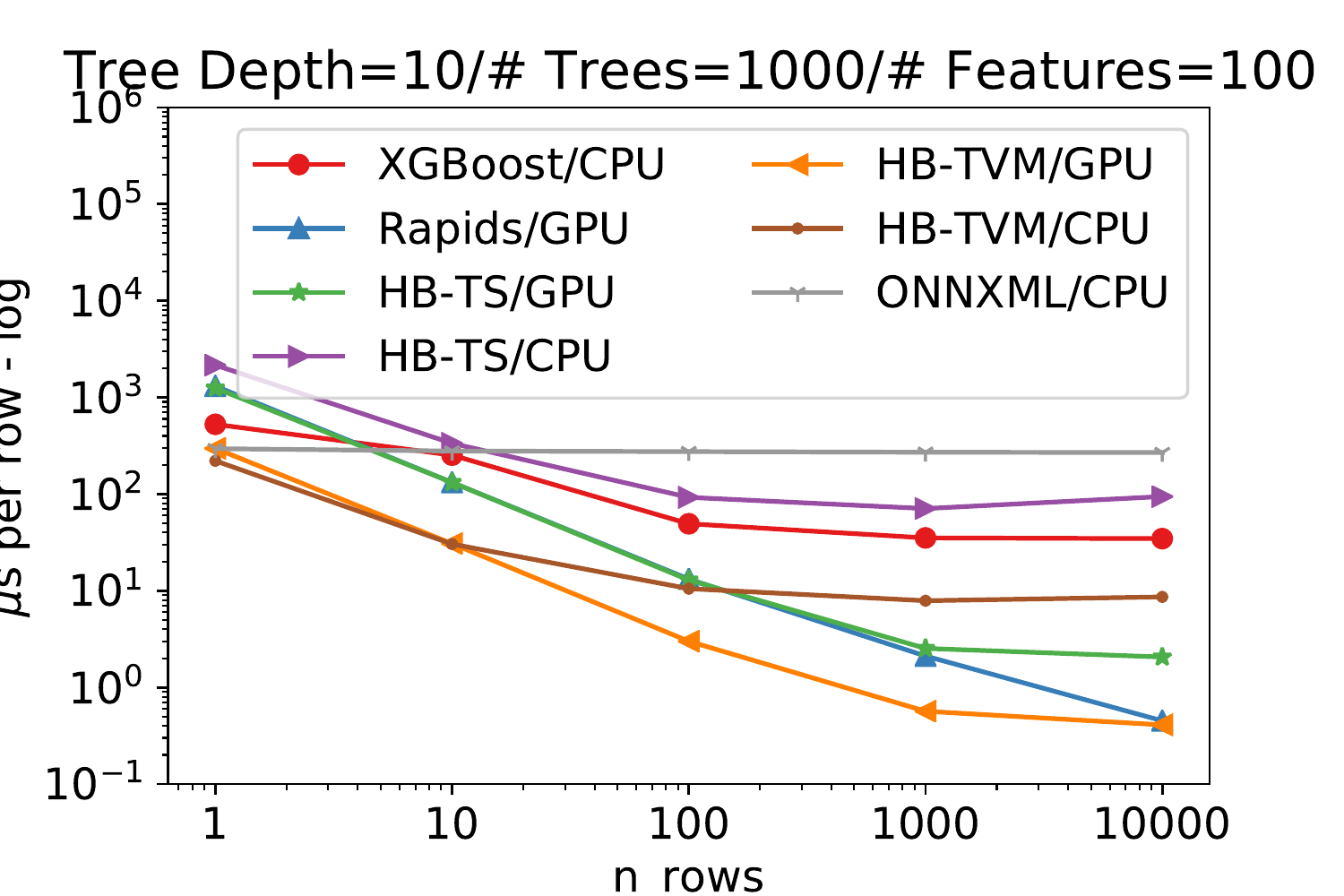}
		\caption{Micro-benchmark}
        \label{fig:mini}
	\end{subfigure}
	\hspace{-1ex}
	\begin{subfigure}{0.33\textwidth}
		\centering
		\includegraphics[trim={0.5cm 0 1.3cm 1.1cm},clip,width=\textwidth]{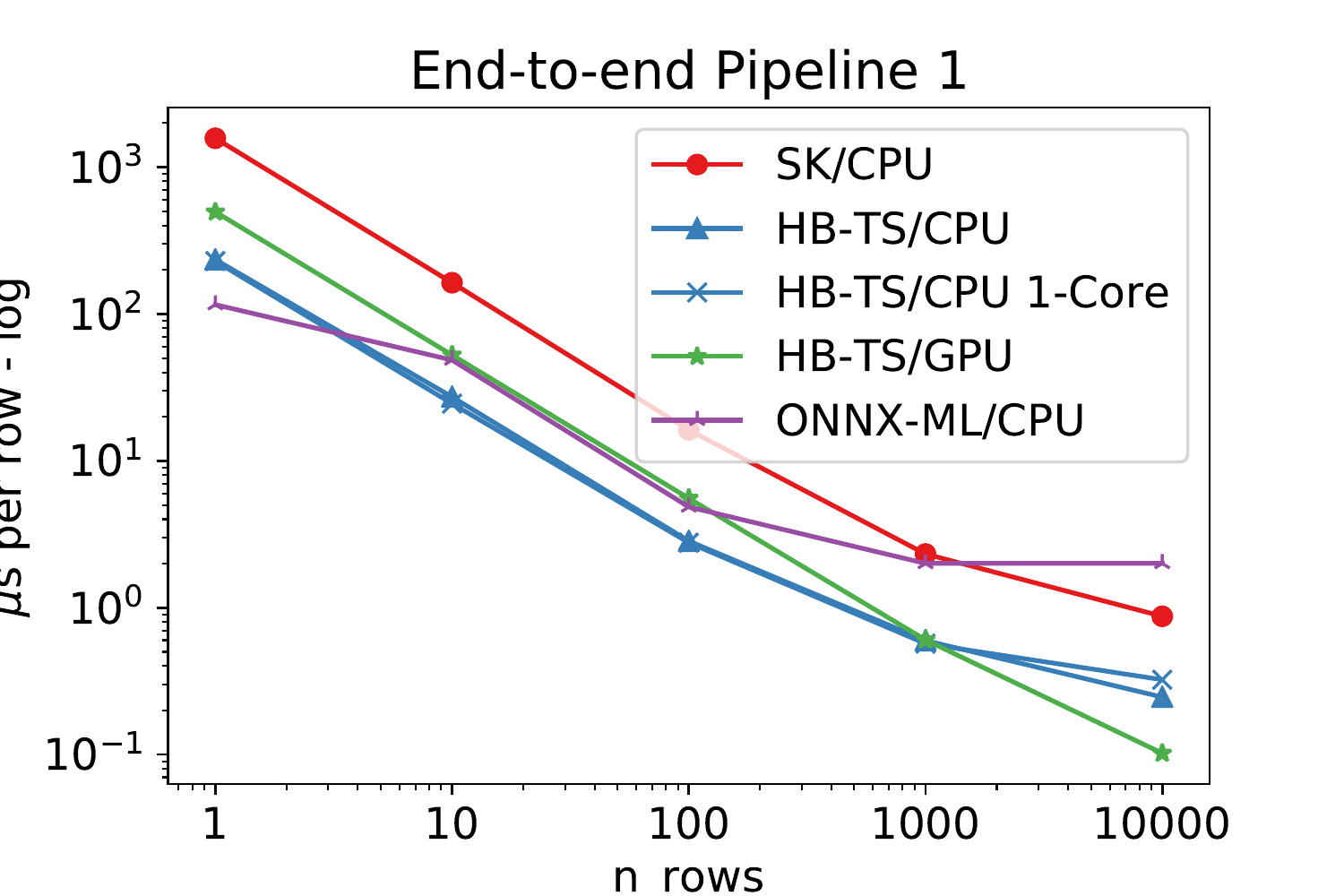}
		\caption{Pipeline 1}
		\label{fig:e2e1}
	\end{subfigure}
	\hspace{-1ex}
	\begin{subfigure}{0.33\textwidth}
		\centering
		\includegraphics[trim={0.8cm 0 1.3cm 1.1cm},clip,width=\textwidth]{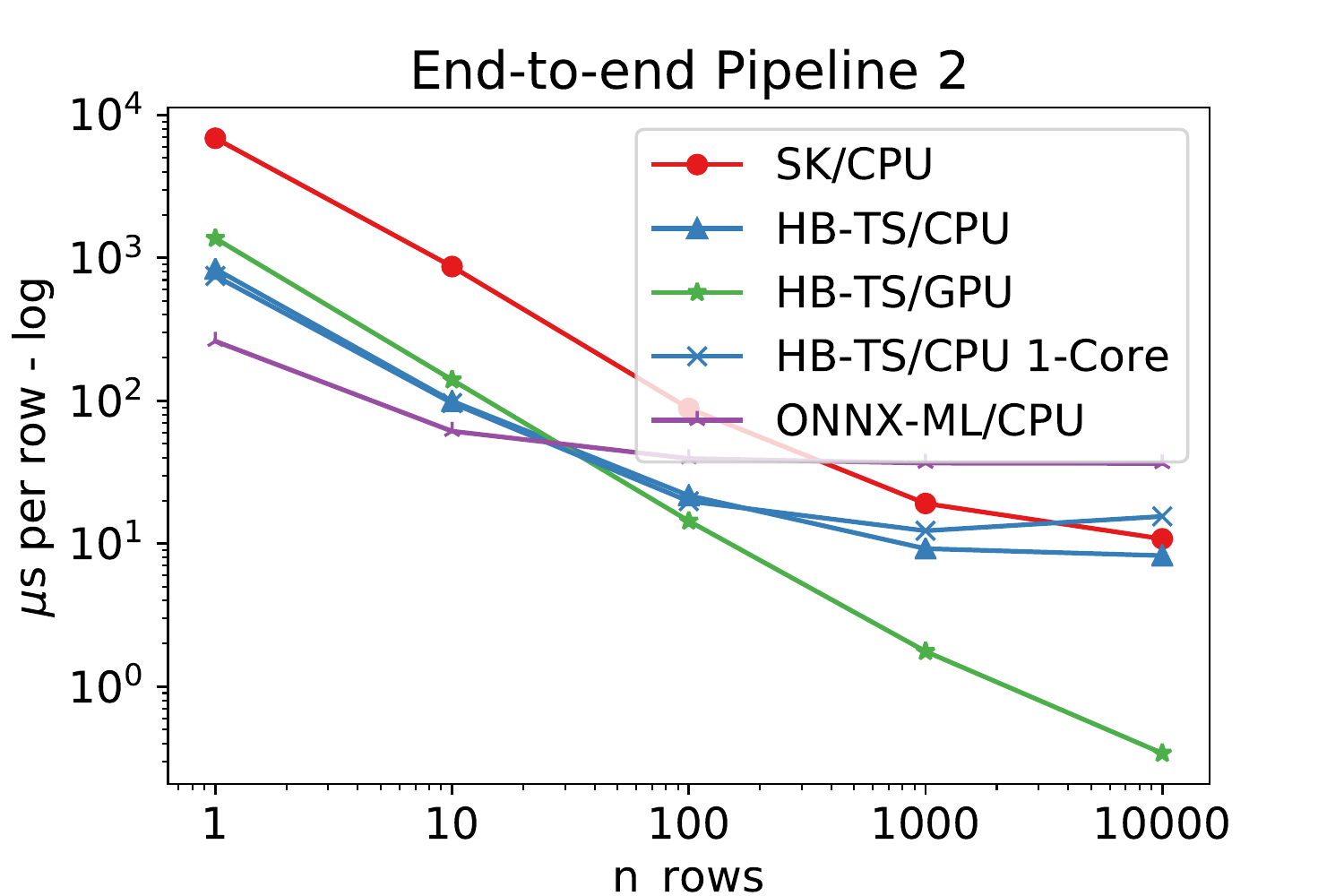}
		\caption{Pipeline 2}
		\label{fig:e2e2}
	\end{subfigure}
	\vspace{-1ex}
 	\caption{\system performance evaluation.}
 	\vspace{-3ex}
\end{figure*}

\eat{
    \begin{figure*}[ht]
    \includegraphics[width=\textwidth]{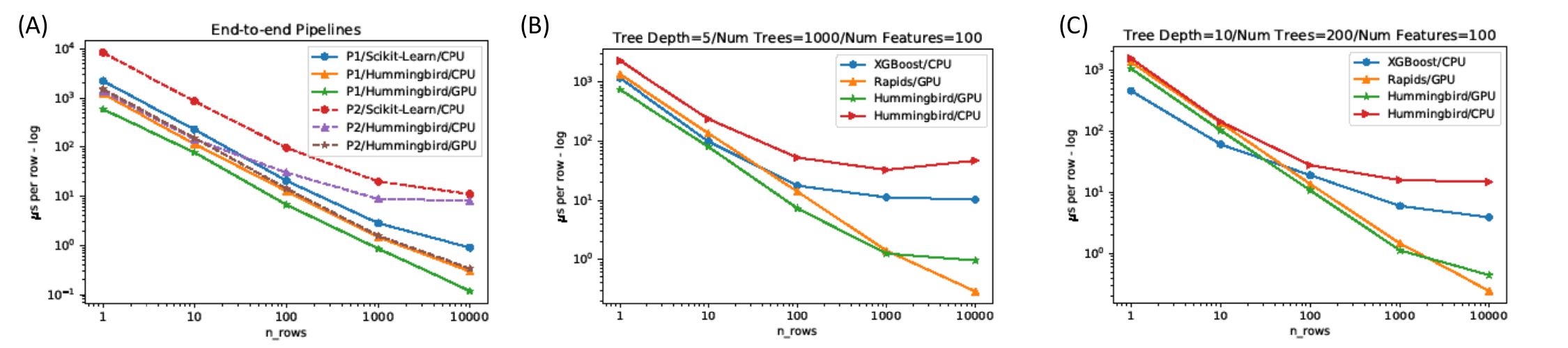}
    \caption{Experimental Results}
    \label{fig:results}
    \vspace{-2mm}
    \end{figure*}
}

\eat{\begin{figure*}[ht]
    \includegraphics[width=\textwidth]{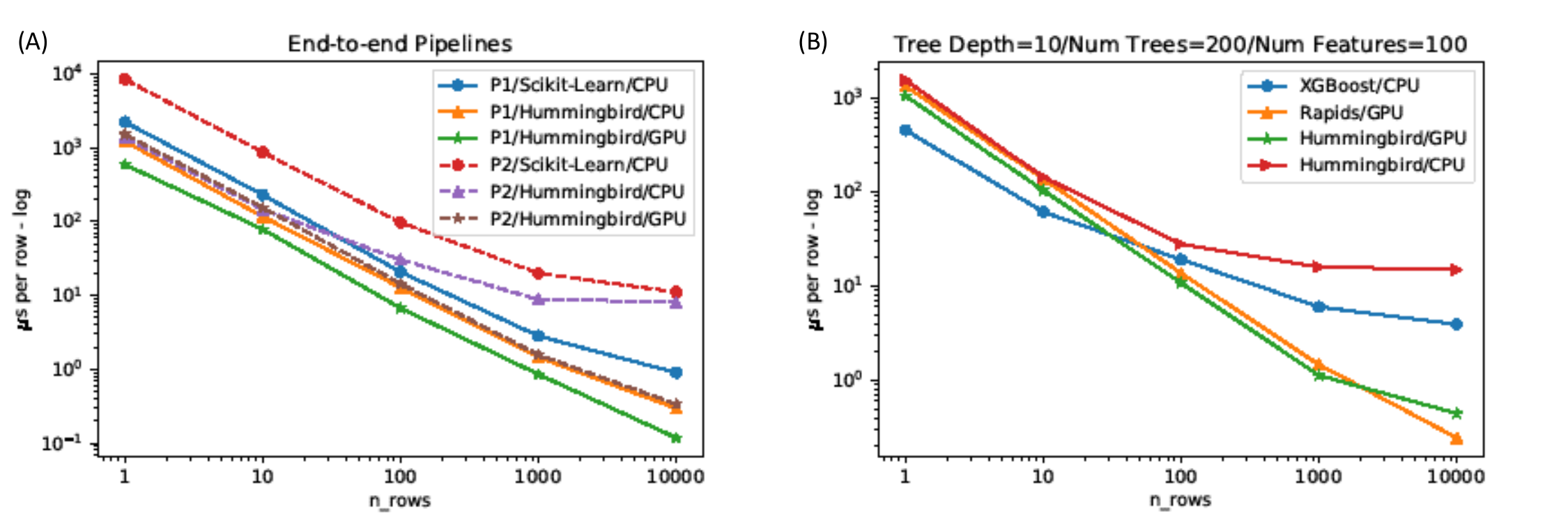}
    \caption{Experimental Results}
    \label{fig:results}
    \vspace{-2mm}
    \end{figure*}
}

\stitle{Gradient Boosting Micro-benchmark.} We first run a micro-benchmark to evaluate the performance of translating tree-based models. We trained a XGBoost~\cite{xgboost} model. We evaluate \system (strategy 3) with TVM ({\sf HB-TVM}) and PyTorch~\footnote{For the PyTorch backend, we compile models into TorchScript for better performance.} ({\sf HB-TS}) backends against three baselines: XGBoost, ONNX-ML~\cite{onnx-ml}, and Nvidia Rapids Forest Inference Library (FIL)~\cite{fil}.
We use a synthetic dataset which has 100 features and 10,000 records.
We vary the inference batch size and use 1000 trees with max tree depth set to 10. 
From the results we can see that {\sf HB-TVM} outperforms the baselines (for batch size of $10K$  we are on par with Rapids) both on CPU and GPU. \system with PyTorch back-end is on par with the natives CPU and GPU implementations for small batch sizes, whereas it is 3$\times$ to 5$\times$ slower for larger batches. This is mainly due to TVM ability to fuse operators into efficient kernels and scheduling tuning.
Note that ONNX-ML's scaling is linear here, as its current implementation of tree inference is neither optimized for batch or multicore execution.

\stitle{End-to-end Pipeline Evaluation.} For this set of experiments, we use the two pipelines shown in \autoref{fig:pipeline} (A), and train them on the Forest Covertype dataset ~\cite{fcovtype}. For the feature selector operator we set the number of features to $15$. For the random forest model we use 100 trees with max depth set to 10.
We compare inference performance for the original pipeline implementations in scikit-Learn and ONNX-ML against \system with PyTorch backed on varying batch sizes. The results are shown in \autoref{fig:e2e1} and c. 
In both pipelines,: (1) ONNX-ML performs best for batch of size but then it is not able to take advantage of bigger batch sizes; (2) \system always performs better than scikit-learn, while it performs better than ONNX-ML for large batches; (3) multicore execution do not improve performance for \system wrt single core; and (4) GPU execution provides better performance for large batches, up to 15$\times$ for pipeline 2. 
}

\vspace{-2ex}
\section{Related Work}
\label{s:related}
\vspace{-2ex}

PyTorch~\cite{pytorch}, TensorFlow~\cite{tensorflow}, MXNet~\cite{mxnet}, CNTK~\cite{cntk} are DNN frameworks that provide easy-to-use (tensor-based) APIs for authoring DNN models, and heterogeneous hardware support for both training and inference.
Beyond these popular frameworks, inference runtimes such as ONNX~\cite{onnx-runtime}, nGraph~\cite{ngraph}, TVM~\cite{tvm}, and  TensorRT~\cite{tensor-rt} provide optimizations and efficient execution targets, specifically for inference. To prove the versatility of our approach, we have tested \system with both PyTorch and TVM. 
\system uses a two-level, logical-physical optimization approach. 
First, logical optimizations are applied based on the operators composing the pipeline. Afterwards, physical operator implementations are selected based on model statistics, and physical rewrites, which are externally implemented by the DNN runtime, are executed (e.g., algebraic rewrites, operator fusion).
Willump~\cite{willump} uses a similar two-level optimization strategy, although it targets Weld~\cite{weld} as its low level runtime and therefore it cannot natively support inference on hardware accelerators.
Conversely, \system casts ML pipelines into tensor computations
and takes advantage of DNN serving systems to ease the deployment on target environments.
Other optimizers for predictive  pipelines, such as Pretzel~\cite{pretzel}, only target logical optimizations. 
We have integrated \system into Raven~\cite{raven} as part of our bigger vision for optimizing ML prediction pipelines.

\eat{
logical optimizations and operator implementations are  at the upper level. Successively, lower-level optimizations such as algebraic rewrites and operator fusion provided by the target runtime are executed. 
Conversely, other inference pipelines optimizers such as Pretzel~\cite{pretzel} only target higher level optimizations.
Willump~\cite{willump} uses a similar approach, although its low level target is Weld~\cite{weld} and therefore it cannot natively support inference on hardware accelerators. }

Several works deal with executing trees (ensembles)~\cite{fil, tree-fpga, tree-gpu} on hardware accelerators.
These systems provide a custom implementation of the \at{PTT} strategy specific to the target hardware (e.g., NVIDIA GPUs for RAPIDS FIL~\cite{fil}, FPGAs for~\cite{tree-fpga}), and where computation is parallelized along on the tree-dimension. 
Alternatively, \system provides three tree inference strategies, including two novel strategies (\at{GEMM} and \at{TT}), and picks the best alternative based on the efficiency and redundancy trade-off.
\eat{Finally, while this work only covers inference, similar techniques can be used for training as well~\cite{hb-training}.
\lsystem can be used alone, but also integrated with other optimizers, e.g., Raven~\cite{raven}.
}



\eat{
Several works~\cite{finetune,Banerjee94initializingneural,Ivanova95initializationof,8478232} suggested to compile tree models into MLPs, although only for training. We instead explore how MLPs can be used to score tree models using NN inference systems.
}

\vspace{-2ex}
\section{Conclusions}
\label{s:conclusion}
\vspace{-2ex}

In this paper, we explore the idea of using DNN frameworks as generic compilers and optimizers for heterogeneous hardware.
Our use-case is ``traditional'' ML inference. 
We ported 40+ data featurizers and traditional ML models into tensor operations and tested their performance over two DNN frameworks (PyTorch and TVM) and over different hardware (CPUs and GPUs).  
The results are compelling: even though we target high-level tensor operations, we are able to outperform custom C++ and CUDA implementations.
To our knowledge, \lsystem is the first system able to run traditional ML inference on heterogeneous hardware. 


\vspace{-2ex}
\section{Acknowledgements}\vspace{-2ex}
We thank the anonymous reviewers and our shepherd, Chen Wenguang, for their feedback and suggestions to improve the paper. We would also like to thank Nellie Gustafsson, Gopal Vashishtha, Emma Ning, and Faith Xu for their support.


{\small
\bibliographystyle{plain}
\bibliography{main}}

\begin{thebibliography}{10}

\bibitem{celebras}
{Cerebras Chip}.
\newblock
  \url{https://www.wired.com/story/power-ai-startup-built-really-big-chip/}.

\bibitem{mem_profiler}
{Memory profiler for Python}.

\bibitem{rapids}
{Nvidia RAPIDS}.
\newblock \url{https://developer.nvidia.com/rapids}.

\bibitem{onnx-ml}
{ONNX ML}.
\newblock \url{https://github.com/onnx/onnx/blob/master/docs/Operators-ml.md}.

\bibitem{onnx-runtime}
{ONNX Runtime}.
\newblock \url{https://github.com/microsoft/onnxruntime}.

\bibitem{onnx-supported}
{ONNX Supported Frameworks and Backends}.
\newblock \url{https://onnx.ai/supported-tools.html}.

\bibitem{pandas}
{Pandas}.
\newblock \url{https://pandas.pydata.org/}.

\bibitem{torchscript}
{TorchScript Documentation}.
\newblock \url{https://pytorch.org/docs/stable/jit.html}.

\bibitem{h2o}
H2{O} {A}lgorithms {R}oadmap.
\newblock
  \url{https://github.com/h2oai/h2o-3/blob/master/h2o-docs/src/product/flow/images/H2O-Algorithms-Road-Map.pdf},
  2015.

\bibitem{cntk}
C{NTK}.
\newblock \url{https://docs.microsoft.com/en-us/cognitive-toolkit/}, 2018.

\bibitem{matplot}
Matplotlib.
\newblock \url{https://matplotlib.org/}, 2018.

\bibitem{mxnet}
M{XN}et.
\newblock \url{https://mxnet.apache.org/}, 2018.

\bibitem{tensorflow}
Tensor{F}low.
\newblock \url{https://www.tensorflow.org}, 2018.

\bibitem{ai-cost}
Esg technical validation: Dell emc ready solutions for ai: Deep learning with
  intel.
\newblock
  \\\url{https://www.esg-global.com/validation/esg-technical-validation-dell-emc-ready-solutions-for-ai-deep-learning-with-intel},
  2019.

\bibitem{graphcore}
{Graphcore IPU}.
\newblock \url{https://www.graphcore.ai/}, 2019.

\bibitem{ngraph}
{nGraph}.
\newblock \url{https://www.ngraph.ai/}, 2019.

\bibitem{onnx-bench}
{{ONNX-ML vs Sklearn Benchmark}}.
\newblock \url{https://github.com/xadupre/scikit-learn_benchmarks}, 2019.

\bibitem{onnxmltools}
{ONNXMLTools}.
\newblock \url{https://github.com/onnx/onnxmltools}, 2019.

\bibitem{fil-blog}
{RAPIDS Forest Inference Library}.
\newblock
  \url{https://medium.com/rapids-ai/rapids-forest-inference-library-prediction-at-100-million-rows}
  \url{-per-second-19558890bc35}, 2019.

\bibitem{tensor-rt}
{Tensor-RT}.
\newblock \url{https://developer.nvidia.com/tensorrt}, 2019.

\bibitem{broadcasting}
{Broadcasting Semantic}.
\newblock \url{https://www.tensorflow.org/xla/broadcasting}, 2020.

\bibitem{gbm-bench}
{Gradient Boosting Algorithm Benchmark}.
\newblock \url{https://github.com/NVIDIA/gbm-bench}, 2020.

\bibitem{iris}
{Iris dataset}.
\newblock
  \url{https://scikit-learn.org/stable/auto_examples/datasets/plot_iris_dataset.html},
  2020.

\bibitem{nomao}
{nomao dataset}.
\newblock \url{https://www.openml.org/d/1486}, 2020.

\bibitem{onnx}
{{ONNX}}.
\newblock \url{https://github.com/onnx/onnx/blob/master/docs/Operators.md},
  2020.

\bibitem{mobile}
{ONNX Portable format}.
\newblock
  \url{https://www.infoworld.com/article/3223401/onnx-makes-machine-learning-models-portable-shareable.html},
  2020.

\bibitem{openml-cc18}
{{OpenML-CC18 Benchmark}}.
\newblock \url{https://www.openml.org/s/99}, 2020.

\bibitem{pytorch-ecosystem}
{Pytorch Ecosystem}.
\newblock \url{https://pytorch.org/ecosystem/}, 2020.

\bibitem{fil}
{{RAPIDS cuML}}.
\newblock \url{https://github.com/rapidsai/cuml}, 2020.

\bibitem{sambanova}
{{Sambanova: Massive Models for Everyone}}.
\newblock \url{https://sambanova.ai/}, 2020.

\bibitem{skl2onnx}
{skl2onnx Converter}.
\newblock \url{https://github.com/onnx/sklearn-onnx/}, 2020.

\bibitem{tensorflowjs}
{Tensorflow JS}.
\newblock \url{https://www.tensorflow.org/js}, 2020.

\bibitem{xla}
{Tensorflow XLA}.
\newblock \url{https://www.tensorflow.org/xla}, 2020.

\bibitem{sparse-tensor-pytorch}
{The Status of Sparse Operations in Pytorch}.
\newblock \url{https://github.com/pytorch/pytorch/issues/9674}, 2020.

\bibitem{flock}
Ashvin {Agrawal}, Rony {Chatterjee}, Carlo {Curino}, Avrilia {Floratou}, Neha
  {Gowdal}, Matteo {Interlandi}, Alekh {Jindal}, Kostantinos {Karanasos}, Subru
  {Krishnan}, Brian {Kroth}, Jyoti {Leeka}, Kwanghyun {Park}, Hiren {Patel},
  Olga {Poppe}, Fotis {Psallidas}, Raghu {Ramakrishnan}, Abhishek {Roy}, Karla
  {Saur}, Rathijit {Sen}, Markus {Weimer}, Travis {Wright}, and Yiwen {Zhu}.
\newblock {Cloudy with high chance of DBMS: A 10-year prediction for
  Enterprise-Grade ML}.
\newblock {\em arXiv e-prints}, page arXiv:1909.00084, Aug 2019.

\bibitem{mldotnet}
Zeeshan Ahmed, Saeed Amizadeh, Mikhail Bilenko, Rogan Carr, Wei-Sheng Chin,
  Yael Dekel, Xavier Dupre, Vadim Eksarevskiy, Senja Filipi, Tom Finley, et~al.
\newblock Machine learning at {Microsoft} with {ML.NET}.
\newblock In {\em Proceedings of the 25th ACM SIGKDD International Conference
  on Knowledge Discovery and Data Mining}, KDD ’19, page 2448–2458, New
  York, NY, USA, 2019. Association for Computing Machinery.

\bibitem{euclidean_distance_trick}
Samuel Albanie.
\newblock Euclidean distance matrix trick.
\newblock 2019.

\bibitem{sagemaker-tco}
Amazon.
\newblock The total cost of ownership (tco) of amazon sagemaker.
\newblock
  \url{https://pages.awscloud.com/rs/112-TZM-766/images/Amazon_SageMaker_TCO_uf.pdf},
  2020.

\bibitem{tfx}
Denis Baylor, Eric Breck, Heng-Tze Cheng, Noah Fiedel, Chuan~Yu Foo, Zakaria
  Haque, Salem Haykal, Mustafa Ispir, Vihan Jain, Levent Koc, Chiu~Yuen Koo,
  Lukasz Lew, Clemens Mewald, Akshay~Naresh Modi, Neoklis Polyzotis, Sukriti
  Ramesh, Sudip Roy, Steven~Euijong Whang, Martin Wicke, Jarek Wilkiewicz, Xin
  Zhang, and Martin Zinkevich.
\newblock Tfx: A tensorflow-based production-scale machine learning platform.
\newblock In {\em Proceedings of the 23rd ACM SIGKDD International Conference
  on Knowledge Discovery and Data Mining}, KDD ’17, page 1387–1395, New
  York, NY, USA, 2017. Association for Computing Machinery.

\bibitem{xgboost}
Tianqi Chen and Carlos Guestrin.
\newblock Xgboost: A scalable tree boosting system.
\newblock In {\em Proceedings of the 22Nd ACM SIGKDD International Conference
  on Knowledge Discovery and Data Mining}, KDD '16, pages 785--794, New York,
  NY, USA, 2016. ACM.

\bibitem{tvm}
Tianqi Chen, Thierry Moreau, Ziheng Jiang, Lianmin Zheng, Eddie Yan, Meghan
  Cowan, Haichen Shen, Leyuan Wang, Yuwei Hu, Luis Ceze, Carlos Guestrin, and
  Arvind Krishnamurthy.
\newblock Tvm: An automated end-to-end optimizing compiler for deep learning.
\newblock In {\em Proceedings of the 12th USENIX Conference on Operating
  Systems Design and Implementation}, OSDI'18, pages 579--594, Berkeley, CA,
  USA, 2018. USENIX Association.

\bibitem{inferline}
Daniel Crankshaw, Gur{-}Eyal Sela, Corey Zumar, Xiangxi Mo, Joseph~E. Gonzalez,
  Ion Stoica, and Alexey Tumanov.
\newblock Inferline: {ML} inference pipeline composition framework.
\newblock {\em CoRR}, abs/1812.01776, 2018.

\bibitem{clipper2}
Daniel Crankshaw, Xin Wang, Guilio Zhou, Michael~J. Franklin, Joseph~E.
  Gonzalez, and Ion Stoica.
\newblock Clipper: {A} low-latency online prediction serving system.
\newblock In {\em {NSDI}}, 2017.

\bibitem{feature_selection}
Manoranjan Dash and Huan Liu.
\newblock Feature selection for classification.
\newblock {\em Intelligent data analysis}, 1(3):131--156, 1997.

\bibitem{bert}
Jacob {Devlin}, Ming-Wei {Chang}, Kenton {Lee}, and Kristina {Toutanova}.
\newblock {BERT: Pre-training of Deep Bidirectional Transformers for Language
  Understanding}.
\newblock {\em arXiv e-prints}, page arXiv:1810.04805, Oct 2018.

\bibitem{firmai}
{FirmAI}.
\newblock {Machine Learning and Data Science Applications in Industry}.
\newblock \url{https://github.com/firmai/industry-machine-learning}.

\bibitem{dl-book}
Ian Goodfellow, Yoshua Bengio, and Aaron Courville.
\newblock {\em Deep learning}.
\newblock MIT press, 2016.

\bibitem{fpga}
Intel.
\newblock Machine learning fpga.
\newblock
  \url{https://www.intel.com/content/www/us/en/products/docs/storage/programmable/applications/machine-learning.html},
  2020.

\bibitem{tpu}
Norman~P. Jouppi, Cliff Young, Nishant Patil, David~A. Patterson, Gaurav
  Agrawal, Raminder Bajwa, Sarah Bates, Suresh Bhatia, Nan Boden, Al~Borchers,
  Rick Boyle, Pierre{-}luc Cantin, Clifford Chao, Chris Clark, Jeremy Coriell,
  Mike Daley, Matt Dau, Jeffrey Dean, Ben Gelb, Tara~Vazir Ghaemmaghami,
  Rajendra Gottipati, William Gulland, Robert Hagmann, Richard~C. Ho, Doug
  Hogberg, John Hu, Robert Hundt, Dan Hurt, Julian Ibarz, Aaron Jaffey, Alek
  Jaworski, Alexander Kaplan, Harshit Khaitan, Andy Koch, Naveen Kumar, Steve
  Lacy, James Laudon, James Law, Diemthu Le, Chris Leary, Zhuyuan Liu, Kyle
  Lucke, Alan Lundin, Gordon MacKean, Adriana Maggiore, Maire Mahony, Kieran
  Miller, Rahul Nagarajan, Ravi Narayanaswami, Ray Ni, Kathy Nix, Thomas
  Norrie, Mark Omernick, Narayana Penukonda, Andy Phelps, Jonathan Ross, Amir
  Salek, Emad Samadiani, Chris Severn, Gregory Sizikov, Matthew Snelham, Jed
  Souter, Dan Steinberg, Andy Swing, Mercedes Tan, Gregory Thorson, Bo~Tian,
  Horia Toma, Erick Tuttle, Vijay Vasudevan, Richard Walter, Walter Wang, Eric
  Wilcox, and Doe~Hyun Yoon.
\newblock In-datacenter performance analysis of a tensor processing unit.
\newblock {\em CoRR}, abs/1704.04760, 2017.

\bibitem{raven}
Konstantinos Karanasos, Matteo Interlandi, Fotis Psallidas, Rathijit Sen,
  Kwanghyun Park, Ivan Popivanov, Doris Xin, Supun Nakandala, Subru Krishnan,
  Markus Weimer, Yuan Yu, Raghu Ramakrishnan, and Carlo Curino.
\newblock Extending relational query processing with {ML} inference.
\newblock In {\em {CIDR} 2020, 10th Conference on Innovative Data Systems
  Research, Amsterdam, The Netherlands, January 12-15, 2020, Online
  Proceedings}. www.cidrdb.org, 2020.

\bibitem{lgbm}
Guolin Ke, Qi~Meng, Thomas Finley, Taifeng Wang, Wei Chen, Weidong Ma, Qiwei
  Ye, and Tie-Yan Liu.
\newblock Lightgbm: A highly efficient gradient boosting decision tree.
\newblock In I.~Guyon, U.~V. Luxburg, S.~Bengio, H.~Wallach, R.~Fergus,
  S.~Vishwanathan, and R.~Garnett, editors, {\em Advances in Neural Information
  Processing Systems 30}, pages 3146--3154. Curran Associates, Inc., 2017.

\bibitem{taco}
Fredrik Kjolstad, Shoaib Kamil, Stephen Chou, David Lugato, and Saman
  Amarasinghe.
\newblock The tensor algebra compiler.
\newblock {\em Proc. ACM Program. Lang.}, 1(OOPSLA):77:1--77:29, October 2017.

\bibitem{willump}
Peter {Kraft}, Daniel {Kang}, Deepak {Narayanan}, Shoumik {Palkar}, Peter
  {Bailis}, and Matei {Zaharia}.
\newblock {Willump: A Statistically-Aware End-to-end Optimizer for Machine
  Learning Inference}.
\newblock {\em arXiv e-prints}, page arXiv:1906.01974, Jun 2019.

\bibitem{alexnet}
Alex Krizhevsky, Ilya Sutskever, and Geoffrey~E. Hinton.
\newblock Imagenet classification with deep convolutional neural networks.
\newblock {\em Commun. ACM}, 60(6):84–90, May 2017.

\bibitem{pretzel}
Yunseong Lee, Alberto Scolari, Byung-Gon Chun, Marco~Domenico Santambrogio,
  Markus Weimer, and Matteo Interlandi.
\newblock {PRETZEL}: Opening the black box of machine learning prediction
  serving systems.
\newblock In {\em 13th {USENIX} Symposium on Operating Systems Design and
  Implementation ({OSDI} 18)}, pages 611--626, Carlsbad, CA, October 2018.
  {USENIX} Association.

\bibitem{tradMLtraining}
Ping Li.
\newblock Robust logitboost and adaptive base class (abc) logitboost.
\newblock In {\em n Proceedings of the Twenty-Sixth Conference Annual
  Conference on Uncertainty in Artificial Intelligence (UAI’10)}.

\bibitem{distributed-pytorch}
Shen Li, Yanli Zhao, Rohan Varma, Omkar Salpekar, Pieter Noordhuis, Teng Li,
  Adam Paszke, Jeff Smith, Brian Vaughan, Pritam Damania, and Soumith Chintala.
\newblock Pytorch distributed: Experiences on accelerating data parallel
  training.
\newblock {\em Proc. VLDB Endow.}, 2020.

\bibitem{blog}
Faith~Xu Matteo~Interlandi, Karla~Saur.
\newblock {Accelerate traditional machine learning models on GPU with ONNX
  Runtime}.
\newblock
  \url{https://cloudblogs.microsoft.com/opensource/2020/09/29/accelerate-machine-learning-models-gpu-onnx-runtime-hummingbird/},
  2020.

\bibitem{tree-fpga}
M.~{Owaida}, H.~{Zhang}, C.~{Zhang}, and G.~{Alonso}.
\newblock Scalable inference of decision tree ensembles: Flexible design for
  cpu-fpga platforms.
\newblock In {\em 2017 27th International Conference on Field Programmable
  Logic and Applications (FPL)}, pages 1--8, Sep. 2017.

\bibitem{weld}
Shoumik Palkar, James Thomas, Deepak Narayanan, Pratiksha Thaker, Rahul
  Palamuttam, Parimajan Negi, Anil Shanbhag, Malte Schwarzkopf, Holger Pirk,
  Saman Amarasinghe, et~al.
\newblock Evaluating end-to-end optimization for data analytics applications in
  weld.
\newblock {\em Proceedings of the VLDB Endowment}, 11(9):1002--1015, 2018.

\bibitem{pytorch}
Adam Paszke, Sam Gross, Soumith Chintala, Gregory Chanan, Edward Yang, Zachary
  DeVito, Zeming Lin, Alban Desmaison, Luca Antiga, and Adam Lerer.
\newblock Automatic differentiation in pytorch.
\newblock In {\em NIPS-W}, 2017.

\bibitem{scikit}
Fabian Pedregosa, Ga\"{e}l Varoquaux, Alexandre Gramfort, Vincent Michel,
  Bertrand Thirion, Olivier Grisel, Mathieu Blondel, Peter Prettenhofer, Ron
  Weiss, Vincent Dubourg, Jake Vanderplas, Alexandre Passos, David Cournapeau,
  Matthieu Brucher, Matthieu Perrot, and \'{E}douard Duchesnay.
\newblock Scikit-learn: Machine learning in python.
\newblock {\em J. Mach. Learn. Res.}, 12:2825--2830, November 2011.

\bibitem{DBLP:journals/sigmod/PolyzotisRWZ18}
Neoklis Polyzotis, Sudip Roy, Steven~Euijong Whang, and Martin Zinkevich.
\newblock Data lifecycle challenges in production machine learning: {A} survey.
\newblock {\em {SIGMOD} Record}, 47(2):17--28, 2018.

\bibitem{dsonds}
Fotis Psallidas, Yiwen Zhu, Bojan Karlas, Matteo Interlandi, Avrilia Floratou,
  Konstantinos Karanasos, Wentao Wu, Ce~Zhang, Subru Krishnan, Carlo Curino,
  and Markus Weimer.
\newblock Data science through the looking glass and what we found there, 2019.

\bibitem{froid}
Karthik Ramachandra, Kwanghyun Park, K.~Venkatesh Emani, Alan Halverson,
  C\'{e}sar Galindo-Legaria, and Conor Cunningham.
\newblock Froid: Optimization of imperative programs in a relational database.
\newblock {\em Proc. VLDB Endow.}, 11(4):432–444, December 2017.

\bibitem{horovod}
Alexander Sergeev and Mike~Del Balso.
\newblock Horovod: fast and easy distributed deep learning in tensorflow, 2018.

\bibitem{tree-gpu}
Toby Sharp.
\newblock Implementing decision trees and forests on a gpu.
\newblock In David Forsyth, Philip Torr, and Andrew Zisserman, editors, {\em
  Computer Vision -- ECCV 2008}, pages 595--608, Berlin, Heidelberg, 2008.
  Springer Berlin Heidelberg.

\bibitem{enterprise-app-lifespan}
Credit Suisse.
\newblock The apps revolution manifesto—volume 1: The technologies.
\newblock \url{https://aka.ms/enterprise-application-lifespan}, 2012.

\bibitem{numpy}
S.~van~der Walt, S.~C. Colbert, and G.~Varoquaux.
\newblock The numpy array: A structure for efficient numerical computation.
\newblock {\em Computing in Science Engineering}, 13(2):22--30, March 2011.

\end{thebibliography}

\balance
\pagebreak 
\appendix
\section{Artifact Appendix}

\subsection{Abstract}

{\em Hummingbird compiles trained traditional ML models into tensor computation for faster inference. Hummingbird allows users to score models both on CPU and hardware accelerators. }

\subsection{Artifact check-list}

{\small
\begin{itemize}
  \item {\bf Program: } PyTorch, ONNX Runtime, TVM.
  \item {\bf Data set: } Fraud, Epsilon, Year, Covtype, Higgs, Airline, Iris, Nomao, OpenMLCC-18.
  \item {\bf Run-time environment: } Ubuntu 18.04. 
  \item {\bf Hardware: } Azure NC6 v2 machine. 
  \item {\bf Experiments: } tree-models (Random Forest, XGBoost, LightGBM), operators ( LogisticRegression,
            SGDClassifier, 
            LinearSVC, 
            NuSVC,
            SVC,
            BernoulliNB,
            MLPClassifier,
            DecisionTreeClassifier,
            Binarizer,
            MinMaxScaler,
            Normalizer, 
            PolynomialFeatures,
            StandardScaler), end-to-end pipelines.
  \item {\bf Public link: } \url{https://github.com/microsoft/hummingbird}.
  \item {\bf Code licenses: } MIT.
\end{itemize}

\subsection{Description}

\subsubsection{How to access}

Hummingbird is open source and can be accessed directly from \url{https://github.com/microsoft/hummingbird}. Otherwise, Hummingbird can also be downloaded from pip with \texttt{pip install hummingbird-ml}. 

\subsubsection{Hardware dependencies} No specific hardware dependencies. The artifact has been evaluated on different NVIDIA GPU generations (K80, P100, V100) but it should work on any hardware supported by the target DNN runtime.

\subsubsection{Software dependencies}
Hummingbird requires 
Python >= 3.5, 
numpy>=1.15, onnxconverter-common>=1.6.0, scikit-learn>=0.21.3, torch>=1.3.1. Additional dependencies for reproducing the results are onnxruntime >= 1.0, onnxmltools>=1.6.0, xgboost>=0.90 and lightgbm>=2.2, psutil, memory-profiler.

\subsubsection{Data sets}
For the experiments on tree algorithms we used Fraud~\footnote{\url{https://www.kaggle.com/mlg-ulb/creditcardfraud}}, Epsilon~\footnote{\url{https://www.csie.ntu.edu.tw/~cjlin/libsvmtools/datasets/binary.html}}, Year~\footnote{\url{https://archive.ics.uci.edu/ml/datasets/yearpredictionmsd}}, Covtype~\footnote{\url{https://archive.ics.uci.edu/ml/datasets/covertype}}, Higgs~\footnote{\url{https://archive.ics.uci.edu/ml/datasets/HIGGS}}, and Airline~\footnote{\url{http://kt.ijs.si/elena_ikonomovska/data.html}}. For the experiments on operators we instead used Iris~\footnote{\url{https://archive.ics.uci.edu/ml/datasets/iris}}.
Finally, for the pipeline experiments we used OpenML-CC18~\cite{openml-cc18}. The experiment scripts automate the download and preparation of all the datasets.

\subsection{Installation}

Hummingbird can be installed from pip with \texttt{pip install hummingbird-ml} or by cloning the code available on GitHub and by calling \texttt{python setup.py install} from the main directory. Hummingbird will automatically detect the available backends at runtime. We refer to \url{https://github.com/microsoft/hummingbird/blob/master/TROUBLESHOOTING.md} for problems related to installations.

\subsection{Experiment workflow}
The scripts for the experiments are divided in three main folders: \emph{trees}, \emph{operators} and \emph{pipelines}. Each folder contains a README.md file containing the specific instructions for that particular set of experiments.

\stitle{Trees:} 
This directory contains the script to generate the result of Section 6.1.1.
We suggest to start with running \texttt{python run.py -dataset fraud,year,covtype,epsilon} (skipping higgs/airline) because the complete script (which can be run with just \texttt{python run.py}) over all backends and datasets takes more than one day to complete. After the script is run for the first time, the datasets and trained models are cached (in datasets and models folders, respectively), so that following executions will be faster. Serveral other arguments can be changed in the script (e.g., batch size, number of trees, etc.).

The output of the above commands is a JSON file reporting the training time and accuracy (if the model is not cached), and prediction (process) time in seconds, as well the peak memory used. The baseline is then compared against Hummingbird with PyTorch (hb-pytorch), TorchScript (hb-torchscript) and TVM (hb-tvm) backends. The entry \texttt{is\_same\_output} specifies whether the results of the translated models match those of the baseline (up to a tolerance of 10\^-6). If the result is \texttt{false}, the script can be re-run with the \texttt{-validate} flag on to check the percentage of wrong results. The \texttt{-gpu} flag can be used to run the experiments on GPU.

\stitle{Operators:} 
This directory contains the scripts to reproduce the experiments of Section 6.1.2. The scripts are configured to run scikit-learn and compare it against ONNX-ML, TorchScript and TVM (the last 2 using Hummingbird), for the Iris dataset over 1 core, and with batch of 1M. \texttt{python run.py} runs the benchmark for CPU, \texttt{python run.py -gpu} runs the benchmark for GPU.

\stitle{Pipelines:} 
This directory contains the script to reproduce the experiments of Section 6.3. There are two main scripts to run for this experiment: 
\begin{itemize}
    \item \texttt{openml\_pipelines.py} is used to download and train all the scikit-learn pipelines of the openML-CC18 benchmark.
    \item \texttt{run.py} is used to run evaluate the performance of scikit-learn and Hummingbird over the trained pipelines.
\end{itemize}

This experiment is composed of two steps.
The first step in this experiment is the generation of the prediction pipelines. This can achieved by running \texttt{python openml\_pipelines.py | tee openml-cc18.log}
This script takes several hours to run.
While executing, this script will log the number of successfully trained pipelines, as well as additional statistics. Once completed, the openml-cc18.log file contains the statistics. Per task statistics are logged into the relative folder.

Once the first step is completed, in the second step we evaluate the scoring time of the generated pipelines, and compare the speed-ups introduced by Hummingbird against scikit-learn. This experiment can be executed both on CPU and GPU, and in both cases it takes about an hour. \texttt{python run.py} runs inference over all the generated pipelines, while \texttt{python run.py -gpu} can be used for GPU execution.

 \subsection{Evaluation and expected result}
 In May we open sourced Hummingbird (blog post: \url{https://azuredata.microsoft.com/articles/ebd95ec0-1eae-44a3-90f5-c11f5c916d15}). Since then we have been pushing our internal code into the open source repository, but the 2 versions do not match yet. Specifically:
\begin{itemize}
    \item TVM integration is not complete. In our internal version we re-implemented all the operators directly in TVM's Relay but this is not a good strategy in the long term. In the open source version, we directly export Relay graphs from PyTorch models. However the exporter does not cover PyTorch 100\% yet. We are however working with the TVM community for bringing full support of TVM in Hummingbird (we suggest to check the related issue \#232 on Hummingbird's GitHub if interested). In practice, this means that: (1) not all operators are currently exportable into TVM; and (2) the performance we reported in the paper for TVM can be a bit different.
    \item The optimizer is not yet open sourced. This means that Figures 9 and 10 are not reproducible as of now. We hope to be able to bring the optimizer open source in the coming months.
\end{itemize} 

Besides the above two limitations, the scripts allow the reproduction of the following main results of the paper:
\begin{itemize}
    \item \texttt{trees} allows the reproduction of the results of Tables 7, 9 and 10. Please check the above description for specifics.
    \item \texttt{operators} allows the reproduction of the results of Table 11 (however not all operators will run on the TVM backend). Again, please check the related description for specifics.
    \item \texttt{pipelines} allows the reproduction of the results of Figure 12. Also in this case we don't cover yet 100\% of the operators, but we are close.
\end{itemize}

Keep in mind that running all the experiments for completely reproducing the results will take several days. 

\subsection{Experiment customization}
The above mentioned scripts can be customized by running them with different input arguments. For instance, Table 8 in the paper can be reproduced by setting the batch size to 1 (using the \texttt{-batch\_size} argument.) in the \texttt{run.py} script.

\subsection{Notes}
The numbers in the paper were run on the reported VM, however:
\begin{itemize}
    \item As this is an Azure VM, the underlying machine can receive upgrades necessitating  the reinstallation of the NVidia drivers.
    \item The original experiments were run inside the context of an Nvidia-docker container.  This setup should not have a large impact on results
\end{itemize}

Additionally, a few operators are not yet available in the open source version of Hummingbird, therefore the final coverage reported in the log file for the pipelines will be different than the one reported in the paper. To check the expected coverage once all the operators are open source, the script allows to add new operators. The same consideration holds for the operators experiment.

As a final note: to allow third-party reproducibility, we are open sourcing all the scripts used for the experiments.
\subsection{AE Methodology}

Submission, reviewing and badging methodology:

\begin{itemize}
  \item \url{https://www.usenix.org/conference/osdi20/call-for-artifacts}
\end{itemize}

\end{document}